\PassOptionsToPackage{table}{xcolor}
\documentclass{article} 
\usepackage{iclr2027_conference,times}

\usepackage{amsmath,amsfonts,bm}

\definecolor{readcolor}{HTML}{7C1D6F}
\definecolor{writecolor}{HTML}{089099}

\newcommand{\textread}{\textcolor{readcolor}{{Read}}}
\newcommand{\textwrite}{\textcolor{writecolor}{{Write}}}
\newcommand{\mathread}{\mathop{\text{\textread}}\nolimits}
\newcommand{\mathwrite}{\mathop{\text{\textwrite}}\nolimits}

\def\eqref#1{equation~\ref{#1}}

\def\1{\bm{1}}

\def\bdelta{\bm{\delta}}

\DeclareMathAlphabet{\mathsfit}{\encodingdefault}{\sfdefault}{m}{sl}
\SetMathAlphabet{\mathsfit}{bold}{\encodingdefault}{\sfdefault}{bx}{n}

\DeclareMathOperator*{\argmin}{arg\,min}

\usepackage{url}
\usepackage{booktabs}
\usepackage[table]{xcolor}
\usepackage{graphicx}
\graphicspath{{media/}}   

\definecolor{linkblue}{RGB}{40,80,190}

\usepackage[
    colorlinks=true,
    linkcolor=linkblue,
    citecolor=linkblue,
    urlcolor=linkblue
]{hyperref}

\usepackage[capitalise,noabbrev]{cleveref}
\usepackage{float}
\usepackage{tikz}
\usepackage[skins]{tcolorbox}
\usepackage{subcaption}
\usepackage{multirow}
\usepackage{wrapfig}
\usepackage{needspace}
\usepackage[section]{placeins}
\usepackage{enumitem}
\usepackage{mathtools}

\usetikzlibrary{arrows.meta,positioning,calc,backgrounds,decorations.markings,decorations.pathreplacing,fit}

\definecolor{memstroke}{HTML}{045275}
\colorlet{memfill}{memstroke!12!white}
\colorlet{memghost}{memstroke!35!white}
\definecolor{residcol}{HTML}{52514E}
\definecolor{resnodestroke}{HTML}{7CCBA2}
\colorlet{residfill}{resnodestroke!12!white}
\colorlet{resnodefill}{resnodestroke!30!white}
\colorlet{blkstroke}{residcol}
\definecolor{blkfill}{HTML}{F1F1F1}
\colorlet{prodstroke}{residcol}
\colorlet{prodfill}{prodstroke!8!white}
\def\figboxcorners{2pt}
\def\figlegendcorners{1.15pt}

\tikzset{ndmstyle/.style={
    x=1cm, y=1cm,
    >={Stealth[round,length=2.3mm,width=1.9mm]},
    font=\footnotesize,
    memnode/.style={
        rounded corners=\figboxcorners, draw=memstroke, fill=memfill,
        line width=0.7pt, inner xsep=3pt, inner ysep=0pt, font=\small,
        minimum width=1.10cm, minimum height=0.60cm},
    memghostnode/.style={
        rounded corners=\figboxcorners, draw=memstroke, fill=memfill!45,
        line width=0.7pt, inner xsep=3pt, inner ysep=0pt, font=\footnotesize,
        dash pattern=on 2.2pt off 1.6pt,
        minimum width=1.10cm, minimum height=0.60cm},
    resnode/.style={
        rounded corners=\figboxcorners, draw=resnodestroke, fill=resnodefill,
        line width=0.7pt, minimum width=1.10cm, minimum height=0.60cm,
        inner sep=1pt, font=\footnotesize},
    blk/.style={
        rounded corners=\figboxcorners, draw=blkstroke, fill=blkfill,
        line width=0.7pt, inner sep=0pt, font=\footnotesize,
        minimum width=2.00cm, minimum height=0.72cm},
    nrm/.style={
        rounded corners=\figboxcorners, draw=blkstroke, fill=white,
        line width=0.6pt, inner xsep=3pt, inner ysep=0pt, font=\tiny,
        minimum height=0.46cm},
    tape/.style={
        rounded corners=\figboxcorners, draw=blkstroke, fill=white,
        line width=0.6pt, inner xsep=4pt, inner ysep=0pt, font=\tiny,
        minimum height=0.40cm},
    oplus/.style={
        rectangle, rounded corners=\figboxcorners,
        draw=residcol, fill=white, line width=0.65pt,
        inner sep=0pt, minimum width=5.0mm, minimum height=5.0mm,
        font=\normalsize},
    otimes/.style={
        rectangle, rounded corners=\figboxcorners,
        draw=residcol, fill=white, line width=0.65pt,
        inner sep=0pt, minimum width=5.0mm, minimum height=5.0mm,
        font=\normalsize},
    reswire/.style={draw=residcol, line width=0.8pt, dash pattern=on 4pt off 2.6pt},
    memwire/.style={draw=residcol, line width=1.8pt},
    statearr/.style={draw=residcol, line width=1.8pt, ->},
    tapwire/.style={draw=residcol, line width=0.8pt, dash pattern=on 4pt off 2.6pt},
    junction/.style={circle, fill=residcol, inner sep=0pt, minimum size=1.7mm},
    sym/.style={font=\small\bfseries, inner sep=1.5pt},
    prod/.style={
        rounded corners=\figboxcorners, draw=prodstroke, fill=prodfill,
        line width=0.7pt,
        inner xsep=4pt, inner ysep=2pt, font=\small,
        minimum width=0.62cm, minimum height=0.50cm},
    DMop/.style={
        rectangle, rounded corners=\figboxcorners,
        draw=residcol, fill=white, line width=0.65pt,
        inner xsep=4pt, inner ysep=1pt, minimum width=1.35cm,
        minimum height=0.62cm, font=\footnotesize},
}}

\newcommand{\architecturelegend}{%
\begin{tikzpicture}[
    x=1cm, y=1cm,
    >={Stealth[round,length=2.3mm,width=1.9mm]},
    font=\scriptsize,
    leglabel/.style={anchor=west, inner xsep=2pt, inner ysep=0pt},
    legswatch/.style={rounded corners=\figlegendcorners, line width=0.65pt,
        minimum width=5.0mm, minimum height=3.4mm, inner sep=0pt}]

\node[legswatch, draw=resnodestroke, fill=resnodefill] (legres) at (0,0) {};
\node[leglabel] at (legres.east) {residual stream};

\node[legswatch, draw=memstroke, fill=memfill] (legmem) at (2.85,0) {};
\node[leglabel] at (legmem.east) {memory stream};

\node[rectangle, rounded corners=\figlegendcorners,
      draw=residcol, fill=white, line width=0.65pt,
      minimum width=5.92mm, minimum height=3.68mm, inner sep=0pt]
      (legop) at (5.45,0) {};
\node[leglabel] at (legop.east) {operator};

\draw[draw=residcol, line width=0.8pt,
      dash pattern=on 4pt off 2.6pt, ->] (6.90,0) -- (7.80,0);
\node[leglabel] at (7.80,0) {vector};

\draw[draw=residcol, line width=1.8pt, ->] (9.10,0) -- (10.00,0);
\node[leglabel] at (10.00,0) {matrix};

\end{tikzpicture}%
}

\newcommand{\DMpanela}{%
\begin{tikzpicture}[ndmstyle]

\def\axres{0.00}
\def\axsub{1.75}
\def\axread{3.10}
\def\axwriteleg{3.30}
\def\axmem{4.80}
\def\ayfoot{0.10}
\def\aystate{0.10}
\def\ayread{0.65}
\def\aymerge{1.50}
\def\ayblock{3.00}
\def\ayout{4.25}
\def\aywrite{5.05}
\def\aynext{6.35}
\def\ayhead{6.35}
\def\DMstreamhalfwidth{0.74}

\begin{scope}[on background layer]
  \fill[memfill!55, rounded corners=\figboxcorners]
      (\axmem-\DMstreamhalfwidth,\ayfoot-0.50) rectangle
      (\axmem+\DMstreamhalfwidth,\ayhead+0.50);
  \fill[residfill, rounded corners=\figboxcorners]
      (\axres-\DMstreamhalfwidth,\ayfoot-0.50) rectangle
      (\axres+\DMstreamhalfwidth,\ayhead+0.50);
  \path
      (\axres-0.84,\ayfoot-0.50) rectangle
      (\axmem+0.84,\ayhead+0.50);
\end{scope}

\node[resnode,minimum width=1.20cm,minimum height=0.60cm,
      inner sep=0pt,font=\footnotesize]
      (axin) at (\axres,\ayfoot) {$\mathbf{x}_l$};
\node[resnode,minimum width=1.20cm,minimum height=0.60cm,
      inner sep=0pt,font=\footnotesize]
      (axout) at (\axres,\ayhead) {$\mathbf{x}_{l+1}$};
\draw[reswire] (axin.north) -- (\axres,\ayout-0.20);
\draw[reswire,->] (\axres,\ayout+0.20) -- (axout.south);

\node[memnode,minimum width=1.20cm,minimum height=0.60cm,
      inner sep=0pt,font=\footnotesize]
      (asin) at (\axmem,\aystate) {$\mathbf M_l$};
\node[DMop] (awrite) at (\axmem,\aywrite) {\textwrite};
\node[memnode,minimum width=1.20cm,minimum height=0.60cm,
      inner sep=0pt,font=\footnotesize]
      (asout) at (\axmem,\aynext) {$\mathbf M_{l+1}$};
\draw[statearr] (asin.north) -- (awrite.south);
\draw[statearr] (awrite.north) -- (asout.south);

\node[DMop] (aread) at (\axread,\ayread) {\textread};
\draw[statearr] (asin.west) -- (\axmem-0.75,\aystate) |- (aread.east);
\node[junction] at (\axres,\ayread) {};
\draw[tapwire,->] (\axres,\ayread) -- (aread.west);

\node[blk] (asub) at (\axsub,\ayblock) {Sublayer};
\draw[tapwire,->] (aread.north) -- (\axread,\aymerge)
    -- (\axsub,\aymerge) -- (asub.south);

\draw[tapwire] (asub.north) -- (\axsub,\ayout);
\node[junction] at (\axsub,\ayout) {};
\draw[tapwire,->] (\axsub,\ayout) -- (\axres+0.20,\ayout);
\node[oplus] at (\axres,\ayout) {$+$};
\draw[tapwire,->] (\axsub,\ayout) -- (\axwriteleg,\ayout) |- (awrite.west);

\end{tikzpicture}%
}

\colorlet{mhcstate}{memstroke}
\colorlet{mhcstatefill}{memfill}
\colorlet{mhcbandfill}{residfill}
\colorlet{mhcstreamstroke}{resnodestroke}
\colorlet{mhcstreamnodefill}{resnodefill}
\colorlet{mhcdark}{residcol}
\colorlet{mhcblockstroke}{blkstroke}
\colorlet{mhcblockfill}{blkfill}
\colorlet{mhcmappingstroke}{prodstroke}
\colorlet{mhcmappingfill}{prodfill}

\tikzset{mhcstyle/.style={
    x=1cm, y=1cm,
    >={Stealth[round,length=2.3mm,width=1.9mm]},
    font=\footnotesize,
    mhcmapping/.style={
        rectangle, rounded corners=\figboxcorners,
        draw=mhcdark, fill=white, line width=0.65pt,
        align=center, inner xsep=1.2pt, inner ysep=0.4pt,
        minimum width=0.95cm, minimum height=0.76cm,
        font=\footnotesize},
    mhclayer/.style={
        rounded corners=\figboxcorners, draw=mhcblockstroke,
        fill=mhcblockfill,
        line width=0.75pt, align=center, inner xsep=5pt, inner ysep=3pt,
        minimum width=2.00cm, minimum height=0.70cm, font=\footnotesize},
    mhcnorm/.style={
        rounded corners=\figboxcorners, draw=mhcblockstroke, fill=white,
        line width=0.6pt, inner xsep=4pt, inner ysep=2pt, font=\tiny},
    mhcstream/.style={
        rounded corners=\figboxcorners, draw=mhcstreamstroke,
        fill=mhcstreamnodefill,
        line width=0.7pt, minimum width=1.10cm, minimum height=0.60cm,
        inner sep=1pt, font=\footnotesize},
    mhcplus/.style={
        rectangle, rounded corners=\figboxcorners,
        draw=mhcdark, fill=white, line width=0.7pt,
        minimum width=5.0mm, minimum height=5.0mm, inner sep=0pt,
        font=\normalsize},
    mhcwire/.style={draw=mhcdark, line width=0.8pt, dash pattern=on 4pt off 2.6pt, ->},
    mhccondition/.style={draw=mhcdark, line width=1.8pt, ->},
    mhcmapwire/.style={draw=mhcdark, line width=0.8pt, dash pattern=on 4pt off 2.6pt, ->},
    mhcstreamwire/.style={draw=mhcdark, line width=1.8pt, ->},
    mhctap/.style={circle, fill=mhcdark, inner sep=0pt,
        minimum size=1.7mm},
    mhcnote/.style={font=\tiny, text=mhcdark, align=center},
}}

\newcommand{\mhcarchitecture}{%
\begin{tikzpicture}[mhcstyle]

\def\mhcstreamhalfwidth{0.765}
\begin{scope}[on background layer]
  \fill[mhcbandfill, rounded corners=\figboxcorners]
      (-1.55-\mhcstreamhalfwidth,-0.307) rectangle
      (-1.55+\mhcstreamhalfwidth,7.145);
  \path
      (-2.40,-0.35) rectangle (3.85,7.15);
\end{scope}

\node[mhcstream] (xin) at (-1.55,0.245) {$\mathbf{X}_{l}$};
\node[mhctap] (tap) at (-1.55,0.80) {};
\draw[mhcstreamwire,-] (xin.north) -- (tap);

\node[mhcmapping] (hres)  at ( 0.00,4.40) {$\mathbf{H}_{l+1}^{\mathrm{res}}$};
\node[mhcmapping] (hpre)  at ( 2.95,1.20) {$\mathbf{H}_{l+1}^{\mathrm{pre}}$};
\node[mhcmapping] (hpost) at ( 2.95,4.40) {$\mathbf{H}_{l+1}^{\mathrm{post}}$};

\draw[mhccondition] (xin.east) -- (0.00,0.245) -- (hres.south);
\draw[mhccondition,-] (xin.east) -- (3.70,0.245) -- (3.70,4.40);
\draw[mhccondition] (3.70,1.20) -- (hpre.east);
\draw[mhccondition] (3.70,4.40) -- (hpost.east);

\node[mhcplus] (preprod) at (1.60,1.20) {$\bullet$};
\draw[mhccondition] (tap) -- (0.80,0.80) |- (preprod.west);
\draw[mhcmapwire] (hpre.west) -- (preprod.east);

\node[mhclayer] (layer) at (1.60,3.18) {Sublayer};
\draw[mhcwire] (preprod.north) -- (layer.south);

\node[mhcplus] (postprod) at (1.60,4.40) {$\times$};
\draw[mhcwire] (layer.north) -- (postprod.south);
\draw[mhcmapwire] (hpost.west) -- (postprod.east);

\node[mhcplus] (resprod) at (-1.55,4.40) {$\times$};
\draw[mhcstreamwire] (tap) -- (resprod.south);
\draw[mhcstreamwire] (hres.west) -- (resprod.east);

\node[mhcplus] (merge) at (-1.55,5.55) {$+$};
\draw[mhcstreamwire] (resprod.north) -- (merge.south);
\draw[mhcstreamwire] (postprod.north) -- (1.60,5.55) -- (merge.east);

\node[mhcstream] (xout) at (-1.55,6.60) {$\mathbf{X}_{l+1}$};
\draw[mhcstreamwire] (merge.north) -- (xout.south);

\end{tikzpicture}%
}

\tikzset{attnresstyle/.style={
    armcache/.style={rounded corners=\figboxcorners,draw=memstroke,
        fill=memfill,line width=0.7pt},
    armlayer/.style={blk},
    armoperator/.style={DMop},
    armquery/.style={prod,draw=writecolor,fill=writecolor!40!white,
        inner sep=1pt,minimum width=0.80cm,minimum height=0.54cm,
        font=\footnotesize},
    armvector/.style={reswire,->},
    armmatrix/.style={statearr},
    armtap/.style={junction},
    armtext/.style={inner sep=0pt,font=\footnotesize},
}}

\newcommand{\attnresarchitecture}{%
\begin{tikzpicture}[ndmstyle,attnresstyle]
\begin{scope}[on background layer]
  \fill[memfill!55,rounded corners=\figboxcorners]
      (1.46,-0.40) rectangle (2.94,6.85);
  \path (-2.05,-0.40) rectangle (3.04,6.85);
\end{scope}

\draw[armcache] (1.60,-0.30) rectangle (2.80,0.63);
\draw[draw=memstroke!45,line width=0.35pt] (2.20,-0.30) -- (2.20,0.63);
\foreach \yy/\idx in {0.45/0,-0.07/l}{
  \node[armtext] at (1.90,\yy) {$\mathbf k_{\idx}$};
  \node[armtext] at (2.50,\yy) {$\mathbf v_{\idx}$};
}
\foreach \yy in {0.135,0.19,0.245}{
  \fill[residcol] (1.90,\yy) circle[radius=0.22pt];
  \fill[residcol] (2.50,\yy) circle[radius=0.22pt];
}
\coordinate (cachein) at (2.20,0.63);

\node[armquery] (query) at (-1.60,2.05) {$\mathbf q_l$};
\node[armoperator] (read) at (0,2.05) {\textread};
\node[armlayer] (layer) at (0,3.00) {Sublayer};
\node[armoperator] (write) at (2.20,4.25) {\textwrite};
\draw[armmatrix] (cachein) -- (write.south);
\node[armtap] (readtap) at (2.20,2.05) {};
\draw[armmatrix] (readtap.west) -- (read.east);
\draw[armvector] (query.east) -- (read.west);
\draw[armvector] (read.north) -- (layer.south);
\draw[armvector] (layer.north) |- (write.west);

\path[fill=memfill,rounded corners=\figboxcorners]
    (1.60,5.48) rectangle (2.80,6.70);
\begin{scope}
  \clip[rounded corners=\figboxcorners] (1.60,5.48) rectangle (2.80,6.70);
  \fill[memstroke!23!white] (1.60,5.48) rectangle (2.80,5.91);
\end{scope}
\draw[draw=memstroke,line width=0.7pt,rounded corners=\figboxcorners]
    (1.60,5.48) rectangle (2.80,6.70);
\draw[draw=memstroke!45,line width=0.35pt] (2.20,5.48) -- (2.20,6.70);
\draw[draw=memstroke!45,line width=0.35pt] (1.60,5.91) -- (2.80,5.91);
\foreach \yy/\idx in {6.54/0,6.08/l,5.70/{l\!+\!1}}{
  \node[armtext] at (1.90,\yy) {$\mathbf k_{\idx}$};
  \node[armtext] at (2.50,\yy) {$\mathbf v_{\idx}$};
}
\foreach \yy in {6.255,6.31,6.365}{
  \fill[residcol] (1.90,\yy) circle[radius=0.22pt];
  \fill[residcol] (2.50,\yy) circle[radius=0.22pt];
}
\draw[armmatrix] (write.north) -- (2.20,5.48);
\end{tikzpicture}%
}

\definecolor{flowstate}{HTML}{045275}
\colorlet{flowstatefill}{flowstate!12!white}
\definecolor{flowadj}{HTML}{DC3977}%
\definecolor{flowcontext}{HTML}{089099}
\colorlet{flowcontextfill}{flowcontext!8!white}
\definecolor{flowink}{HTML}{52514E}
\definecolor{flowbackground}{HTML}{F1F1F1}
\colorlet{flowghost}{flowink!60!white}

\tikzset{flowstyle/.style={
    x=0.86cm, y=1cm,
    >={Stealth[round,length=2.2mm,width=1.8mm,inset=0.715mm,line width=0.6pt]},
    font=\footnotesize,
    stored/.style={rounded corners=1.8pt, draw=flowstate, fill=flowstatefill, line width=0.9pt,
        inner xsep=3pt, inner ysep=0pt, font=\small, minimum width=1.12cm, minimum height=0.42cm},
    mat/.style={rounded corners=1.8pt, draw=flowstate, fill=white, line width=0.9pt,
        inner xsep=3pt, inner ysep=0pt, font=\small, minimum width=1.12cm, minimum height=0.42cm},
    virtS/.style={rounded corners=1.8pt, draw=flowghost, fill=none, line width=0.6pt, dotted, text=flowghost,
        inner xsep=3pt, inner ysep=0pt, font=\small, minimum width=1.12cm, minimum height=0.42cm},
    adjmat/.style={rounded corners=1.8pt, draw=flowadj, fill=white, line width=0.9pt,
        inner xsep=2pt, inner ysep=0pt, font=\footnotesize, minimum width=1.12cm, minimum height=0.42cm},
    virtA/.style={rounded corners=1.8pt, draw=flowadj!45, fill=none, line width=0.6pt, dotted, text=flowadj!60,
        inner xsep=2pt, inner ysep=0pt, font=\footnotesize, minimum width=1.12cm, minimum height=0.42cm},
    tape/.style={rounded corners=1.5pt, draw=flowcontext, fill=flowcontextfill, line width=0.6pt,
        inner xsep=4pt, inner ysep=0pt, font=\small, minimum height=0.32cm, minimum width=0.55cm},
    win/.style={rounded corners=5pt, draw=black!45, line width=0.6pt, dash pattern=on 3pt off 2pt},
    midarrow/.style={postaction={decorate},
        decoration={markings,mark=at position 0.5 with {
            \pgftransformxshift{1.1mm}\arrow{>}}}},
    prejunctionarrow/.style={postaction={decorate},
        decoration={markings,mark=at position 0.25 with {\arrow{>}}}},
    flowline/.style={draw=flowink, line width=0.7pt},
    memwire/.style={flowline, midarrow},
    memwireJ/.style={flowline, prejunctionarrow},
    adjwire/.style={flowline, midarrow},
    adjwireJ/.style={flowline, prejunctionarrow},
    layerwire/.style={flowline, midarrow},
    layerwireJ/.style={flowline, prejunctionarrow},
    layerwireA/.style={flowline, midarrow},
    layerwireAJ/.style={flowline, prejunctionarrow},
    tapewire/.style={flowline},
    tapflow/.style={tapewire, midarrow},
    couplingflow/.style={flowline, midarrow},
    virtcoupling/.style={flowline, draw=flowghost, dotted, midarrow},
    junction/.style={circle, fill=flowink, inner sep=0pt, minimum size=1.7mm},
    lab/.style={font=\footnotesize, inner sep=1pt},
    hdr/.style={font=\scriptsize\bfseries, inner sep=1pt},
}}

\tikzset{
  sm/.style={font=\scriptsize, inner sep=1pt},
}
\tikzset{
  sbox/.style={rounded corners=1.8pt, line width=0.9pt, minimum width=1.12cm, minimum height=0.42cm, inner sep=0pt, outer sep=0pt, font=\small},
  smat/.style={sbox, draw=flowstate, fill=white},
  sstored/.style={sbox, draw=flowstate, fill=flowstatefill},
  sghost/.style={sbox, draw=flowghost, line width=0.6pt, dash pattern=on 1.6pt off 1.2pt, text=flowghost, fill=white},
  gmat/.style={sbox, draw=flowadj, fill=white},
  gghost/.style={sbox, draw=flowadj!55, line width=0.6pt, dash pattern=on 1.6pt off 1.2pt, text=flowadj!70, fill=white},
}

\newcommand{\ndmflows}{%
\begin{tikzpicture}[flowstyle]
\def\P{2.2}
\def\yF{0.00} \def\yT{-1.05} \def\yBS{-2.10} \def\yBA{-2.95}
\begin{scope}[on background layer]
  \fill[flowstatefill!35, rounded corners=4pt] (-0.85,\yF-0.40) rectangle ({6*\P+0.85},\yF+0.40);
  \fill[flowbackground!70, rounded corners=4pt] (-0.85,\yBA-0.40) rectangle ({6*\P+0.85},\yBS+0.40);
  \foreach \w in {0,3}{
    \draw[win] ({\w*\P+0.90},\yF-0.32) rectangle ({(\w+3)*\P+0.74},\yF+0.32);
    \draw[win] ({\w*\P+0.90},\yBA-0.32) rectangle ({(\w+3)*\P+0.74},\yBA+0.32);
  }
\end{scope}
\node[hdr,flowstate,rotate=90] at (-1.15,\yF) {Forward};
\node[hdr,flowadj,rotate=90] at (-1.15,{(\yBS+\yBA)/2}) {Backward};
\foreach \s/\n/\st in {-1/0/smat, 0/1/sghost, 1/2/sghost, 2/3/smat, 3/4/sghost, 4/5/sghost, 5/6/sstored}{
  \node[\st] (f\n) at ({\n*\P},\yF) {$\mathbf M_{\s}$};
}
\foreach \s/\n/\st in {0/1/gghost, 1/2/gghost, 2/3/gmat, 3/4/gghost, 4/5/gghost, 5/6/gmat}{
  \node[\st] (g\n) at ({\n*\P},\yBA) {$\mathrm d\mathbf M_{\s}$};
}
\node[smat]    (b0) at (0,\yBS)        {$\mathbf M_{-1}$};
\node[smat]    (b3) at ({3*\P},\yBS)   {$\mathbf M_{2}$};
\node[sstored] (b6) at ({6*\P},\yBS)   {$\mathbf M_{5}$};
\draw[flowline,->] (f6.south) -- (b6.north);
\draw[flowline,->] (b6.west) -- (b3.east);
\draw[flowline,->] (b3.west) -- (b0.east);
\foreach \i/\n/\m in {0/0/1, 1/1/2, 2/2/3, 3/3/4, 4/4/5, 5/5/6}{
  \pgfmathsetmacro{\x}{(\i+0.5)*\P}
  \draw[flowline,->] (f\n.east) -- (f\m.west);
  \ifnum\i=0 \draw[flowline,draw=flowadj!60,->] (g\m.west) -- ({\x+0.09},\yBA); \else \draw[flowline,draw=flowadj!60,->] (g\m.west) -- (g\n.east); \fi
  \fill[flowink] (\x,\yF) circle (0.6mm);
  \fill[flowadj] (\x,\yBA) circle (0.6mm);
  \draw[flowline,->] (\x,\yF+0.62) node[sm,above] {$(\mathbf q,\mathbf k,\mathbf v)_{\i}$} -- (\x,\yF+0.06);
  \node[tape] (c\i) at (\x,\yT) {$\mathbf c_{\i}$};
  \draw[flowline,->] (\x,\yF) -- (c\i.north);
  \draw[flowline] (c\i.south) -- (\x,\yT-0.42);
  \draw[flowline,->] (\x,\yBA+0.48) -- (\x,\yBA+0.07);
  \draw[flowline,draw=flowadj,->] (\x,\yBA) -- (\x,\yBA-0.62) node[sm,below,flowadj] {$\mathrm d(\mathbf q,\mathbf k,\mathbf v)_{\i}$};
}
\foreach \b/\lo/\hi in {0/0.5/2.5, 3/3.5/5.5}{
  \draw[flowline] ({\hi*\P},\yT-0.42) -- ({\b*\P},\yT-0.42);
  \draw[flowline,->] ({\b*\P},\yT-0.42) -- (b\b.north);
  \draw[flowline] (b\b.south) |- ({\hi*\P},\yBA+0.48);
}
\matrix[anchor=north,inner sep=0pt,column sep=3pt,ampersand replacement=\&] at ({3*\P},\yBA-1.17) {
  \node[draw=black!55,fill=white,rounded corners=1.8pt,line width=0.9pt,minimum width=0.5cm,minimum height=0.32cm] {}; \& \node[sm] {materialized}; \&[12pt]
  \node[rounded corners=1.8pt,line width=0.9pt,minimum width=0.5cm,minimum height=0.32cm,draw=black!55,fill=black!14] {}; \& \node[sm] {stored}; \&[12pt]
  \node[draw=flowghost,line width=0.6pt,dash pattern=on 1.6pt off 1.2pt,rounded corners=1.8pt,minimum width=0.5cm,minimum height=0.32cm] {}; \& \node[sm] {not materialized}; \&[12pt]
  \node[inner sep=0pt,minimum width=1.2mm] {}; \& \node[sm] (legdot) {\textwrite{} $l$ and \textread{} for layer $l{+}1$}; \\ };
\fill[flowink] ([xshift=-3pt-0.6mm,yshift=1.6pt]legdot.base west) circle (0.6mm);
\end{tikzpicture}%
}

\usepackage{amsmath,amssymb}
\usepackage{algorithm}
\usepackage[noend]{algpseudocode}

\usepackage{comment}

\title{\centering Depth Memory}

\author{%
\begin{minipage}[t]{\dimexpr\textwidth-2\tabcolsep\relax}
\centering
\textbf{David González-Martínez$^{*1,2,3}$\hspace{0.7em}Philipp Nazari$^{*1,2,3,4}$} \\
\textbf{T. Konstantin Rusch$^{1,2,3,5}$\hspace{0.7em}Shiwei Liu$^{1,2,3}$} \\
\normalfont
$^1$ELLIS Institute Tübingen \quad $^2$Max Planck Institute for Intelligent Systems, Tübingen \\
$^3$Tübingen AI Center \quad $^4$ETH Zürich \quad $^5$Liquid AI \hspace{0.7em}
$^*$Equal contribution
\end{minipage}%
}

\definecolor{darkgreen}{HTML}{006400}

\newif\ifpreprint
\preprinttrue 
\ifpreprint\iclrfinalcopy\fi

\newcommand{\paperlink}[2]{%
    \if\relax\detokenize{#1}\relax
        {\color{linkblue}#2}%
    \else
        \href{#1}{#2}%
    \fi
}
\begin{document}

\maketitle

\begin{abstract}
Residual connections propagate information through a network by additively accumulating layer outputs. Recently proposed alternatives, such as Manifold-Constrained Hyper-Connections (mHC) and Attention Residuals (AttnRes), instead use more expressive depth aggregation mechanisms that improve language modeling performance at scale. In this work, we consider the mixing of representations over depth as a problem of storing and retrieving information across layers. Formalizing this perspective, we introduce \emph{Depth Memory (DM)}, a unified framework that describes the standard pre-layer-norm (PreLN) transformer, mHC, and AttnRes in terms of streams that carry representations through the network. One such stream, for example, could be an explicit memory stream that is queried to enrich the input to every layer, and whose output is subsequently written back into the memory for use by later layers. For AttnRes, this memory stream is a growing cache of explicit layer representations. For mHC, it is an expanded residual stream. Building on this view, we propose \emph{Linear Depth Memory (LDM)}, which formulates information retrieval across depth as a regression problem solved using recursive least squares. The resulting architecture propagates three streams across depth: the standard residual stream, a linear associative memory stream, and an inverse covariance stream. For high-throughput training and inference, we propose a hardware-aware implementation of LDM that reduces peak activation memory and memory traffic. LDM consistently achieves the lowest perplexities among PreLN, mHC and AttnRes baselines at the 400M and 1.5B parameter scales, alongside competitive zero-shot accuracy.
\end{abstract}

\ifpreprint
    \par\vspace{-0.5\baselineskip}
    {\centering
    \raisebox{-0.2\height}{\includegraphics[height=1.2em]{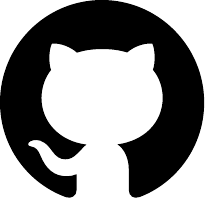}}\hspace{0.35em}%
    \paperlink{https://github.com/davidgonmar/depth-memory}{\textbf{Code}}%
    \hspace{4em}%
    \raisebox{-0.2\height}{\includegraphics[height=1.45em]{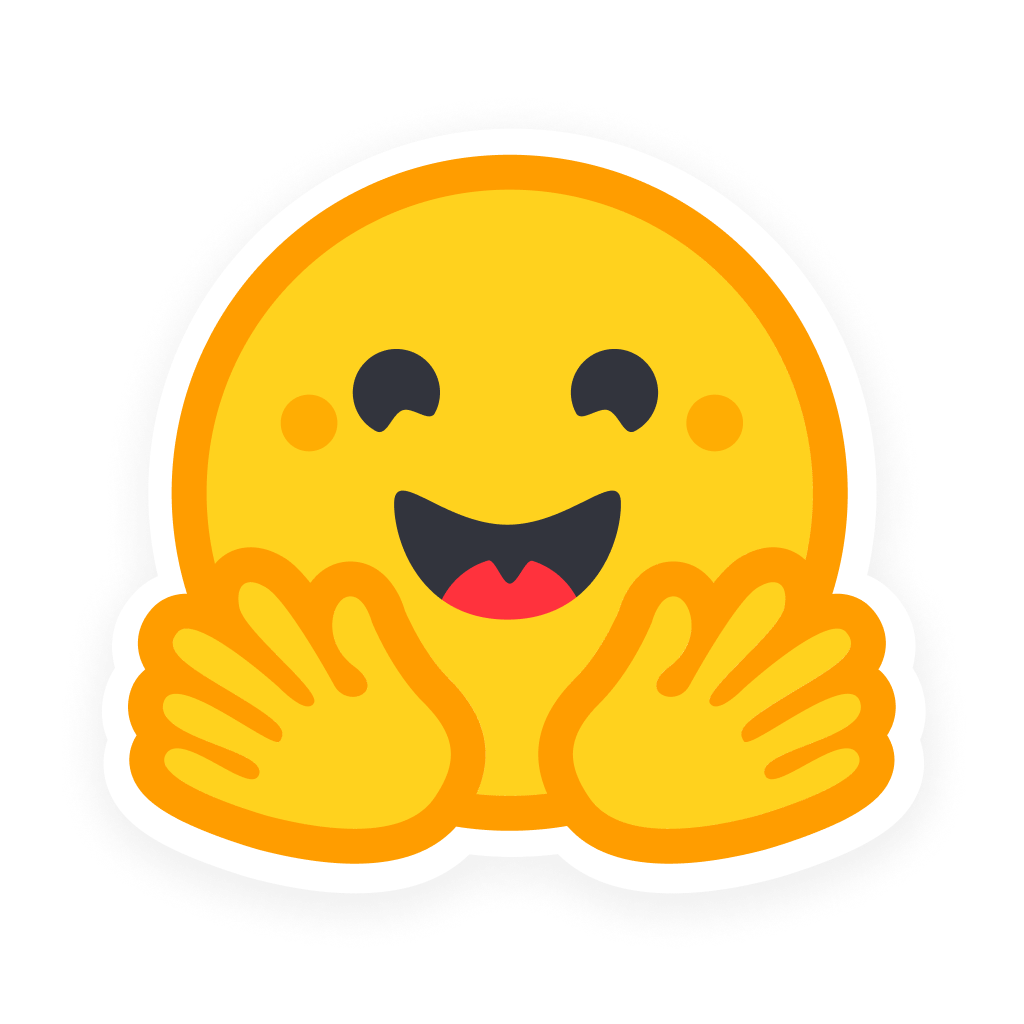}}\hspace{0.35em}%
    \paperlink{https://huggingface.co/collections/davidgonmar/depth-memory-6ac5045a1bd8e798d2ed1392}{\textbf{Checkpoints}}\par
    }
    \lhead{Preprint}
\fi

\section{Introduction}

\begin{wrapfigure}[14]{r}{0.3\textwidth}
    \centering
    \vspace{-\intextsep}
    \begingroup%
\definecolor{sequence}{HTML}{DC3977}%
\definecolor{channel}{HTML}{089099}%
\definecolor{depth}{HTML}{045275}%
\definecolor{edge}{HTML}{52514E}%
\definecolor{neutral}{HTML}{F1F1F1}%
\begin{tikzpicture}[x={(0.6397569cm,0.2584786cm)},y={(-0.6397569cm,0.2584786cm)},z={(0cm,0.8276203cm)},
  scale=0.3967934,font=\scriptsize,
  line join=round,line cap=round,>={Stealth[length=3pt,width=2.8pt]}]
\foreach \k in {0,...,4}{
  \foreach \s in {8,7,...,0}{
    \foreach \i in {0,...,4}{
      \pgfmathtruncatemacro{\j}{\s-\i}
      \ifnum\j>-1\ifnum\j<5
        \def\cube{neutral}
        \ifnum\j=0\ifnum\k=0\def\cube{sequence}\fi\fi
        \ifnum\i=0\ifnum\k=0\def\cube{channel}\fi\fi
        \ifnum\i=0\ifnum\j=0\def\cube{depth}\fi\fi
        \pgfmathsetmacro{\a}{\i+0.045}
        \pgfmathsetmacro{\b}{\j+0.045}
        \pgfmathsetmacro{\c}{\k+0.045}
        \pgfmathsetmacro{\A}{\i+0.955}
        \pgfmathsetmacro{\B}{\j+0.955}
        \pgfmathsetmacro{\C}{\k+0.955}
        \filldraw[fill=\cube!62!white,draw=edge!65,line width=0.13pt]
          (\a,\b,\c)--(\A,\b,\c)--(\A,\b,\C)--(\a,\b,\C)--cycle;
        \filldraw[fill=\cube!82!white,draw=edge!65,line width=0.13pt]
          (\a,\b,\c)--(\a,\B,\c)--(\a,\B,\C)--(\a,\b,\C)--cycle;
        \filldraw[fill=\cube!37!white,draw=edge!65,line width=0.13pt]
          (\a,\b,\C)--(\A,\b,\C)--(\A,\B,\C)--(\a,\B,\C)--cycle;
      \fi\fi
    }
  }
}
\draw[->,sequence,line width=0.5pt] (0,-0.62,-0.42)--(5.55,-0.62,-0.42)
  node[midway,below=2pt,sloped,text=sequence,inner sep=1pt]{Sequence mixing};
\draw[->,channel,line width=0.5pt] (-0.62,0,-0.42)--(-0.62,5.55,-0.42)
  node[midway,below=2pt,sloped,text=channel,inner sep=1pt]{Channel mixing};
\draw[->,depth,line width=0.5pt] (-0.62,5.60,0)--(-0.62,5.60,5.55)
  node[midway,left=3pt,text=depth,rotate=90,anchor=south,inner sep=1pt]{Depth mixing};
\end{tikzpicture}
\endgroup
    \caption{The axes of representation mixing: sequence (tokens), channel (features), and depth (layers).}
    \label{fig:mixing-axes}
    \vspace{-\intextsep}
\end{wrapfigure}
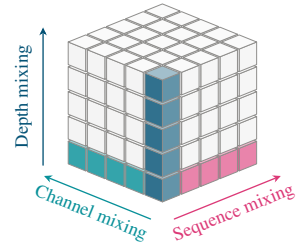

The transformer architecture underpins modern language modeling~\citep{vaswani2023attentionneed}. Since its introduction, architecture research has mainly seen innovation along two axes: sequence mixing, which includes softmax attention~\citep{vaswani2023attentionneed}, linear attention~\citep{pmlr-v119-katharopoulos20a}, and state-space models~\citep{gu2022efficientlymodelinglongsequences}, and channel mixing, through works like gated MLPs~\citep{shazeer2020gluvariantsimprovetransformer} and MoEs~\citep{JMLR:v23:21-0998}. The standard mechanism for propagating information across depth, however, has remained largely unchanged since the introduction of ResNet~\citep{resnet}: layer outputs are accumulated additively through residual connections. Recent works have established alternatives to this standard mechanism at scale~\citep{kimiteam2026kimik3openfrontier,deepseekai2026deepseekv4highlyefficientmilliontoken}. For example, Hyper-Connections (HC,~\citet{zhu2025hyperconnections}) and Manifold-Constrained Hyper-Connections (mHC,~\citet{xie2026mhcmanifoldconstrainedhyperconnections}) maintain an expanded residual stream, while Attention Residuals (AttnRes,~\citet{kimiteam2026attentionresiduals}) explicitly attend to previous layer outputs (\cref{fig:attnres-architecture}). Analogous to sequence and channel mixing operators (\cref{fig:mixing-axes}), we refer to these methods as \emph{depth mixing} mechanisms.

In this work, we approach depth mixing from a different angle. Viewing it as a problem of storing and retrieving information across layers, we introduce \emph{Depth Memory}, a framework that interprets the propagation of information across depth as a problem of managing layer representations using a set of \emph{streams}. At each layer, and independently for each token, a \textread{} operation forms the layer input from these streams, and a \textwrite{} operation updates the streams using the layer output. This framework encompasses the standard residual stream, AttnRes's growing cache of layer representations, and mHC's expanded residual stream, covering both explicit and implicit forms of depth mixing (\cref{fig:architecture})\looseness-1.

Besides taxonomizing existing residual methods using our depth memory framework, we propose instantiating it as a \emph{Linear Depth Memory (LDM)}. LDM maintains three streams: the standard residual stream, a fixed-size matrix-valued memory stream, and an inverse covariance stream (\cref{fig:DM-architecture}). Analogous to test-time regression for sequence mixing~\citep{wang2025test}, we formulate matrix-valued memory stream updates as the solution to a \emph{depth regression} problem, which fits a linear map to reconstruct previous layer outputs from their corresponding keys. LDM solves this optimization problem optimally through recursive least squares (RLS, \cref{sec:lDM}).

Our hardware-aware implementation of LDM addresses efficiency challenges posed by a naive instantiation of the framework. First, we implement an unwinding mechanism that does not require storing each layer's memory stream. Second, we employ a windowed \textwrite{} algorithm that defers parts of the memory updates. Implemented as efficient CuTe DSL kernels, LDM achieves comparable speed to the baselines (\cref{fig:efficiency-results}).

We evaluate LDM on language modeling benchmarks across 400M Llama-like \citep{touvron2023llamaopenefficientfoundation} and 1.5B Qwen-like \citep{yang2025qwen3technicalreport} transformer backbones. LDM achieves lower perplexity than a standard pre-layer-norm (PreLN) transformer, mHC and AttnRes at both scales, along with competitive zero-shot accuracy. These gains hold across standard aspect ratios. We furthermore perform extensive ablations to isolate the contributions of different LDM architectural components.

In summary, our main contributions are:
\begin{enumerate}[
    itemsep=2pt,
    topsep=2pt,
    parsep=0pt,
    leftmargin=2em
]
\item We introduce depth memory, a framework that describes the propagation of information across depth through a collection of streams and their \textread{} and \textwrite{} operations.
\item We instantiate this framework in the form of LDM, a linear depth memory that solves a depth regression problem using recursive least squares to store layer representations in a fixed-size state.\looseness-1
\item We provide an efficient implementation of LDM that combines unwinding and deferred writing techniques to reduce activation memory overhead and memory traffic of LDM.
\item We show that LDM consistently achieves the lowest perplexities among tuned PreLN, mHC, and AttnRes baselines at the 400M and 1.5B scales, while maintaining competitive zero-shot accuracy, and that these gains hold across standard model aspect ratios.\looseness-1
\end{enumerate}

\begin{figure*}[t]
    \centering
    \hbox to \textwidth{%
    \hfil
    \begin{minipage}[b]{3.84cm}
        \centering
        \resizebox*{!}{4.30cm}{\DMpanela}
    \end{minipage}\hfil
    \begin{minipage}[b]{3.20cm}
        \centering
        \resizebox*{!}{4.30cm}{\attnresarchitecture}
    \end{minipage}\hfil
    \begin{minipage}[b]{3.60cm}
        \centering
        \resizebox*{!}{4.30cm}{\mhcarchitecture}
    \end{minipage}\hfil
    }
    \par\vspace{0.5mm}
    \architecturelegend
    \par\vspace{-0.5mm}
    \hbox to \textwidth{%
    \hfil
    \begin{subfigure}[t]{3.84cm}
        \centering
        \captionsetup{skip=0pt}
        \caption{LDM (ours)}
        \label{fig:DM-architecture}
    \end{subfigure}\hfil
    \begin{subfigure}[t]{3.20cm}
        \centering
        \captionsetup{skip=0pt}
        \caption{AttnRes}
        \label{fig:attnres-architecture}
    \end{subfigure}\hfil
    \begin{subfigure}[t]{3.60cm}
        \centering
        \captionsetup{skip=0pt}
        \caption{mHC}
        \label{fig:mhc-architecture}
    \end{subfigure}\hfil
    }
    \par\vspace{-3mm}
    \caption{
    (\subref{fig:DM-architecture}) LDM carries a separate per-token depth memory alongside the residual stream. \textread{} and \textwrite{} are simplified for brevity (the inverse covariance stream and the write address generation are omitted). (\subref{fig:attnres-architecture}) AttnRes reads from a growing key--value memory $\mathbf M_l$ and writes the sublayer output back into it. (\subref{fig:mhc-architecture}) mHC generates its three mappings from the current expanded state and applies them around one transformer branch.}
    \label{fig:architecture}
    \vspace{-1.5em}
\end{figure*}

\section{Background}\label{sec:background}

Consider a standard pre-normalized transformer (PreLN, \citet{layernormplacement}) with $B$ blocks and $L=2B$ layers. We write equations for a single token and suppress its index. Let $\mathbf{x}_l\in\mathbb R^D$ denote its representation after layer $l$, with $\mathbf{x}_0=\mathbf e$ as the token embedding. In PreLN, representations across depth follow the recurrence
\begin{equation}
\label{eq:pre-norm-transformer}
\mathbf{x}_l=\mathbf{x}_{l-1}+\boldsymbol{\delta}_l,
\qquad
\boldsymbol{\delta}_l
=F_l\left(\operatorname{Norm}_l(\mathbf{x}_{l-1})\right),
\qquad l=1,\ldots,L,
\end{equation}
where $\operatorname{Norm}_l$ is typically RMSNorm~\citep{DBLP:journals/corr/abs-1910-07467}, and $F_l$ is either an attention layer or an MLP. Unrolling this recurrence gives
\begin{equation*}
\mathbf{x}_l=\mathbf e+\sum_{i=1}^{l}\boldsymbol{\delta}_i.
\end{equation*}
In particular, residual connections establish a \emph{residual stream} that aggregates layer outputs additively (Figure~\ref{fig:DM-architecture}). Recent work extends this additive aggregation with richer representations and learned mechanisms for mixing information across depth. We review two prominent approaches below, then unify them within our depth memory framework in Section~\ref{sec:DM}.

\subsection{Attention Residuals}

Attention Residuals (AttnRes, \citet{kimiteam2026attentionresiduals}) replace simple additive accumulation of layer outputs via the residual stream with softmax attention over earlier layer outputs. Set $\mathbf v_0=\mathbf k_0=\mathbf e$ and $\mathbf{v}_i = \mathbf{k}_i =  \boldsymbol{\delta}_i$, for $i\geq1$. Full AttnRes forms the input to layer $l$ as
\begin{equation}
\label{eq:attnres-background}
\mathbf a_l=\sum_{i=0}^{l-1}\alpha_{l,i}\mathbf v_i,
\qquad
\alpha_{l,i}=\frac{\exp\left(\mathbf q_{l-1}^\top\operatorname{RMSNorm}(\mathbf k_i)\right)}
{\sum_{j=0}^{l-1}\exp\left(\mathbf q_{l-1}^\top\operatorname{RMSNorm}(\mathbf k_j)\right)}.
\end{equation}
The layer then produces $\boldsymbol{\delta}_l=F_l(\operatorname{Norm}_l(\mathbf a_l))$, which is stored for subsequent layers to attend to. The query $\mathbf q_{l-1}\in\mathbb R^D$ is a learned per-layer parameter. Through the attention mechanism, each layer can selectively weight previous representations, while PreLN gives every previous layer the same weight (\cref{fig:attnres-architecture}).

The above formulation of AttnRes, known as Full AttnRes, incurs memory overhead during inference because it needs to retain previous representations. Block AttnRes groups layers into blocks and summarizes each block by the sum of its layer outputs. At every layer, Block AttnRes forms the input from the embedding, the completed blocks, and the current block's partial sum.\looseness-1

\subsection{Hyper-Connections}

Hyper-Connections (HC, \citet{zhu2025hyperconnections}) expand the residual representation into $N$ streams, collected as columns of $\mathbf X_l\in\mathbb R^{D\times N}$. Three learned mappings take the incoming state $\mathbf X_{l-1}$ and produce vectors $\mathbf H_l^{\mathrm{pre}}(\mathbf X_{l-1}),\mathbf H_l^{\mathrm{post}}(\mathbf X_{l-1})\in\mathbb R^N$ as well as the matrix $\mathbf H_l^{\mathrm{res}}(\mathbf X_{l-1})\in\mathbb R^{N\times N}$. A layer acts on this expanded residual stream via
\begin{equation*}
\boldsymbol{\delta}_l=F_l\left(\operatorname{Norm}_l(\mathbf X_{l-1}\mathbf H_l^{\mathrm{pre}}(\mathbf X_{l-1}))\right),\quad \mathbf X_l =\mathbf X_{l-1}\mathbf H_l^{\mathrm{res}}(\mathbf X_{l-1})
+\boldsymbol{\delta}_l\bigl(\mathbf H_l^{\mathrm{post}}(\mathbf X_{l-1})\bigr)^\top.
\end{equation*}
The pre-mapping $\mathbf H_l^{\mathrm{pre}}$ combines streams into a $D$-dimensional layer input, the post-mapping $\mathbf H_l^{\mathrm{post}}$ distributes the layer output across streams, and the residual mapping $\mathbf H_l^{\mathrm{res}}$ mixes streams (\Cref{fig:mhc-architecture}). Manifold-Constrained Hyper-Connections (mHC, \citet{xie2026mhcmanifoldconstrainedhyperconnections}) constrain this architecture to stabilize signal propagation over depth. In particular, $\mathbf H_l^{\mathrm{res}}$ is constrained to be approximately doubly stochastic through applications of the Sinkhorn-Knopp algorithm~\citep{sinkhorn1967concerning}.

\textbf{Hyper-Connections as a memory.} (m)HC updates the residual stream with a rank-1 matrix, and contracts it to generate layer inputs. Informally, this mechanism can be interpreted as a linear associative memory over depth~\citep{correlationmatrixmemories,kimiteam2026attentionresiduals}. The pre-mapping $\mathbf H_l^{\mathrm{pre}}$ generates a query that reads from the state, and the post-mapping $\mathbf H_l^{\mathrm{post}}$ generates a key under which the layer output is written to the expanded residual stream through an outer product update. The residual mapping $\mathbf H_l^{\mathrm{res}}$ acts as a transition matrix on the state. Since this state combines earlier layer outputs, subsequent reads mix information across depth without explicitly retaining each output. HC and mHC can therefore be understood as performing a form of implicit depth mixing, with both queries and keys generated from the incoming matrix-valued state itself. Building on this perspective, we next establish depth memory as a framework that encompasses both (m)HC and AttnRes, before instantiating the framework as a linear depth memory.

\section{Depth Memory}

\begin{table}[!t]
\centering
\renewcommand{\arraystretch}{1.05}
\caption{Overview of depth mixing methods. Each method maintains streams that are accessed through \textread{} and \textwrite{} operators. LDM uses the RLS update derived in \cref{sec:lDM}, with pseudo-value $\mathbf u_l$ and gain $\boldsymbol{\kappa}_l$. Gates and the inverse covariance update for $\mathbf P_l$ are omitted for brevity.\looseness-1}
\label{tab:depth-mixing}
\setlength{\tabcolsep}{4.8pt}
\begin{tabular}{@{}llll@{}}
\toprule
Method & Streams & State update (\textwrite{}) & Memory readout (\textread{}) \\
\midrule
PreLN
& $(\mathbf x_l)$
& $\mathbf x_l=\mathbf x_{l-1}+\mathbf v_l$
& $\mathbf x_l$ \\
AttnRes
& $((\mathbf K_{\le l},\mathbf V_{\le l}))$
& $\operatorname{Append}(\mathbf k_l,\mathbf v_l)$
& $\operatorname{Attend}(\mathbf M_l,\mathbf q_l)$ \\
mHC
& $(\mathbf X_l)$
& $\mathbf X_l
=\mathbf X_{l-1}\mathbf H_l^{\mathrm{res}}(\mathbf X_{l-1})
+\mathbf v_l\mathbf H_l^{\mathrm{post}}(\mathbf X_{l-1})^\top$
& $\mathbf X_l\mathbf H_{l+1}^{\mathrm{pre}}(\mathbf X_l)$ \\
\midrule
\textbf{LDM}
& $(\mathbf x_l,\mathbf M_l,\mathbf P_l)$
& $(\mathbf x_l,\mathbf M_l)
=(\mathbf x_{l-1}+\mathbf v_l,\,
\mathbf M_{l-1}+\mathbf u_l\boldsymbol{\kappa}_l^\top)$
& $\mathbf x_l+\mathbf M_l\mathbf q_l$ \\
\bottomrule
\end{tabular}
\end{table}

In this section, we first introduce depth memory (DM), a framework that takes a memory-centric view to describe how information in a transformer flows over depth and encompasses PreLN as well as AttnRes and mHC. We then instantiate DM as a linear depth memory, which we formulate as solving a regression problem over depth. Finally, we describe how to use unwinding and windowed updates to implement linear depth memory efficiently.

\subsection{The Framework}
\label{sec:DM}
The depth memory framework assumes a tuple $\mathbf S_l = (\mathbf S_l^{(1)},\ldots,\mathbf S_l^{(P)})$ of $P$ \emph{streams}, which are read from and written to using \textread{} and \textwrite{} operators. For a single stream, we identify $\mathbf S_l = \mathbf S_l^{(1)}$. The input to sublayer $F_l$ is enhanced by reading from the streams $\mathbf S_{l-1}$ using a query $\mathbf q_{l-1}$, and it is updated using keys $\mathbf k_l$ and values $\mathbf v_l$, where the value is the sublayer output $\mathbf v_l=\bm\delta_l$,\looseness-1
\begin{equation*}
\mathbf{\bdelta}_l
    =
    F_l\left(
        \operatorname{Norm}_l
        \left(
            \mathread(\mathbf{S}_{l-1}, \mathbf{q}_{l-1})
        \right)
    \right), \qquad
    \mathbf{\mathbf{S}}_l
    =
    \mathwrite
    \left(
        \mathbf{\mathbf{S}}_{l-1},
        \mathbf{k}_l, \mathbf{v}_l
    \right).
\end{equation*}
Any instantiation of DM requires specifying the tuple $\mathbf S_l$ of streams as well as the \textread{} and \textwrite{} operators. The following paragraphs incorporate PreLN, AttnRes and mHC.

\textbf{PreLN.}
For PreLN there exists exactly one stream, which is the classic residual stream: $\mathbf S_l = \mathbf x_l$. The \textread{} and \textwrite{} operators retrieve the stream and write layer outputs into it,
\begin{equation*}
    \mathread{(\mathbf x_{l-1}, \cdot) = \mathbf x_{l-1}}, \qquad \mathbf x_l = \mathwrite{(\mathbf x_{l-1}, \cdot, \bm{\delta}_l)} = \mathbf x_{l-1} + \bm{\delta}_l.
\end{equation*}
\textbf{AttnRes.}
Like PreLN, AttnRes maintains exactly one stream. This stream consists of a growing set of key--value pairs: $\mathbf S_l = \mathbf M_l = (\mathbf K_{\leq l}, \mathbf V_{ \leq l})$. Here, $\mathbf M_l$ acts as a memory for the model that explicitly maintains every previous layer's representation. Reading and writing correspond to attending and appending to the cache:
\begin{equation*}
\begin{aligned}
\mathread(\mathbf M_{l-1}, \mathbf q_{l-1}) &= {\operatorname{Attend}}(\mathbf M_{l-1}, \mathbf q_{l-1}) \\
\mathbf M_l = \mathwrite(\mathbf M_{l-1}, \mathbf k_l, \mathbf v_l) &= \mathbf M_{l-1}.{\operatorname{Append}}(\mathbf k_l, \mathbf v_l).
\end{aligned}
\end{equation*}
Since this mechanism keeps a copy of every representation, we refer to it as an \emph{explicit} depth mixer. Its memory stream grows over depth, as incoming representations are appended.\looseness-1

\textbf{mHC.}
For mHC, there again is exactly one stream. This time it is the expanded residual stream, $\mathbf S_l = (\mathbf X_l)$. The queries and keys are generated via $\mathbf q_l = \mathbf{H}_{l+1}^{\mathrm{pre}}(\mathbf{X}_l)$ and $\mathbf k_l = \mathbf{H}_l^{\mathrm{post}}(\mathbf{X}_{l-1})$. Reading and writing correspond to
\begin{equation*}
    \mathread{(\mathbf X_{l-1}, \mathbf q_{l-1})} = \mathbf X_{l-1} \mathbf q_{l-1}, \qquad \mathbf X_l = \mathwrite{(\mathbf X_{l-1}, \mathbf k_l, \mathbf v_l)} = \mathbf{X}_{l-1}\mathbf{H}_l^{\mathrm{res}}(\mathbf{X}_{l-1})+\mathbf{v}_l\mathbf k_l^\top.
\end{equation*}
The memory stream of mHC has a fixed size (see also Section~\ref{sec:background}). Compared to AttnRes, incoming representations are additively written into the stream, not appended. We thus refer to it as an \emph{implicit} depth mixer. Table~\ref{tab:depth-mixing} provides an overview of depth mixing architectures in our framework. In the next section, we introduce linear depth memory, an implicit depth mixer derived from first principles.

\subsection{Linear Depth Memory}
\label{sec:lDM}
In this section, we develop \emph{linear depth memory (LDM)}, a novel depth mixing module. At its core, LDM implements a linear associative memory~\citep{correlationmatrixmemories}. Such memories store key--value associations and lay the foundation for linear attention~\citep{pmlr-v139-schlag21a}. The simplest version of LDM maintains two streams, $\mathbf S_l = (\mathbf x_l, \mathbf M_l)$, where $\mathbf x_l$ is the standard PreLN residual stream and $\mathbf M_l$ is a matrix-valued hidden state. The read and write operations are
\begin{equation}
\begin{aligned}
\label{eq:dr-update-rule}
    \mathread((\mathbf x_{l-1}, \mathbf M_{l-1}), \mathbf{q}_{l-1}) &= \mathbf x_{l-1} + \mathbf M_{l-1} \mathbf q_{l-1}, \\
    (\mathbf x_l, \mathbf M_l) = \mathwrite((\mathbf x_{l-1}, \mathbf M_{l-1}), \mathbf k_l, \mathbf v_l) &= (\mathbf x_{l-1} + \mathbf v_l, \mathbf M_{l-1} + \mathbf v_l \mathbf k_l^\top).
\end{aligned}
\end{equation}
Note how, since simple LDM carries two streams, \textread{} and \textwrite{} need to incorporate both streams. One of the streams is the classic PreLN residual stream, while the other one maintains a matrix-valued memory stream via outer-product writes. Our ablations in Section~\ref{sec:ablations} reveal the importance of maintaining these two streams separately.

The memory stream is closely related to softmax depth attention (\cref{eq:attnres-background}). Indeed, the latter takes the form
\begin{equation}
\label{eq:softmax-attention}
    \mathbf o_{l-1}
    =
    \mathbf V_{<l}^{\top}
    \operatorname{softmax}
    \left(
        \mathbf K_{<l}\mathbf q_{l-1}
    \right).
\end{equation}
Following \citet{pmlr-v119-katharopoulos20a}, replacing the exponential kernel in \cref{eq:softmax-attention} with the linear dot-product kernel without normalization yields
\begin{equation*}
    \mathbf o_{l-1}
    =
    \mathbf V_{<l}^{\top}
    \mathbf K_{<l}
    \mathbf q_{l-1}
    =
    \left(
        \sum_{i=0}^{l-1}
        \mathbf v_i \mathbf k_i^\top
    \right)
    \mathbf q_{l-1}
    = 
    \mathbf M_{l-1}
    \mathbf q_{l-1},
\end{equation*}
where $\mathbf M_{l-1}=\sum_{i=0}^{l-1}\mathbf v_i\mathbf k_i^\top$ is exactly maintained by the LDM outer product recurrence in \cref{eq:dr-update-rule}. Thus, this simple instantiation of LDM implements linear attention over depth, replacing a growing collection of layer representations with a fixed-size matrix summary. The additive update rule approximates the solution of a regression problem~\citep{wang2025test}. We adapt this perspective for our main instantiation.\looseness-1

\textbf{An exact solution to depth regression.}
Following \citet{wang2025test}, we interpret the simple outer product update in \cref{eq:dr-update-rule} as an approximate solution to
\begin{equation}
\label{eq:least-squares}
    \mathbf{M}_l^* = \argmin_{\mathbf M_l} \sum_{i=0}^l \|\mathbf v_i - \mathbf M_l \mathbf k_i \|_2^2.
\end{equation}
We call this the \emph{depth regression} objective, by analogy with test-time regression along the sequence axis. If $\mathbf K_{\le l}^\top\mathbf K_{\le l}$ is invertible, the unique solution to~\cref{eq:least-squares} is $\mathbf M_l^* = \mathbf V_{\le l}^\top \mathbf K_{\le l} \left(\mathbf K_{\le l}^\top \mathbf K_{\le l}\right)^{-1}$. Approximating the key Gram matrix by the identity recovers the update in \cref{eq:dr-update-rule}.\looseness-1

For sequence mixing, efficient training and prefill motivate parallelizable (approximate) solvers of \cref{eq:least-squares}. Delta-rule-based models take an online gradient step at each token~\citep{pmlr-v139-schlag21a,yang2024parallelizing,yang2025gated,team2025kimi}. DeltaProduct takes multiple steps per token~\citep{siems2025deltaproductimprovingstatetrackinglinear}. MesaNet uses an iterative conjugate gradient solve~\citep{vonoswald2026mesanetsequencemodelinglocally}.\looseness-1

For depth mixing, however, transformer layers are already evaluated sequentially, allowing the use of non-parallelizable solvers. LDM maintains the exact regression solution through recursive least squares (RLS), as in \citet{vonoswald2024uncoveringmesaoptimizationalgorithmstransformers}. We apply RLS to a generalized version of \cref{eq:least-squares}, introducing learnable per-layer forgetting factors $\lambda_j>0$, dynamic importance weights $\omega_i>0$, and a weighted ridge penalty:
\begin{equation}
\label{eq:ridge-least-squares}
\mathbf M_l^* = \argmin_{\mathbf M}
\left[
\sum_{i=0}^l
\left(\prod_{j=i+1}^l\lambda_j\right)\omega_i
\|\mathbf v_i-\mathbf M\mathbf k_i\|_2^2
+
\left(\prod_{j=0}^l\lambda_j\right)\alpha^{-1}
\|\mathbf M\|_F^2
\right].
\end{equation}

\Cref{eq:ridge-least-squares} admits an exact recursive solution via Sherman--Morrison. Initialize $\mathbf M_{-1}=\mathbf 0$ and the inverse key covariance as $\mathbf P_{-1}=\alpha\mathbf I \in \mathbb R^{D_k \times D_k}$, and define the weighted observations \mbox{$\overline{\mathbf k}_l=\sqrt{\omega_l}\mathbf k_l$} and \mbox{$\overline{\mathbf v}_l=\sqrt{\omega_l}\mathbf v_l$.} The RLS gain and pseudo-value (the prediction error of $\mathbf M_{l-1}$) are
\begin{equation}
\label{eq:pseudo-updates}
\bm\kappa_l={\mathbf P_{l-1}\overline{\mathbf k}_l}/({\lambda_l+\overline{\mathbf k}_l^\top\mathbf P_{l-1}\overline{\mathbf k}_l}),
\qquad
\mathbf u_l=\overline{\mathbf v}_l-\mathbf M_{l-1}\overline{\mathbf k}_l.
\end{equation}
These can be used to solve~\cref{eq:ridge-least-squares} online,
\begin{equation}
\mathbf M_l=\mathbf M_{l-1}+\mathbf u_l\bm\kappa_l^\top,
\qquad
\mathbf P_l= ({\mathbf P_{l-1}-\bm\kappa_l\overline{\mathbf k}_l^\top\mathbf P_{l-1}})/{\lambda_l}.
\label{eq:rls-recursions}
\end{equation}
The complete derivation is given in Appendix~\ref{app:rls-derivation}. As shown, the RLS update introduces a third stream, an inverse covariance stream $\mathbf P_l$, to LDM. 

\begin{figure}[t]
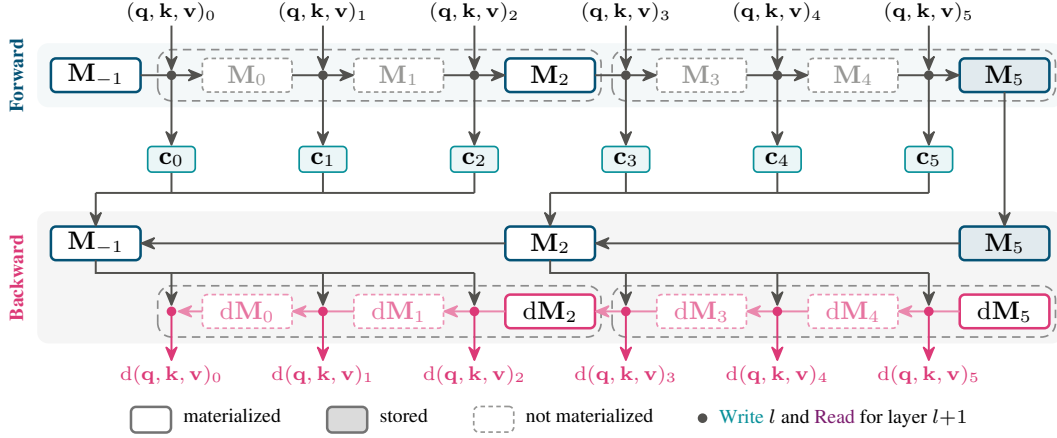

    \centering
    \resizebox{\linewidth}{!}{\ndmflows}
    \caption{{Efficient implementation of LDM} for $L=5$ layers and windows (denoted by the dashed boxes) of size $E=3$. Each dot denotes the \textwrite{} at layer $l$ fused with the \textread{} for layer $l+1$. In the \textcolor{flowstate}{forward} pass, each dot saves a small context $\mathbf c_l$, which is used for the \textcolor{flowadj}{backward} pass. Some details are omitted for clarity (see \cref{app:efficiency}).}
    \label{fig:hardware-aware-impl}
    \vspace{-0.5em}
\end{figure}

\textbf{Multiple groups.} We find that partitioning the memory state $\mathbf M_l$ into $G$ groups, each with value dimension $D_v = D/G$, improves expressivity and empirically improves performance slightly (Section~\ref{sec:ablations}). Each group maintains its own inverse covariance matrix and is queried independently. The resulting $D_v$-dimensional readouts are concatenated to form the full $D$-dimensional memory readout. The state then becomes $\mathbf S_l = (\mathbf x_l, \mathbf M_l, \mathbf P_l)$, where $\mathbf P_l \in \mathbb R^{G \times D_k \times D_k}$. The forgetting factors and gates are also applied and learned independently for each group (thus, in practice, we treat them as vectors, with each element corresponding to a group).

\textbf{Efficient parameterization through shared latent representations.} Having established LDM as an abstract architectural component above, we now present how to parametrize it efficiently. Let $\operatorname{Aff}$ denote a learned affine map. For the \textread{} operator, we generate a shared latent representation
\begin{equation*}
\mathbf{h}_{l-1}^{\textcolor{readcolor}{(R)}}
=
\operatorname{SiLU}\left(
\operatorname{Aff}(\,
\operatorname{Norm}_{\text{RMS}}(\mathbf{x}_{l-1}))
\right)
\in \mathbb{R}^{H}.
\end{equation*}
From this latent representation, we obtain the query as well as a per-group read gate
\begin{equation*}
\mathbf{q}_{l-1} =
\operatorname{Norm}_{L^2}(
\operatorname{Aff}(\mathbf{h}^{\textcolor{readcolor}{(R)}}_{l-1})) \in \mathbb{R}^{G\times D_k}, \quad \mathbf{g}_{l-1} =
2\sigma(
\operatorname{Aff}(\mathbf{h}^{{\textcolor{readcolor}{(R)}}}_{l-1})
)\in \mathbb{R}^{G}.
\end{equation*}
The gated query retrieves data from the streams as
\begin{equation*}
\mathread((\mathbf x_{l-1}, \mathbf{M}_{l-1}, \mathbf P_{l-1}),\mathbf{q}_{l-1})
=
\mathbf x_{l-1} + \boldsymbol{\gamma}_{l-1} \odot \mathbf{M}_{l-1} \left(\,
\mathbf{g}_{l-1}
\odot
 \mathbf{q}_{l-1} \right),
\end{equation*}
where $\boldsymbol{\gamma}_{l-1} \in \mathbb{R}^G$ is a learnable per-group gain that controls the magnitude of the read. For the \textwrite{} operator, we analogously first generate a latent representation
\begin{equation*}
\mathbf{h}^{{\textcolor{writecolor}{(W)}}}_{l}     
=
\operatorname{SiLU}\left(
\operatorname{Aff}\left(\,
\operatorname{Norm}_{\text{RMS}}(\mathbf{x}_{l})
\right)\right)
\in \mathbb{R}^{H},
\end{equation*}
from which we obtain the keys $\mathbf{k}_{l}$ and groupwise weight $\boldsymbol{\omega}_{l}$,
\begin{equation*}
\mathbf{k}_{l}
=
\operatorname{Norm}_{L^2}(
\operatorname{Aff}(\mathbf{h}^{{\textcolor{writecolor}{(W)}}}_{l}))
\in \mathbb{R}^{G\times D_k}, \qquad \boldsymbol{\omega}_{l}
=
2\sigma(
\operatorname{Aff}(\mathbf{h}^{{\textcolor{writecolor}{(W)}}}_{l})
)
\in \mathbb{R}^{G}.
\end{equation*}
The RLS write then computes
\begin{equation*}
(\mathbf x_l, \mathbf{M}_{l}, \mathbf P_l)
=
\mathwrite(
(\mathbf x_{l-1}, \mathbf{M}_{l-1}, \mathbf P_{l-1}),
\mathbf{k}_{l},
\mathbf v_{l}, \bm{\omega}_l, \bm{\lambda}_l
)
\end{equation*}
in the sense of \cref{eq:rls-recursions}. We additionally write the embedding into the first memory state and perform a read before the LM head (see \cref{app:boundary}).

\subsection{Hardware-Aware Implementation}
\label{sec:efficiency}
A naive implementation of LDM (1) stores a state $\mathbf M_l$ per layer for later reuse during backpropagation and (2) reads and writes the current state at every \textread{} and \textwrite{}, which incurs memory traffic. We propose an efficient implementation that reduces both activation memory overhead and traffic:\looseness-1

\textbf{(1) State unwinding.}
The RLS recurrence is reversible given compact per-layer auxiliary information. Indeed, \cref{eq:rls-recursions} shows that $\mathbf M_{l-1}$ can be reconstructed from $\mathbf M_l$ via
\begin{equation}\label{eq:unwinding}
\mathbf M_{l-1}=\mathbf M_l-\mathbf u_l\bm\kappa_l^{\top}.
\end{equation}
In particular, the backward pass can undo the state update given a context $\mathbf c_l$ saved for each layer, which holds $\mathbf u_l$ and $\bm\kappa_l$, the inverse covariance $\mathbf P_{l-1}$ before the update, and a few small vectors and scalars needed for the gradients, such as $\overline{\mathbf k}_l$ and $\boldsymbol\lambda_l$ (\cref{app:efficiency}). Since $\mathbf P_l \in \mathbb R^{G \times D_k \times D_k}$ and typically $D_k$ is small (see Section~\ref{sec:experiments}), storing $\mathbf c_l$ is much cheaper than storing $\mathbf M_l \in \mathbb R^{G \times D_v \times D_k}$. Our implementation of LDM updates the state $\mathbf M_l$ in place during the forward pass and unwinds the depth recurrence from the final state $\mathbf M_L$ during the backward pass (\cref{fig:hardware-aware-impl}).

\textbf{(2) Windowed writes.}
To reduce memory traffic, we fuse each \textwrite{} with the \textread{} for the next layer. We additionally defer the \textwrite{} operations within a window \mbox{$\mathcal W=\{b+1,\ldots,e\}$} of $E$ layers (let the state be $\mathbf M_b$ before the window) and compute every \textread{} and \textwrite{} from $\mathbf M_b$, since
\begin{equation}
\label{eq:served}
    \mathbf M_j \mathbf x = \mathbf M_b \mathbf x + \sum_{b<i\le j} \mathbf u_i (\bm{\kappa}_i^\top \mathbf x),
\end{equation}
for any $j \in \mathcal W$ and any vector $\mathbf x$ (e.g., the gated query of a \textread{} or the weighted key of the next \textwrite{}), with the pseudo-values $\mathbf u_i$ and RLS gains $\bm\kappa_i$ corresponding to \cref{eq:pseudo-updates}. No intermediate state is materialized inside a window, reducing \textwrite{} traffic. The window size $E$ trades state traffic against accesses to the pseudo-values. Our main experiments use $E=4$. In the backward pass, we similarly avoid materializing the unwound states and state gradients inside a window.

We implement LDM in CuTe DSL for Hopper and Blackwell GPUs. \Cref{fig:hardware-aware-impl} summarizes our implementation, and \cref{app:efficiency} provides further details.
\section{Experiments}
\label{sec:experiments}

\begin{figure}[!t]
    \centering
    \input{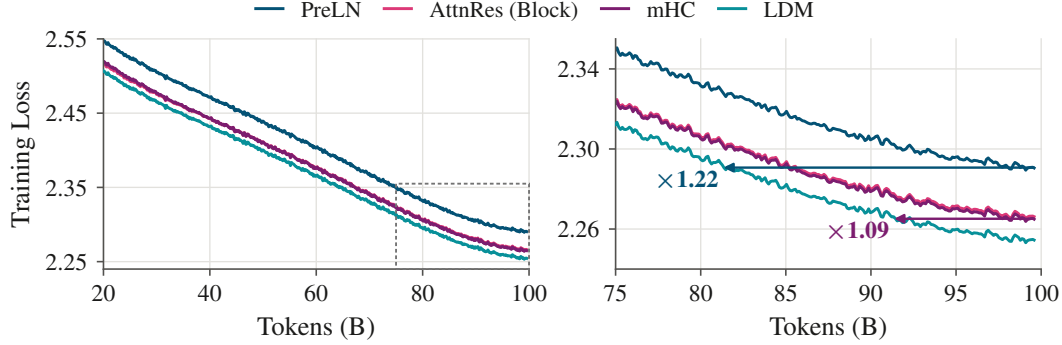}
    \caption{Left: Training loss at 1.5B model size (100B tokens, learning rate $10^{-3}$). Right: the last 25B tokens. Arrows: tokens LDM needs to reach each baseline's final loss. }
    \label{fig:main-loss}
    \vspace{-0.7em}
\end{figure}

In this section, we evaluate LDM by pretraining 400M Llama-like and 1.5B Qwen-like transformers on 50B and 100B FineWeb-Edu tokens, respectively~\citep{penedo2024finewebdatasetsdecantingweb}. Unless stated otherwise, the queries and keys are generated from hidden representations of dimension $H=32$ at 400M and $H=64$ at 1.5B. We set the per-group value dimension to the attention head dimension, yielding $G=16$ groups for both models. We furthermore fix $\alpha=1$, which we found to work better than making it learnable.

\subsection{Main Results}

\textbf{Setup.}
We compare LDM with PreLN, mHC, and AttnRes using matched backbone configurations. At both 400M and 1.5B, LDM adds approximately 1\% more parameters than PreLN, comparable to the overhead of mHC (\cref{tab:model-configs}). Our main LDM configuration uses $D_k=8$ at both scales, and we additionally evaluate $D_k=4$ at 400M. We train both Full and Block AttnRes at 400M, but only Block AttnRes at 1.5B because of the higher training cost of Full AttnRes. Following~\citet{kimiteam2026attentionresiduals}, we target approximately eight depth blocks for Block AttnRes, using eight at 400M and seven at 1.5B to avoid splitting transformer blocks across depth blocks. All models use AdamW with a weight decay of $0.1$ and linear warmup followed by cosine decay to $10\%$ of the peak learning rate. We sweep the peak learning rate for each method and scale using a grid spaced by factors of two, selecting the value with the lowest held-out validation loss. We report held-out validation, WikiText-2~\citep{merity2016pointersentinelmixturemodels}, and LAMBADA~\citep{paperno2016lambadadatasetwordprediction} perplexity, together with zero-shot accuracy on ARC-Easy, ARC-Challenge~\citep{clark2018thinksolvedquestionanswering}, HellaSwag~\citep{zellers2019hellaswagmachinereallyfinish}, PIQA~\citep{bisk2019piqareasoningphysicalcommonsense}, WinoGrande~\citep{sakaguchi2020winogrande}, OpenBookQA~\citep{mihaylov2018suitarmorconductelectricity}, and LAMBADA. Full implementation details are in \cref{app:exp_details}.

\begin{table*}[t!]
\centering
\small
\caption{Pretraining evaluations for both 400M and 1.5B models, generated using the lm-eval-harness~\citep{eval-harness}. Val is the pretraining validation perplexity.}
\label{tab:main_results}

\setlength{\tabcolsep}{2.8pt}
\renewcommand{\arraystretch}{1.08}

\begin{tabular}{clccccccccccc}
\toprule
& \multirow{2}{*}{Method} & ARC-E & ARC-C & Hella & PIQA & Wino & OBQA & Lmb & Avg & Lmb & Wiki & Val \\
& & $\mathrm{acc}\,\uparrow$ & $\mathrm{acc}_n\,\uparrow$
& $\mathrm{acc}_n\,\uparrow$ & $\mathrm{acc}\,\uparrow$
& $\mathrm{acc}\,\uparrow$ & $\mathrm{acc}_n\,\uparrow$
& $\mathrm{acc}\,\uparrow$ & $\mathrm{acc}\,\uparrow$
& $\mathrm{ppl}\,\downarrow$ & $\mathrm{ppl}\,\downarrow$ & $\mathrm{ppl}\,\downarrow$ \\
\midrule

\multirow{6}{*}{\rotatebox{90}{\tiny 400M @ 50B}}
& PreLN
& 65.66 & 31.23 & 46.80 & 68.93 & 55.25 & \underline{37.00} & 38.46 & 49.05 & 25.32 & 22.59 & 12.73 \\

& AttnRes (Full)
& 64.94 & 32.08 & 49.34 & \underline{70.08} & \underline{55.64} & 36.40 & 41.35 & 49.98 & 19.62 & 21.46 & 12.35 \\

& AttnRes (Block)
& \textbf{67.63} & \textbf{32.85} & \textbf{49.55} & \underline{70.08} & \textbf{56.04} & \textbf{37.20} & 40.91 & \textbf{50.61} & 20.71 & 20.86 & \underline{12.12} \\

& mHC
& 65.11 & 31.66 & 49.26 & 70.02 & 55.09 & 35.00 & 40.69 & 49.55 & 20.23 & 21.20 & 12.23 \\

& \cellcolor{gray!15}\raisebox{-0.20ex}[0pt][0pt]{LDM, $D_k{=}4$}
& \cellcolor{gray!15}\raisebox{-0.20ex}[0pt][0pt]{66.79}
& \cellcolor{gray!15}\raisebox{-0.20ex}[0pt][0pt]{\underline{32.34}}
& \cellcolor{gray!15}\raisebox{-0.20ex}[0pt][0pt]{48.71}
& \cellcolor{gray!15}\raisebox{-0.20ex}[0pt][0pt]{69.42}
& \cellcolor{gray!15}\raisebox{-0.20ex}[0pt][0pt]{54.70}
& \cellcolor{gray!15}\raisebox{-0.20ex}[0pt][0pt]{\textbf{37.20}}
& \cellcolor{gray!15}\raisebox{-0.20ex}[0pt][0pt]{\underline{42.27}}
& \cellcolor{gray!15}\raisebox{-0.20ex}[0pt][0pt]{50.20}
& \cellcolor{gray!15}\raisebox{-0.20ex}[0pt][0pt]{\underline{19.59}}
& \cellcolor{gray!15}\raisebox{-0.20ex}[0pt][0pt]{\underline{20.84}}
& \cellcolor{gray!15}\raisebox{-0.20ex}[0pt][0pt]{12.14} \\

& \cellcolor{gray!15}\raisebox{-0.20ex}[0pt][0pt]{\textbf{LDM (Ours)}}
& \cellcolor{gray!15}\raisebox{-0.20ex}[0pt][0pt]{\underline{67.13}}
& \cellcolor{gray!15}\raisebox{-0.20ex}[0pt][0pt]{\underline{32.34}}
& \cellcolor{gray!15}\raisebox{-0.20ex}[0pt][0pt]{\underline{49.41}}
& \cellcolor{gray!15}\raisebox{-0.20ex}[0pt][0pt]{\textbf{70.40}}
& \cellcolor{gray!15}\raisebox{-0.20ex}[0pt][0pt]{54.93}
& \cellcolor{gray!15}\raisebox{-0.20ex}[0pt][0pt]{36.40}
& \cellcolor{gray!15}\raisebox{-0.20ex}[0pt][0pt]{\textbf{42.58}}
& \cellcolor{gray!15}\raisebox{-0.20ex}[0pt][0pt]{\underline{50.46}}
& \cellcolor{gray!15}\raisebox{-0.20ex}[0pt][0pt]{\textbf{19.10}}
& \cellcolor{gray!15}\raisebox{-0.20ex}[0pt][0pt]{\textbf{20.70}}
& \cellcolor{gray!15}\raisebox{-0.20ex}[0pt][0pt]{\textbf{12.08}} \\

\midrule

\multirow{4}{*}{\rotatebox{90}{\tiny 1.5B @ 100B}}
& PreLN
& 74.07 & 41.47 & 62.04 & \underline{75.14} & 62.35 & 42.60 & 51.19 & 58.41 & 10.49 & 15.06 & 9.48 \\

& AttnRes (Block)
& \underline{75.88} & \textbf{42.92} & \underline{63.51} & 75.03 & 60.77 & \underline{43.60} & \underline{54.22} & 59.42 & \underline{8.84} & 14.54 & 9.26 \\

& mHC
& 75.67 & 41.72 & 63.32 & \textbf{75.57} & \textbf{62.98} & \textbf{44.40} & 54.08 & \underline{59.68} & 9.05 & \underline{14.44} & \underline{9.25} \\

& \cellcolor{gray!15}\raisebox{-0.20ex}[0pt][0pt]{\textbf{LDM (Ours)}}
& \cellcolor{gray!15}\raisebox{-0.20ex}[0pt][0pt]{\textbf{76.35}}
& \cellcolor{gray!15}\raisebox{-0.20ex}[0pt][0pt]{\underline{42.66}}
& \cellcolor{gray!15}\raisebox{-0.20ex}[0pt][0pt]{\textbf{63.98}}
& \cellcolor{gray!15}\raisebox{-0.20ex}[0pt][0pt]{\underline{75.14}}
& \cellcolor{gray!15}\raisebox{-0.20ex}[0pt][0pt]{\underline{62.83}}
& \cellcolor{gray!15}\raisebox{-0.20ex}[0pt][0pt]{42.00}
& \cellcolor{gray!15}\raisebox{-0.20ex}[0pt][0pt]{\textbf{55.00}}
& \cellcolor{gray!15}\raisebox{-0.20ex}[0pt][0pt]{\textbf{59.71}}
& \cellcolor{gray!15}\raisebox{-0.20ex}[0pt][0pt]{\textbf{8.42}}
& \cellcolor{gray!15}\raisebox{-0.20ex}[0pt][0pt]{\textbf{14.21}}
& \cellcolor{gray!15}\raisebox{-0.20ex}[0pt][0pt]{\textbf{9.15}} \\

\bottomrule
\end{tabular}
\vspace{-0.7em}
\end{table*}

\textbf{Lower perplexity and competitive accuracy.} \Cref{tab:main_results} reports our main pretraining results. It reveals that our main LDM configuration ($D_k=8$) achieves the lowest validation, WikiText-2, and LAMBADA perplexity at both model sizes. It also achieves the highest average zero-shot accuracy at 1.5B and ranks second at 400M. \Cref{fig:main-loss} shows the training loss at 1.5B.\looseness-1

\textbf{Matched memory matrix size.} Both mHC and LDM maintain states of fixed size across depth, whereas Full AttnRes retains a growing cache of layer outputs. With $D_k=8$, LDM's memory matrix contains $8D$ entries per token, twice the $4D$ entries in mHC's expanded residual state. We therefore also evaluate LDM with $D_k=4$ at 400M, matching the number of entries in its memory matrix to mHC's expanded residual state (note that LDM additionally maintains a classic residual stream and an inverse covariance stream). This smaller configuration still achieves lower validation, WikiText-2, and LAMBADA perplexity than mHC, alongside higher average zero-shot accuracy. Thus, LDM remains competitive with other baselines even with a smaller memory matrix.

\begin{figure*}[t]
\centering
\input{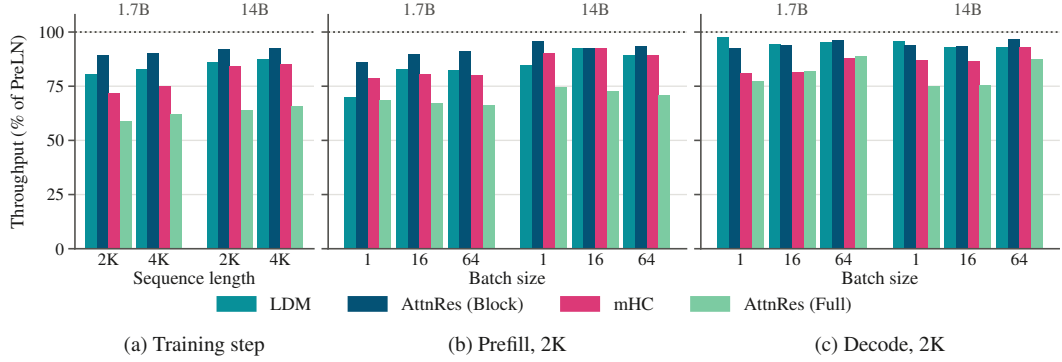}
\caption{Throughput relative to PreLN on B200 for Qwen-like 1.7B and 14B models, comparing LDM, mHC and AttnRes. The mHC results use an improved version of the official mHC kernels. (a) Training step with 2K and 4K sequence length. (b) Prefill of 2K prompts at different batch sizes. (c) KV-cached decode after 2K prefill at different batch sizes. More settings in \cref{app:efficiency-results}.}
\label{fig:efficiency-results}
\end{figure*}

\textbf{Throughput.} \Cref{fig:efficiency-results} compares LDM's training, prefill and decode throughput with AttnRes, implemented using Flash Linear Attention kernels~\citep{yang2024fla}, and mHC using an improved version of the official mHC kernels. More details and settings can be found in \cref{app:efficiency-results}. Across 1.7B--14B models, sequence lengths, context lengths, and batch sizes, LDM exceeds Full AttnRes throughput in every setting and mHC in all training and most inference settings, while remaining competitive with Block AttnRes, especially during decode.

\subsection{Ablations and Architectural Choices}
\label{sec:ablations}

\begin{table}[t]
\centering
\caption{LDM ablations, 400M @ 15B. $\Delta$ denotes change in held-out perplexity ($\downarrow$) relative to our main configuration. (a) $F(\mathbf r)$: memory-only layer input (normalization omitted). $\mathbf x=\boldsymbol\delta$: no post-layer residual addition. (b) Delta/Outer: replace RLS with delta rule/outer product update. (c) $\mathbf v=\mathbf x$: store the residual stream. $\mathbf{QK}(\boldsymbol\delta)$: queries and keys from layer outputs. Static $\mathbf q$: learned query per read with read gate fixed to one. (d) $-\boldsymbol\gamma$, $\mathbf g=\mathbf 1$, $\boldsymbol\omega=\mathbf 1$: fix read gain, read gate, or write weight to one, respectively. $\boldsymbol\lambda=\mathbf 1$: disable forgetting. Selected training curves in \cref{fig:ablations}, full definitions and results in \cref{app:ablation-definitions,tab:ablation-validation-full}.}
\label{tab:ablation-validation}
\footnotesize
\setlength{\tabcolsep}{1.5pt}
\renewcommand{\arraystretch}{1.1}
\begin{tabular*}{\textwidth}{@{\extracolsep{\fill}}l*{13}{c}@{}}
\toprule
 & \multicolumn{2}{c}{Baselines} & \multicolumn{2}{c}{(a) Residual} & \multicolumn{2}{c}{(b) Update} & \multicolumn{3}{c}{(c) Addressing} & \multicolumn{4}{c}{(d) Gain, gates, forgetting} \\
\cmidrule(lr){2-3}\cmidrule(lr){4-5}\cmidrule(lr){6-7}\cmidrule(lr){8-10}\cmidrule(l){11-14}
 & PreLN & \textbf{LDM} & $F(\mathbf r)$ & $\mathbf x=\boldsymbol\delta$ & Delta & Outer & $\mathbf v=\mathbf x$ & $\mathbf{QK}(\boldsymbol\delta)$ & Static $\mathbf q$ & $-\boldsymbol\gamma$ & $\mathbf g=\mathbf 1$ & $\boldsymbol\omega=\mathbf 1$ & $\boldsymbol\lambda=\mathbf 1$ \\
\midrule
$\Delta$ & 0.78 & 0.00 & 1.88 & 0.87 & 0.06 & 0.07 & 0.33 & 0.07 & 0.19 & 0.04 & 0.07 & 0.01 & $-$0.01 \\
\bottomrule
\end{tabular*}
\vspace{-0.5em}
\end{table}

We ablate LDM's main design choices at 400M, training each variant on 15B FineWeb-Edu tokens with a peak learning rate of $10^{-3}$ and otherwise following our main training recipe. \Cref{tab:ablation-validation} reports changes in held-out perplexity relative to the default LDM configuration. \Cref{app:ablation-definitions} provides full definitions of each variant. Our main findings are as follows:

\textbf{(a) Keeping the residual stream is crucial.}
We assess the importance of LDM’s separate residual stream by interrupting its flow at one of two points: before the layer, so that the layer receives only the memory readout, or after the layer, so that the stream is replaced by the layer output. Both variants underperform PreLN.

\textbf{(b) Depth regression approximations.} Replacing RLS with either a plain outer product or a delta rule update~\citep{pmlr-v139-schlag21a,yang2024parallelizing} still yields better performance than PreLN, but neither alternative matches RLS. This result is consistent with reported benefits of more accurate test-time regression solvers in sequence modeling~\citep{vonoswald2026mesanetsequencemodelinglocally,tumma2026preconditioneddeltanetcurvatureawaresequence}.\looseness-1

\begin{wrapfigure}[12]{r}{0.34\textwidth}
\centering
\input{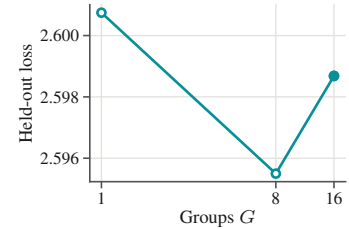}
\caption{Held-out loss versus group count (400M, 15B tokens).}
\label{fig:group-sweep}
\vspace{-5em}
\end{wrapfigure}
\textbf{(c) Addressing and stored values.}
We next examine how LDM's performance depends on the choice of stored representations, the construction of queries and keys, and the number of memory groups. Storing $\mathbf x_l$ instead of $\boldsymbol\delta_l$ hurts most, potentially due to redundancy in the stored information (given that earlier representations would be stored at each \textwrite{}). Deriving queries and keys from $\boldsymbol\delta_l$ instead of $\mathbf x_l$ also hurts, and static per-layer queries (read gate fixed to one) hurt more. Grouping provides small but consistent gains (\cref{fig:group-sweep}).

\textbf{(d) Gain, gates and forgetting.} Disabling either the read gain $\boldsymbol\gamma$ or the read gate $\mathbf g$ hurts slightly. Fixing the write weight $\bm{\omega}_l$ to one or disabling forgetting ($\boldsymbol\lambda_l=\mathbf 1$) has a negligible effect in this setting.

\begin{figure}[t]
    \centering
    \input{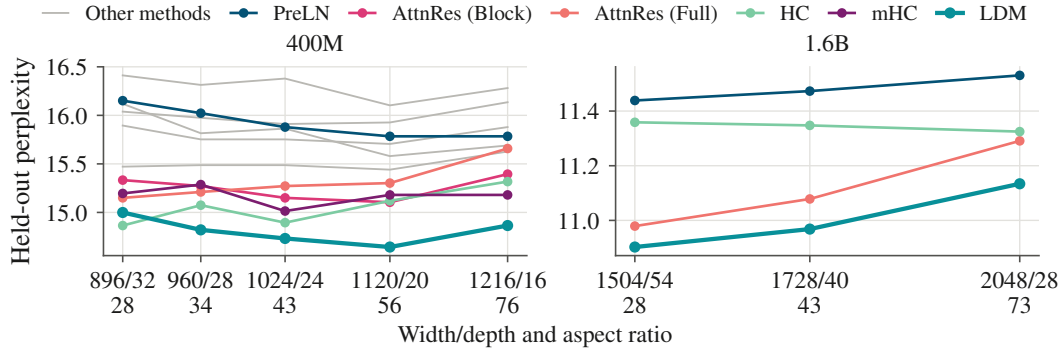}
    \caption{Final held-out perplexity on DepthBench against the aspect ratio (width/depth) at a fixed budget of 400M (left) and 1.6B (right) parameters. Tick labels denote the width/depth of each shape and its aspect ratio.}
    \label{fig:depthbench}
    \vspace{-0.7em}
\end{figure}

\subsection{Different Aspect Ratios}
\label{sec:depthbench}
Following the protocol of DepthBench~\citep{wang2026depthbenchmeasuringresidualconnections}, we evaluate whether LDM's gains persist across different model aspect ratios (width/depth) at fixed parameter budgets of 400M and 1.6B (for the latter, DepthBench provides results for their most promising baselines). We compare against the published DepthBench baseline results, matching their training setup exactly to ensure comparability. At 400M, DepthBench tunes the learning rate separately for each baseline and finds $2\times10^{-3}$ to be optimal in most cases. At 1.6B, it uses $5\times10^{-4}$ for all methods. We adopt these learning rates for LDM without further tuning. Additional experimental details are provided in \cref{app:depthbench}.

\textbf{Results and shape preferences.} LDM improves over PreLN at every evaluated aspect ratio at both scales (\cref{fig:depthbench}). At 400M, LDM favors moderate aspect ratios and outperforms the baselines, except in the deepest configuration, where HC performs slightly better. At 1.6B, however, LDM achieves the lowest perplexity at every considered aspect ratio. In particular, HC, which becomes less stable as scale increases~\citep{wang2026depthbenchmeasuringresidualconnections}, no longer improves with depth, whereas LDM's perplexity decreases monotonically. \Cref{app:depthbench} extends our 400M sweep to extreme aspect ratios, further supporting LDM's preference for shallower configurations at that scale.
\section{Discussion and Conclusion}
In this work, we introduce Depth Memory, a framework that interprets the propagation of information over depth as a problem of managing layer representations using a set of streams. This perspective unifies standard residual connections, AttnRes, and mHC, while distinguishing methods that retain explicit layer representations from those that mix them into a fixed-size state. It also provides a starting point for designing new depth mixers. Building on this view, we then propose Linear Depth Memory (LDM), which formulates memory updates as a depth regression problem, solving it exactly using recursive least squares. To do so, LDM maintains a standard residual stream, a memory stream, and an inverse covariance stream. Each token thus carries a compressed, queryable representation of earlier layer outputs, which enriches the inputs to subsequent layers. Our implementation uses state unwinding and windowed updates to reduce activation storage and memory traffic, making LDM feasible in practice. At both 400M and 1.5B, trained on 50B and 100B tokens, respectively, our main LDM configuration achieves the lowest reported perplexities among the tuned PreLN, AttnRes, and mHC baselines, alongside competitive zero-shot accuracy. These gains hold across standard aspect ratios. 

\textbf{Limitations and future work.} In this work, we focus on language models with up to 1.6B parameters that are trained on up to 100B tokens. Future work could evaluate LDM at larger scales and in other domains such as image classification. Moreover, LDM can add peak memory during training, which other stable formulations of the backward recurrence or checkpointing (as used in mHC) could reduce. Finally, our Depth Memory framework admits a plethora of different instantiations, such as alternative linear recurrences or memories with nonlinear read and write operations. We consider the exploration of that design space a promising avenue for future work.

\subsection*{Acknowledgments}
This work was supported in part by the Hector Foundation and by the Max Planck ETH Center for Learning Systems.
We thank the authors of DepthBench \citep{wang2026depthbenchmeasuringresidualconnections} for providing early access to their codebase, on which ours is based.

\bibliography{iclr2027_conference}

@misc{penedo2024finewebdatasetsdecantingweb,
      title={The FineWeb Datasets: Decanting the Web for the Finest Text Data at Scale}, 
      author={Guilherme Penedo and Hynek Kydlíček and Loubna Ben allal and Anton Lozhkov and Margaret Mitchell and Colin Raffel and Leandro Von Werra and Thomas Wolf},
      year={2024},
      eprint={2406.17557},
      archivePrefix={arXiv},
      primaryClass={cs.CL},
      url={https://arxiv.org/abs/2406.17557}, 
}

@inproceedings{
zhu2025hyperconnections,
title={Hyper-Connections},
author={Defa Zhu and Hongzhi Huang and Zihao Huang and Yutao Zeng and Yunyao Mao and Banggu Wu and Qiyang Min and Xun Zhou},
booktitle={The Thirteenth International Conference on Learning Representations},
year={2025},
url={https://openreview.net/forum?id=9FqARW7dwB}
}

@misc{li2026siamesenormbreakingbarrierreconciling,
      title={SiameseNorm: Breaking the Barrier to Reconciling Pre/Post-Norm}, 
      author={Tianyu Li and Dongchen Han and Zixuan Cao and Haofeng Huang and Mengyu Zhou and Ming Chen and Erchao Zhao and Xiaoxi Jiang and Guanjun Jiang and Gao Huang},
      year={2026},
      eprint={2602.08064},
      archivePrefix={arXiv},
      primaryClass={cs.LG},
      url={https://arxiv.org/abs/2602.08064}, 
}

@misc{mak2025residualmatrixtransformersscaling,
      title={Residual Matrix Transformers: Scaling the Size of the Residual Stream}, 
      author={Brian Mak and Jeffrey Flanigan},
      year={2025},
      eprint={2506.22696},
      archivePrefix={arXiv},
      primaryClass={cs.LG},
      url={https://arxiv.org/abs/2506.22696}, 
}

@misc{fang2023crosslayerretrospectiveretrievinglayer,
      title={Cross-Layer Retrospective Retrieving via Layer Attention}, 
      author={Yanwen Fang and Yuxi Cai and Jintai Chen and Jingyu Zhao and Guangjian Tian and Guodong Li},
      year={2023},
      eprint={2302.03985},
      archivePrefix={arXiv},
      primaryClass={cs.CV},
      url={https://arxiv.org/abs/2302.03985}, 
}

@misc{yang2026mhclitedontneed20,
      title={mHC-lite: You Don't Need 20 Sinkhorn-Knopp Iterations}, 
      author={Yongyi Yang and Jianyang Gao},
      year={2026},
      eprint={2601.05732},
      archivePrefix={arXiv},
      primaryClass={cs.LG},
      url={https://arxiv.org/abs/2601.05732}, 
}

@inproceedings{
xie2026mhcmanifoldconstrainedhyperconnections,
title={m{HC}: Manifold-Constrained Hyper-Connections},
author={Zhenda Xie and Yixuan Wei and Huanqi Cao and Chenggang Zhao and Chengqi Deng and Jiashi Li and Damai Dai and Huazuo Gao and Mingyu Xu and Kuai Yu and Liang Zhao and Shangyan Zhou and Zhean Xu and Zhengyan Zhang and Wangding Zeng and Shengding Hu and Yuqing Wang and Jingyang Yuan and Lean Wang and Wenfeng Liang},
booktitle={Forty-third International Conference on Machine Learning},
year={2026},
url={https://openreview.net/forum?id=mDhyxu8WRb}
}

@misc{zhang2026deepdeltalearning,
      title={Deep Delta Learning}, 
      author={Yifan Zhang and Yifeng Liu and Mengdi Wang and Quanquan Gu},
      year={2026},
      eprint={2601.00417},
      archivePrefix={arXiv},
      primaryClass={cs.LG},
      url={https://arxiv.org/abs/2601.00417}, 
}

@misc{vonoswald2026mesanetsequencemodelinglocally,
      title={MesaNet: Sequence Modeling by Locally Optimal Test-Time Training}, 
      author={Johannes von Oswald and Nino Scherrer and Seijin Kobayashi and Luca Versari and Songlin Yang and Sarthak Mittal and Maximilian Schlegel and Kaitlin Maile and Yanick Schimpf and Oliver Sieberling and Alexander Meulemans and Rif A. Saurous and Guillaume Lajoie and Charlotte Frenkel and Razvan Pascanu and Blaise Agüera y Arcas and João Sacramento},
      year={2026},
      eprint={2506.05233},
      archivePrefix={arXiv},
      primaryClass={cs.LG},
      url={https://arxiv.org/abs/2506.05233}, 
}

@misc{gomez2017reversibleresidualnetworkbackpropagation,
      title={The Reversible Residual Network: Backpropagation Without Storing Activations}, 
      author={Aidan N. Gomez and Mengye Ren and Raquel Urtasun and Roger B. Grosse},
      year={2017},
      eprint={1707.04585},
      archivePrefix={arXiv},
      primaryClass={cs.CV},
      url={https://arxiv.org/abs/1707.04585}, 
}

@misc{olmo20242olmo2furious,
      title={{2 OLMo 2 Furious}},
      author={{Team OLMo} and Pete Walsh and Luca Soldaini and Dirk Groeneveld and Kyle Lo and Shane Arora and Akshita Bhagia and Yuling Gu and Shengyi Huang and Matt Jordan and Nathan Lambert and Dustin Schwenk and Oyvind Tafjord and Taira Anderson and David Atkinson and Faeze Brahman and Christopher Clark and Pradeep Dasigi and Nouha Dziri and Michal Guerquin and Hamish Ivison and Pang Wei Koh and Jiacheng Liu and Saumya Malik and William Merrill and Lester James V. Miranda and Jacob Morrison and Tyler Murray and Crystal Nam and Valentina Pyatkin and Aman Rangapur and Michael Schmitz and Sam Skjonsberg and David Wadden and Christopher Wilhelm and Michael Wilson and Luke Zettlemoyer and Ali Farhadi and Noah A. Smith and Hannaneh Hajishirzi},
      year={2024},
      eprint={2501.00656},
      archivePrefix={arXiv},
      primaryClass={cs.CL},
      url={https://arxiv.org/abs/2501.00656},
}

@misc{zhang2026xhcexpandedhyperconnections,
      title={{xHC}: Expanded Hyper-Connections}, 
      author={Xiangdong Zhang and Xiaohan Qin and Sunan Zou and Tuo Dai and Xiaoming Shi and Huaijin Wu and Yebin Yang and Zhuo Xia and Shaofeng Zhang and Lin Yao and Yuliang Liu and Yu Cheng and Junchi Yan},
      year={2026},
      eprint={2607.14530},
      archivePrefix={arXiv},
      primaryClass={cs.LG},
      url={https://arxiv.org/abs/2607.14530}, 
}

@misc{vonoswald2024uncoveringmesaoptimizationalgorithmstransformers,
      title={Uncovering mesa-optimization algorithms in Transformers}, 
      author={Johannes von Oswald and Maximilian Schlegel and Alexander Meulemans and Seijin Kobayashi and Eyvind Niklasson and Nicolas Zucchet and Nino Scherrer and Nolan Miller and Mark Sandler and Blaise Agüera y Arcas and Max Vladymyrov and Razvan Pascanu and João Sacramento},
      year={2024},
      eprint={2309.05858},
      archivePrefix={arXiv},
      primaryClass={cs.LG},
      url={https://arxiv.org/abs/2309.05858}, 
}

@inproceedings{
xiao2025muddformer,
title={{MUDDF}ormer: Breaking Residual Bottlenecks in Transformers via Multiway Dynamic Dense Connections},
author={Da Xiao and Qingye Meng and Shengping Li and Xingyuan Yuan},
booktitle={Forty-second International Conference on Machine Learning},
year={2025},
url={https://openreview.net/forum?id=qkhgzNiEdj}
}

@misc{kimiteam2026kimik3openfrontier,
      title={Kimi K3: Open Frontier Intelligence}, 
      author={{Kimi Team} and Tongtong Bai and Yifan Bai and Yiping Bao and M. C. and Jianfeng Cai and Xinyuan Cai and Peizhou Cao and Yuxuan Cao and Ziwei Chai and Y. Charles and H. S. Che and Guanduo Chen and Guangyu Chen and Guanzheng Chen and Huarong Chen and Jia Chen and Jianlong Chen and Jun Chen and Kexin Chen and Peng Chen and Ruijue Chen and Wentao Chen and Xin Chen and Yang Chen and Yanru Chen and Yifei Chen and Yingjiang Chen and Yuankun Chen and Yujie Chen and Yutian Chen and Zhirong Chen and Dazhi Cheng and Yean Cheng and Jialei Cui and Jingbing Cui and Anqi Dai and Jiaqi Deng and Hao Ding and Rui Ding and Shaofeng Ding and Mengfan Dong and Mengnan Dong and Yuhao Dong and Yuxin Dong and Angang Du and Chenzhuang Du and Dikang Du and Jusen Du and Yulun Du and Yu Fan and Jing Feng and Qiulin Feng and Yichen Feng and Kelin Fu and Qiang Fu and Fuxuan Gao and Hongcheng Gao and Jingyue Gao and Tong Gao and Weijia Gao and Shangyi Geng and Jie Gong and Linhu Gong and Shengao Gong and Xiaochen Gong and Qizheng Gu and Yicheng Gu and Shuhao Guan and Haiqing Guo and Shiqi Guo and Xiang Guo and Zhengyan Guo and Beixi Hao and Wenxin Hao and Xiaoru Hao and Dailan He and Haotian He and Lehan He and Qi He and Weiran He and Xinran He and Xinyi He and Yibo He and Yunjia He and Chao Hong and Tiange Hong and Hao Hu and Jiaxi Hu and Ruikun Hu and Weiming Hu and Yangyang Hu and Zhenxing Hu and Liang Hua and Jinbin Huang and Ke Huang and Ruiyuan Huang and Siying Huang and Weixiao Huang and Yan Huang and Zhengjie Huang and Zhiqi Huang and Yulong Hui and Chaobo Jia and Yutong Jiang and Zhejun Jiang and Zuoyou Jiang and Wenyi Jin and Xinyi Jin and Yu Jing and Huanjun Kong and Guokun Lai and Aidi Li and Cheng Li and Chengyuan Li and Cong Li and Fang Li and Guanyu Li and Haoyang Li and Jia Li and Junxiong Li and Lei Li and Letian Li and Lincan Li and Weihong Li and Wentao Li and Xintong Li and Yang Li and Yishen Li and Yiwei Li and Yuxiao Li and Zhaowei Li and Zhaoxi Li and Zheming Li and Zhengxiao Li and Zhiyuan Li and Jiawei Lin and Xiaohan Lin and Yibo Lin and Zichao Lin and Ziyan Lin and Bill Liu and Boxiao Liu and Chuan Liu and Liang Liu and Shaowei Liu and Shudong Liu and Shuran Liu and Tianwei Liu and Weizhou Liu and Yangyang Liu and Yanming Liu and Yibo Liu and Yipeng Liu and Zhengying Liu and Zhiheng Liu and Enzhe Lu and Haoyu Lu and Linqiang Lu and Tingzhan Lu and Zhiyuan Lu and Aotian Luo and G. Luo and Junyu Luo and Yifan Luo and B. Lyu and Wenzhou Lyu and Shaoguang Mao and Yuan Mei and Xin Men and Minqing Ni and Yixuan Niu and Siyuan Pan and Shujun Peng and Zhangyang Qi and Ruoyu Qin and ZeChao Qin and Zeyu Qin and Haiquan Qiu and Jianxin Qiu and Jiezhong Qiu and Bowen Qu and Yuhao Qu and Zeyu Shang and Youbo Shao and Han Shen and Jincheng Shi and Juanfeng Shi and Lidong Shi and Shengyuan Shi and Wingchun Siu and Pengwei Song and Xiaoxi Song and Jianlin Su and Yunfeng Su and Zhaochen Su and Lin Sui and Jingsong Sun and Junyao Sun and Shaoning Sun and Shuzhe Sun and Tongyu Sun and Yujun Sun and Yunpeng Tai and Chuning Tang and Heyi Tang and Sirui Tang and Zecheng Tang and Chaoran Tian and Rongpeng Tian and Yu Tian and Wei Tu and Chensi Wang and Chuang Wang and Chunjie Wang and Dinglu Wang and Feng Wang and Hailong Wang and Haiming Wang and Hao Wang and Hao Wang and Huaqing Wang and Hui Wang and Jiayi Wang and Jinglong Wang and Jinhong Wang and Jiuzheng Wang and Linian Wang and Shaobo Wang and Shenzhi Wang and Shuyi Wang and Si Wang and Siyuan Wang and Tianfu Wang and Wenjue Wang and Xingran Wang and Xinmei Wang and Xinyuan Wang and Xusheng Wang and Yalin Wang and Yangkun Wang and Yao Wang and Yaoyu Wang and Yejie Wang and Yiqin Wang and Yucheng Wang and Yuzhi Wang and Zhaoji Wang and Zhaowei Wang and Zhengtao Wang and Zhenhao Wang and Zhongsheng Wang and Zifan Wang and Chu Wei and Ming Wei and Shouxin Wei and Zichen Wen and Fan Wu and Haoning Wu and Rucong Wu and Wenhao Wu and Xiaoxue Wu and Yingcong Wu and Yongqi Wu and Yuxin Wu and Zijian Wu and Xinglang Xian and Chenxuan Xiang and Yuye Xiang and Bocheng Xiao and Chenjun Xiao and Xin Xiao and Jin Xie and Xiaotong Xie and Yifeng Xie and Zhe Xie and Bowei Xing and Yiming Xiong and Baosheng Xu and Boyu Xu and Jiale Xu and Jianfan Xu and Jing Xu and Jinjing Xu and L. H. Xu and Qingtao Xu and Shuyao Xu and Suting Xu and Tiantian Xu and Tianxiang Xu and Weixin Xu and Xinran Xu and Yangchuan Xu and Ye Xu and Yueni Xu and Ziyao Xu and Haonan Xue and Junjie Yan and Yaoyao Yan and Fan Yang and Guangyao Yang and Hao Yang and Junwei Yang and Ruoyu Yang and Wenjie Yang and Xiaofei Yang and Xinyu Yang and Yi Yang and Yiling Yang and Ying Yang and Yuchen Yang and Zhen Yang and Zhilin Yang and Zian Yang and Zuhao Yang and Haotian Yao and Dan Ye and Haoran Ye and Wenjie Ye and Zhanbo Ye and Bohong Yin and Haoxiang Yin and Xietong Yin and Chengzhen Yu and Haozhen Yu and Longhui Yu and Shengnan Yu and Shuying Yu and Tianxiang Yu and Enming Yuan and Mengjie Yuan and Tongtian Yue and Wei Yue and Yang Yue and Dunyuan Zha and Haobing Zhan and B. H. Zhang and Dehao Zhang and Fei Zhang and Hao Zhang and Haoyuan Zhang and Huanyu Zhang and Jiapei Zhang and Jiaxuan Zhang and Jin Zhang and Kaiyi Zhang and Miaozhen Zhang and Puqi Zhang and Qinglei Zhang and Rong Zhang and Rui Zhang and Shaoshuai Zhang and Shiyi Zhang and Xiaobin Zhang and Xiaoyun Zhang and Y. Zhang and Yangkun Zhang and Ye Zhang and Yichi Zhang and Yikun Zhang and Yizhi Zhang and Yongting Zhang and Yu Zhang and Yutao Zhang and Yutong Zhang and Zheng Zhang and Zijing Zhang and Bin Zhao and Chenguang Zhao and Feifan Zhao and Jinglun Zhao and Jinxiang Zhao and Shuai Zhao and Wenshuo Zhao and Xiangyu Zhao and Xuanle Zhao and Yikai Zhao and Zijia Zhao and Haozhi Zheng and Huabin Zheng and Ruihan Zheng and Shaojie Zheng and Tengyang Zheng and Haofeng Zhong and Lei Zhong and Longguang Zhong and M. Zhou and Qiankang Zhou and Runjie Zhou and Ruozhang Zhou and Xinyu Zhou and Yiqiao Zhou and Zaida Zhou and Jinguo Zhu and Liya Zhu and Xinhao Zhu and Yangjunfeng Zhu and Yuxuan Zhu and Zhen Zhu and Chen Zhuang and Weiyu Zhuang and Xinxing Zu},
      year={2026},
      eprint={2607.24653},
      archivePrefix={arXiv},
      primaryClass={cs.CL},
      url={https://arxiv.org/abs/2607.24653}, 
}

@misc{qiu2026designqwen38nextarchitectureevaluation,
      title={On the Design of Qwen3.8-Next Architecture: Evaluation, Efficiency, and Training Stability}, 
      author={Zihan Qiu and Zekun Wang and Xiao Li and Yanpeng Li and Yang Xu and Yixuan Wang and Huaqing Zhang and Rui Men and Bochao Mao and Chengruidong Zhang and Fan Zhou and Hao Luo and Haofeng Huang and Haoran Lian and Haoyan Huang and Hongqing Chen and Jianwei Zhang and Jing Xu and Junjie Wang and Langshi Chen and Liangyu Wang and Linlang Jiang and Man Yuan and Minmin Sun and Peng Jin and Siqi Zhang and Siyu Wang and Xingzhang Ren and Yakai Wang and Yi Zhang and Yiming Dong and Yizhong Cao and Yubo Ma and Yunfei Mao and Bo Zheng and Dayiheng Liu},
      year={2026},
      eprint={2608.30320},
      archivePrefix={arXiv},
      primaryClass={cs.CL},
      url={https://arxiv.org/abs/2608.30320}, 
}

@misc{kimiteam2026attentionresiduals,
      title={Attention Residuals}, 
      author={{Kimi Team} and Guangyu Chen and Yu Zhang and Jianlin Su and Weixin Xu and Siyuan Pan and Yaoyu Wang and Yucheng Wang and Guanduo Chen and Bohong Yin and Yutian Chen and Junjie Yan and Ming Wei and Y. Zhang and Fanqing Meng and Chao Hong and Xiaotong Xie and Shaowei Liu and Enzhe Lu and Yunpeng Tai and Yanru Chen and Xin Men and Haiqing Guo and Y. Charles and Haoyu Lu and Lin Sui and Jinguo Zhu and Zaida Zhou and Weiran He and Weixiao Huang and Xinran Xu and Yuzhi Wang and Guokun Lai and Yulun Du and Yuxin Wu and Zhilin Yang and Xinyu Zhou},
      year={2026},
      eprint={2603.15031},
      archivePrefix={arXiv},
      primaryClass={cs.CL},
      url={https://arxiv.org/abs/2603.15031}, 
}

@inproceedings{
heddes2025deepcrossattention,
title={DeepCrossAttention: Supercharging Transformer Residual Connections},
author={Mike Heddes and Adel Javanmard and Kyriakos Axiotis and Gang Fu and Mohammadhossein Bateni and Vahab Mirrokni},
booktitle={Forty-second International Conference on Machine Learning},
year={2025},
url={https://openreview.net/forum?id=j3JBfFnGYh}
}

@inproceedings{denseformer,
author = {Pagliardini, Matteo and Mohtashami, Amirkeivan and Fleuret, Francois and Jaggi, Martin},
title = {DenseFormer: enhancing information flow in transformers via depth weighted averaging},
year = {2024},
isbn = {9798331314385},
publisher = {Curran Associates Inc.},
address = {Red Hook, NY, USA},
booktitle = {Proceedings of the 38th International Conference on Neural Information Processing Systems},
articleno = {4336},
numpages = {30},
location = {Vancouver, BC, Canada},
series = {NIPS '24}
}

@misc{gerasimov2025fullyutilizetransformersrepresentation,
      title={You Do Not Fully Utilize Transformer's Representation Capacity}, 
      author={Gleb Gerasimov and Yaroslav Aksenov and Nikita Balagansky and Viacheslav Sinii and Daniil Gavrilov},
      year={2025},
      eprint={2502.09245},
      archivePrefix={arXiv},
      primaryClass={cs.LG},
      url={https://arxiv.org/abs/2502.09245}, 
}

@misc{su2026lowrankattentionresiduals,
      title={Low-Rank Attention Residuals}, 
      author={Jonathan Su},
      year={2026},
      eprint={2607.09694},
      archivePrefix={arXiv},
      primaryClass={cs.LG},
      url={https://arxiv.org/abs/2607.09694}, 
}

@misc{luo2026multiheadattentionresiduals,
      title={Multi-Head Attention Residuals}, 
      author={Cheng Luo and Zefan Cai and Junjie Hu},
      year={2026},
      eprint={2607.27230},
      archivePrefix={arXiv},
      primaryClass={cs.AI},
      url={https://arxiv.org/abs/2607.27230}, 
}

@inproceedings{alkhamissi-etal-2021-adapting,
    title = "Adapting {MARBERT} for Improved {A}rabic Dialect Identification: Submission to the {NADI} 2021 Shared Task",
    author = "AlKhamissi, Badr  and
      Gabr, Mohamed  and
      ElNokrashy, Muhammad  and
      Essam, Khaled",
    editor = "Habash, Nizar  and
      Bouamor, Houda  and
      Hajj, Hazem  and
      Magdy, Walid  and
      Zaghouani, Wajdi  and
      Bougares, Fethi  and
      Tomeh, Nadi  and
      Abu Farha, Ibrahim  and
      Touileb, Samia",
    booktitle = "Proceedings of the Sixth Arabic Natural Language Processing Workshop",
    month = apr,
    year = "2021",
    address = "Kyiv, Ukraine (Virtual)",
    publisher = "Association for Computational Linguistics",
    url = "https://aclanthology.org/2021.wanlp-1.29/",
    pages = "260--264"
}

@misc{elnokrashy2024depthwiseattentiondwattlayer,
      title={Depth-Wise Attention (DWAtt): A Layer Fusion Method for Data-Efficient Classification}, 
      author={Muhammad ElNokrashy and Badr AlKhamissi and Mona Diab},
      year={2024},
      eprint={2209.15168},
      archivePrefix={arXiv},
      primaryClass={cs.CL},
      url={https://arxiv.org/abs/2209.15168}, 
}

@article{wang2025test,
  title={Test-time regression: a unifying framework for designing sequence models with associative memory},
  author={Wang, Ke Alexander and Shi, Jiaxin and Fox, Emily B},
  journal={arXiv preprint arXiv:2501.12352},
  year={2025}
}

@INPROCEEDINGS {resnet,
author = { He, Kaiming and Zhang, Xiangyu and Ren, Shaoqing and Sun, Jian },
booktitle = { 2016 IEEE Conference on Computer Vision and Pattern Recognition (CVPR) },
title = {{ Deep Residual Learning for Image Recognition }},
year = {2016},
volume = {},
ISSN = {1063-6919},
pages = {770-778},
doi = {10.1109/CVPR.2016.90},
url = {https://doi.ieeecomputersociety.org/10.1109/CVPR.2016.90},
publisher = {IEEE Computer Society},
address = {Los Alamitos, CA, USA},
month =Jun}

@misc{vaswani2023attentionneed,
      title={Attention Is All You Need}, 
      author={Ashish Vaswani and Noam Shazeer and Niki Parmar and Jakob Uszkoreit and Llion Jones and Aidan N. Gomez and Lukasz Kaiser and Illia Polosukhin},
      year={2023},
      eprint={1706.03762},
      archivePrefix={arXiv},
      primaryClass={cs.CL},
      url={https://arxiv.org/abs/1706.03762}, 
}

@misc{deepseekai2026deepseekv4highlyefficientmilliontoken,
      title={DeepSeek-V4: Towards Highly Efficient Million-Token Context Intelligence}, 
      author={DeepSeek-AI and Anyi Xu and Bangcai Lin and Bing Xue and Bingxuan Wang and Bingzheng Xu and Bochao Wu and Bowei Zhang and Chaofan Lin and Chen Dong and Chenchen Ling and Chengda Lu and Chenggang Zhao and Chengqi Deng and Chengyu Hou and Chenhao Xu and Chenze Shao and Chong Ruan and Conner Sun and Damai Dai and Daya Guo and Dejian Yang and Deli Chen and Donghao Li and Dongjie Ji and Erhang Li and Fang Wei and Fangyun Lin and Fangzhou Yuan and Feiyu Xia and Fucong Dai and Guangbo Hao and Guanting Chen and Guoai Cao and Guolai Meng and Guowei Li and Han Yu and Han Zhang and Hanwei Xu and Hao Li and Haofen Liang and Haoling Zhang and Haoming Luo and Haoran Wei and Haotian Yuan and Haowei Zhang and Haowen Luo and Haoyu Chen and Haozhe Ji and Hengqing Zhang and Honghui Ding and Hongxuan Tang and Huanqi Cao and Huazuo Gao and Hui Qu and Hui Zeng and J Yang and JQ Zhu and Jia Luo and Jia Song and Jia Yu and Jialiang Huang and Jialu Cai and Jian Liang and Jiangting Zhou and Jiasheng Ye and Jiashi Li and Jiaxin Xu and Jiewen Hu and Jieyu Yang and Jin Chen and Jin Yan and Jingchang Chen and Jingli Zhou and Jingting Xiang and Jingyang Yuan and Jingyuan Cheng and Jingzi Zhou and Jinhua Zhu and Jiping Yu and Joseph Sun and Jun Ran and Junguang Jiang and Junjie Qiu and Junlong Li and Junmin Zheng and Junxiao Song and Kai Dong and Kaige Gao and Kang Guan and Kexing Zhou and Kezhao Huang and Kuai Yu and Lean Wang and Lecong Zhang and Lei Wang and Leyi Xia and Li Zhang and Liang Zhao and Lihua Guo and Lingxiao Luo and Linwang Ma and Linyan Zhu and Litong Wang and Liyu Cai and Liyue Zhang and Longhao Chen and MS Di and MY Xu and Max Mei and Miaojun Wang and Mingchuan Zhang and Minghua Zhang and Minghui Tang and Mingming Li and Mingxu Zhou and Minmin Han and Ning Wang and Panpan Huang and Panpan Wang and Peixin Cong and Peiyi Wang and Peng Zhang and Qiancheng Wang and Qihao Zhu and Qingyang Li and Qinyu Chen and Qiushi Du and Qiwei Jiang and Rui Tian and Ruifan Xu and Ruijie Lu and Ruiling Xu and Ruiqi Ge and Ruisong Zhang and Ruizhe Pan and Runji Wang and Runqian Chen and Runqiu Yin and Runxin Xu and Ruomeng Shen and Ruoyu Zhang and Ruyi Chen and SH Liu and Shanghao Lu and Shangmian Sun and Shangyan Zhou and Shanhuang Chen and Shaofei Cai and Shaoheng Nie and Shaoqing Wu and Shaoyuan Chen and Shengding Hu and Shengyu Liu and Shiqiang Hu and Shirong Ma and Shiyu Wang and Shuiping Yu and Shunfeng Zhou and Shuting Pan and Shuying Yu and Songyang Zhou and Tao Ni and Tao Yun and Tian Jin and Tian Pei and Tian Ye and Tianle Lin and Tianran Ji and Tianyi Cui and Tianyuan Yue and Tingting Yu and Tun Wang and W Zhang and WL Xiao and Wangding Zeng and Wei An and Weilin Zhao and Wen Liu and Wenfeng Liang and Wenjie Pang and Wenjing Luo and Wenjing Yao and Wenjun Gao and Wenkai Yang and Wenlve Huang and Wenqing Hou and Wentao Zhang and Wenting Ma and Xi Gao and Xiang He and Xiangwen Wang and Xianzu Wang and Xiao Bi and Xiaodong Liu and Xiaohan Wang and Xiaokang Chen and Xiaokang Zhang and Xiaotao Nie and Xiaowen Sun and Xiaoxiang Wang and Xin Cheng and Xin Liu and Xin Xie and Xingchao Liu and Xingchen Liu and Xingkai Yu and Xingyou Li and Xinyu Yang and Xinyu Zhang and Xu Chen and Xuanyu Wang and Xuecheng Su and Xueyin Chen and Xuheng Lin and Xuwei Fu and YC Yan and YQ Wang and YW Ma and Yanfeng Luo and Yang Zhang and Yanhong Xu and Yanru Ma and Yanwen Huang and Yao Li and Yao Li and Yao Xu and Yao Zhao and Yaofeng Sun and Yaohui Wang and Yi Qian and Yi Shao and Yi Yu and Yichao Zhang and Yifan Ding and Yifan Shi and Yijia Wu and Yiliang Xiong and Yiling Ma and Ying He and Ying Tang and Ying Zhou and Yingjia Luo and Yinmin Zhong and Yishi Piao and Yisong Wang and Yixiang Zhang and Yixiao Chen and Yixuan Tan and Yixuan Wei and Yiyang Ma and Yiyuan Liu and Yonglun Yang and Yongqiang Guo and Yongtong Wu and Yu Wu and YuKun Li and Yuan Cheng and Yuan Ou and Yuanfan Xu and Yuanhao Li and Yuduan Wang and Yuehan Yang and Yuer Xu and Yuhan Wu and Yuhao Meng and Yuheng Zou and Yukun Zha and Yunfan Xiong and Yupeng Chen and Yuping Lin and Yuqian Cao and Yuqian Wang and Yushun Zhang and Yuting Yan and Yutong Lin and Yuxian Gu and Yuxiang Luo and Yuxiang You and Yuxuan Liu and Yuxuan Zhou and Yuyang Zhou and Yuzhen Huang and ZF Wu and Zehao Wang and Zehua Zhao and Zehui Ren and Zekai Zhang and Zhangli Sha and Zhe Fu and Zhe Ju and Zhean Xu and Zhenda Xie and Zhengyan Zhang and Zheren Gao and Zhewen Hao and Zhibin Gou and Zhicheng Ma and Zhigang Yan and Zhihong Shao and Zhixian Huang and Zhixuan Chen and Zhiyu Wu and Zhizhou Ren and Zhongyu Wu and Zhuoshu Li and Zhuping Zhang and Zian Xu and Zihao Wang and Zihua Qu and Zihui Gu and Zijia Zhu and Zilin Li and Zipeng Zhang and Ziwei Xie and Ziyi Gao and Ziyi Wan and Zizheng Pan and Zongqing Yao},
      year={2026},
      eprint={2606.19348},
      archivePrefix={arXiv},
      primaryClass={cs.CL},
      url={https://arxiv.org/abs/2606.19348}, 
}

@ARTICLE{correlationmatrixmemories,
  author={Kohonen, Teuvo},
  journal={IEEE Transactions on Computers}, 
  title={Correlation Matrix Memories}, 
  year={1972},
  volume={C-21},
  number={4},
  pages={353-359},
  doi={10.1109/TC.1972.5008975}}

@InProceedings{pmlr-v119-katharopoulos20a,
  title = 	 {Transformers are {RNN}s: Fast Autoregressive Transformers with Linear Attention},
  author =       {Katharopoulos, Angelos and Vyas, Apoorv and Pappas, Nikolaos and Fleuret, Fran{\c{c}}ois},
  booktitle = 	 {Proceedings of the 37th International Conference on Machine Learning},
  pages = 	 {5156--5165},
  year = 	 {2020},
  editor = 	 {III, Hal Daumé and Singh, Aarti},
  volume = 	 {119},
  series = 	 {Proceedings of Machine Learning Research},
  month = 	 {13--18 Jul},
  publisher =    {PMLR},
  url = 	 {https://proceedings.mlr.press/v119/katharopoulos20a.html}
}

@InProceedings{pmlr-v139-schlag21a,
  title = 	 {Linear Transformers Are Secretly Fast Weight Programmers},
  author =       {Schlag, Imanol and Irie, Kazuki and Schmidhuber, J{\"u}rgen},
  booktitle = 	 {Proceedings of the 38th International Conference on Machine Learning},
  pages = 	 {9355--9366},
  year = 	 {2021},
  editor = 	 {Meila, Marina and Zhang, Tong},
  volume = 	 {139},
  series = 	 {Proceedings of Machine Learning Research},
  month = 	 {18--24 Jul},
  publisher =    {PMLR},
  url = 	 {https://proceedings.mlr.press/v139/schlag21a.html}
}

@misc{touvron2023llamaopenefficientfoundation,
      title={LLaMA: Open and Efficient Foundation Language Models}, 
      author={Hugo Touvron and Thibaut Lavril and Gautier Izacard and Xavier Martinet and Marie-Anne Lachaux and Timothée Lacroix and Baptiste Rozière and Naman Goyal and Eric Hambro and Faisal Azhar and Aurelien Rodriguez and Armand Joulin and Edouard Grave and Guillaume Lample},
      year={2023},
      eprint={2302.13971},
      archivePrefix={arXiv},
      primaryClass={cs.CL},
      url={https://arxiv.org/abs/2302.13971}, 
}

@misc{yang2025qwen3technicalreport,
      title={Qwen3 Technical Report}, 
      author={An Yang and Anfeng Li and Baosong Yang and Beichen Zhang and Binyuan Hui and Bo Zheng and Bowen Yu and Chang Gao and Chengen Huang and Chenxu Lv and Chujie Zheng and Dayiheng Liu and Fan Zhou and Fei Huang and Feng Hu and Hao Ge and Haoran Wei and Huan Lin and Jialong Tang and Jian Yang and Jianhong Tu and Jianwei Zhang and Jianxin Yang and Jiaxi Yang and Jing Zhou and Jingren Zhou and Junyang Lin and Kai Dang and Keqin Bao and Kexin Yang and Le Yu and Lianghao Deng and Mei Li and Mingfeng Xue and Mingze Li and Pei Zhang and Peng Wang and Qin Zhu and Rui Men and Ruize Gao and Shixuan Liu and Shuang Luo and Tianhao Li and Tianyi Tang and Wenbiao Yin and Xingzhang Ren and Xinyu Wang and Xinyu Zhang and Xuancheng Ren and Yang Fan and Yang Su and Yichang Zhang and Yinger Zhang and Yu Wan and Yuqiong Liu and Zekun Wang and Zeyu Cui and Zhenru Zhang and Zhipeng Zhou and Zihan Qiu},
      year={2025},
      eprint={2505.09388},
      archivePrefix={arXiv},
      primaryClass={cs.CL},
      url={https://arxiv.org/abs/2505.09388}, 
}

@inproceedings{widrow1960adaptive,
  author    = {Widrow, Bernard and Hoff, Marcian E.},
  title     = {Adaptive Switching Circuits},
  booktitle = {IRE WESCON Convention Record},
  pages     = {96--104},
  address   = {Los Angeles, CA, USA},
  year      = {1960},
  month     = aug,
  note      = {Part 4}
}

@inproceedings{NEURIPS2018_4ff6fa96,
 author = {MacKay, Matthew and Vicol, Paul and Ba, Jimmy and Grosse, Roger B},
 booktitle = {Advances in Neural Information Processing Systems},
 editor = {S. Bengio and H. Wallach and H. Larochelle and K. Grauman and N. Cesa-Bianchi and R. Garnett},
 pages = {},
 publisher = {Curran Associates, Inc.},
 title = {Reversible Recurrent Neural Networks},
 url = {https://proceedings.neurips.cc/paper_files/paper/2018/file/4ff6fa96179cdc2838e8d8ce64cd10a7-Paper.pdf},
 volume = {31},
 year = {2018}
}

@misc{gu2022efficientlymodelinglongsequences,
      title={Efficiently Modeling Long Sequences with Structured State Spaces}, 
      author={Albert Gu and Karan Goel and Christopher Ré},
      year={2022},
      eprint={2111.00396},
      archivePrefix={arXiv},
      primaryClass={cs.LG},
      url={https://arxiv.org/abs/2111.00396}, 
}

@misc{shazeer2020gluvariantsimprovetransformer,
      title={GLU Variants Improve Transformer}, 
      author={Noam Shazeer},
      year={2020},
      eprint={2002.05202},
      archivePrefix={arXiv},
      primaryClass={cs.LG},
      url={https://arxiv.org/abs/2002.05202}, 
}

@article{JMLR:v23:21-0998,
  author  = {William Fedus and Barret Zoph and Noam Shazeer},
  title   = {Switch Transformers: Scaling to Trillion Parameter Models with Simple and Efficient Sparsity},
  journal = {Journal of Machine Learning Research},
  year    = {2022},
  volume  = {23},
  number  = {120},
  pages   = {1--39},
  url     = {http://jmlr.org/papers/v23/21-0998.html}
}

@InProceedings{layernormplacement,
  title = 	 {On Layer Normalization in the Transformer Architecture},
  author =       {Xiong, Ruibin and Yang, Yunchang and He, Di and Zheng, Kai and Zheng, Shuxin and Xing, Chen and Zhang, Huishuai and Lan, Yanyan and Wang, Liwei and Liu, Tieyan},
  booktitle = 	 {Proceedings of the 37th International Conference on Machine Learning},
  pages = 	 {10524--10533},
  year = 	 {2020},
  editor = 	 {III, Hal Daumé and Singh, Aarti},
  volume = 	 {119},
  series = 	 {Proceedings of Machine Learning Research},
  month = 	 {13--18 Jul},
  publisher =    {PMLR},
  url = 	 {https://proceedings.mlr.press/v119/xiong20b.html}
}

@article{yang2024parallelizing,
  title={Parallelizing linear transformers with the delta rule over sequence length},
  author={Yang, Songlin and Wang, Bailin and Zhang, Yu and Shen, Yikang and Kim, Yoon},
  journal={Advances in neural information processing systems},
  volume={37},
  pages={115491--115522},
  year={2024}
}

@article{team2025kimi,
  title={Kimi linear: An expressive, efficient attention architecture},
  author={{Kimi Team} and Zhang, Yu and Lin, Zongyu and Yao, Xingcheng and Hu, Jiaxi and Meng, Fanqing and Liu, Chengyin and Men, Xin and Yang, Songlin and Li, Zhiyuan and others},
  journal={arXiv preprint arXiv:2510.26692},
  year={2025}
}

@inproceedings{yang2025gated,
  title={Gated delta networks: Improving mamba2 with delta rule},
  author={Yang, Songlin and Kautz, Jan and Hatamizadeh, Ali},
  booktitle={International Conference on Learning Representations},
  volume={2025},
  pages={29687--29707},
  year={2025}
}

@article{sinkhorn1967concerning,
  title={Concerning nonnegative matrices and doubly stochastic matrices},
  author={Sinkhorn, Richard and Knopp, Paul},
  journal={Pacific Journal of Mathematics},
  volume={21},
  number={2},
  pages={343--348},
  year={1967},
  publisher={Mathematical Sciences Publishers}
}

@misc{clark2018thinksolvedquestionanswering,
      title={Think you have Solved Question Answering? Try ARC, the AI2 Reasoning Challenge}, 
      author={Peter Clark and Isaac Cowhey and Oren Etzioni and Tushar Khot and Ashish Sabharwal and Carissa Schoenick and Oyvind Tafjord},
      year={2018},
      eprint={1803.05457},
      archivePrefix={arXiv},
      primaryClass={cs.AI},
      url={https://arxiv.org/abs/1803.05457}, 
}

@misc{zellers2019hellaswagmachinereallyfinish,
      title={HellaSwag: Can a Machine Really Finish Your Sentence?}, 
      author={Rowan Zellers and Ari Holtzman and Yonatan Bisk and Ali Farhadi and Yejin Choi},
      year={2019},
      eprint={1905.07830},
      archivePrefix={arXiv},
      primaryClass={cs.CL},
      url={https://arxiv.org/abs/1905.07830}, 
}

@misc{bisk2019piqareasoningphysicalcommonsense,
      title={PIQA: Reasoning about Physical Commonsense in Natural Language}, 
      author={Yonatan Bisk and Rowan Zellers and Ronan Le Bras and Jianfeng Gao and Yejin Choi},
      year={2019},
      eprint={1911.11641},
      archivePrefix={arXiv},
      primaryClass={cs.CL},
      url={https://arxiv.org/abs/1911.11641}, 
}

@misc{mihaylov2018suitarmorconductelectricity,
      title={Can a Suit of Armor Conduct Electricity? A New Dataset for Open Book Question Answering}, 
      author={Todor Mihaylov and Peter Clark and Tushar Khot and Ashish Sabharwal},
      year={2018},
      eprint={1809.02789},
      archivePrefix={arXiv},
      primaryClass={cs.CL},
      url={https://arxiv.org/abs/1809.02789}, 
}

@misc{merity2016pointersentinelmixturemodels,
      title={Pointer Sentinel Mixture Models}, 
      author={Stephen Merity and Caiming Xiong and James Bradbury and Richard Socher},
      year={2016},
      eprint={1609.07843},
      archivePrefix={arXiv},
      primaryClass={cs.CL},
      url={https://arxiv.org/abs/1609.07843}, 
}

@misc{paperno2016lambadadatasetwordprediction,
      title={The LAMBADA dataset: Word prediction requiring a broad discourse context}, 
      author={Denis Paperno and Germán Kruszewski and Angeliki Lazaridou and Quan Ngoc Pham and Raffaella Bernardi and Sandro Pezzelle and Marco Baroni and Gemma Boleda and Raquel Fernández},
      year={2016},
      eprint={1606.06031},
      archivePrefix={arXiv},
      primaryClass={cs.CL},
      url={https://arxiv.org/abs/1606.06031}, 
}

@misc{eval-harness,
  author       = {Gao, Leo and Tow, Jonathan and Abbasi, Baber and Biderman, Stella and Black, Sid and DiPofi, Anthony and Foster, Charles and Golding, Laurence and Hsu, Jeffrey and Le Noac'h, Alain and Li, Haonan and McDonell, Kyle and Muennighoff, Niklas and Ociepa, Chris and Phang, Jason and Reynolds, Laria and Schoelkopf, Hailey and Skowron, Aviya and Sutawika, Lintang and Tang, Eric and Thite, Anish and Wang, Ben and Wang, Kevin and Zou, Andy},
  title        = {The Language Model Evaluation Harness},
  month        = 07,
  year         = 2024,
  publisher    = {Zenodo},
  version      = {v0.4.3},
  doi          = {10.5281/zenodo.12608602},
  url          = {https://zenodo.org/records/12608602}
}

@software{yang2024fla,
  title  = {FLA: A Triton-Based Library for Hardware-Efficient Implementations of Linear Attention Mechanism},
  author = {Yang, Songlin and Zhang, Yu},
  url    = {https://github.com/fla-org/flash-linear-attention},
  month  = jan,
  year   = {2024}
}

@inproceedings{groeneveld-etal-2024-olmo,
    title = "{OLM}o: Accelerating the Science of Language Models",
    author = "Groeneveld, Dirk  and
      Beltagy, Iz  and
      Walsh, Evan  and
      Bhagia, Akshita  and
      Kinney, Rodney  and
      Tafjord, Oyvind  and
      Jha, Ananya  and
      Ivison, Hamish  and
      Magnusson, Ian  and
      Wang, Yizhong  and
      Arora, Shane  and
      Atkinson, David  and
      Authur, Russell  and
      Chandu, Khyathi  and
      Cohan, Arman  and
      Dumas, Jennifer  and
      Elazar, Yanai  and
      Gu, Yuling  and
      Hessel, Jack  and
      Khot, Tushar  and
      Merrill, William  and
      Morrison, Jacob  and
      Muennighoff, Niklas  and
      Naik, Aakanksha  and
      Nam, Crystal  and
      Peters, Matthew  and
      Pyatkin, Valentina  and
      Ravichander, Abhilasha  and
      Schwenk, Dustin  and
      Shah, Saurabh  and
      Smith, William  and
      Strubell, Emma  and
      Subramani, Nishant  and
      Wortsman, Mitchell  and
      Dasigi, Pradeep  and
      Lambert, Nathan  and
      Richardson, Kyle  and
      Zettlemoyer, Luke  and
      Dodge, Jesse  and
      Lo, Kyle  and
      Soldaini, Luca  and
      Smith, Noah  and
      Hajishirzi, Hannaneh",
    editor = "Ku, Lun-Wei  and
      Martins, Andre  and
      Srikumar, Vivek",
    booktitle = "Proceedings of the 62nd Annual Meeting of the Association for Computational Linguistics (Volume 1: Long Papers)",
    month = aug,
    year = "2024",
    address = "Bangkok, Thailand",
    publisher = "Association for Computational Linguistics",
    url = "https://aclanthology.org/2024.acl-long.841/",
    doi = "10.18653/v1/2024.acl-long.841",
    pages = "15789--15809"
}

@misc{shoeybi2020megatronlmtrainingmultibillionparameter,
      title={Megatron-LM: Training Multi-Billion Parameter Language Models Using Model Parallelism}, 
      author={Mohammad Shoeybi and Mostofa Patwary and Raul Puri and Patrick LeGresley and Jared Casper and Bryan Catanzaro},
      year={2020},
      eprint={1909.08053},
      archivePrefix={arXiv},
      primaryClass={cs.CL},
      url={https://arxiv.org/abs/1909.08053}, 
}

@misc{sakaguchi2020winogrande,
      title={WinoGrande: An Adversarial Winograd Schema Challenge at Scale}, 
      author={Keisuke Sakaguchi and Ronan Le Bras and Chandra Bhagavatula and Yejin Choi},
      year={2019},
      eprint={1907.10641},
      archivePrefix={arXiv},
      primaryClass={cs.CL},
      url={https://arxiv.org/abs/1907.10641}, 
}

@article{DBLP:journals/corr/abs-1910-07467,
  author       = {Biao Zhang and
                  Rico Sennrich},
  title        = {Root Mean Square Layer Normalization},
  journal      = {CoRR},
  volume       = {abs/1910.07467},
  year         = {2019},
  url          = {http://arxiv.org/abs/1910.07467},
  eprinttype   = {arXiv},
  eprint       = {1910.07467},
  bibsource    = {dblp computer science bibliography, https://dblp.org}
}

@misc{liou2026replayssm,
  title        = {{ReplaySSM}: Cache {SSM} Inputs, Not State},
  author       = {Liou, Ze-Wei and Dao, Tri},
  year         = {2026},
  month        = jun,
  howpublished = {\url{https://tridao.me/blog/2026/replayssm/}},
  note         = {Blog post}
}

@misc{siems2025deltaproductimprovingstatetrackinglinear,
      title={DeltaProduct: Improving State-Tracking in Linear RNNs via Householder Products}, 
      author={Julien Siems and Timur Carstensen and Arber Zela and Frank Hutter and Massimiliano Pontil and Riccardo Grazzi},
      year={2025},
      eprint={2502.10297},
      archivePrefix={arXiv},
      primaryClass={cs.LG},
      url={https://arxiv.org/abs/2502.10297}, 
}

@misc{tumma2026preconditioneddeltanetcurvatureawaresequence,
      title={Preconditioned DeltaNet: Curvature-aware Sequence Modeling for Linear Recurrences}, 
      author={Neehal Tumma and Noel Loo and Daniela Rus},
      year={2026},
      eprint={2604.21100},
      archivePrefix={arXiv},
      primaryClass={cs.LG},
      url={https://arxiv.org/abs/2604.21100}, 
}

@misc{wang2026depthbenchmeasuringresidualconnections,
      title={DepthBench: Measuring How Residual Connections Enable More Computational Depth}, 
      author={Keyu Wang and Yangyi Huang and Jiale Kang and David González-Martínez and Weiyang Liu and Shiwei Liu},
      year={2026},
      eprint={2609.32534},
      archivePrefix={arXiv},
      primaryClass={cs.CV},
      url={https://arxiv.org/abs/2609.32534}, 
}
\bibliographystyle{iclr2027_conference}

\appendix

\section{Extended Related Work}
\label{sec:related-work}

Our work sits at the intersection of two lines of research: mechanisms that enrich information flow across transformer depth, particularly through depth mixing, and associative memory mechanisms developed primarily for sequence modeling. We discuss both connections below.

\subsection{Information Flow Across Transformer Depth}

\textbf{Enriching the residual representation.} The standard transformer propagates information across depth through a single additive residual stream. A number of works have explored richer alternatives to this formulation. SiameseNorm \citep{li2026siamesenormbreakingbarrierreconciling} maintains two streams that complement each other with different normalization approaches. Moreover, residual matrix transformers \citep{mak2025residualmatrixtransformersscaling} propagate a matrix-valued residual representation. Hyper-Connections \citep{zhu2025hyperconnections} introduce multiple residual channels along with learned mappings between them, and Manifold-Constrained Hyper-Connections (mHC) \citep{xie2026mhcmanifoldconstrainedhyperconnections} (and its follow-up mHC-lite \citep{yang2026mhclitedontneed20}) enhance their stability. xHC~\citep{zhang2026xhcexpandedhyperconnections} expands the residual stream further, updating only a subset of its streams in each layer and enriching what each layer writes back. Related approaches include Deep Delta Learning \citep{zhang2026deepdeltalearning}, which uses a delta-like rule to update a (possibly matrix-sized) residual stream, and other architectures that increase the capacity or flexibility of the representations propagated between layers.

These methods modify the residual representation itself. LDM instead preserves the standard additive residual stream and introduces a separate recurrent associative state.

\textbf{Explicit depth mixing.} A second family of methods directly combines representations produced at different depths. We refer to this general class of mechanisms as explicit depth mixing. Unlike ordinary residual accumulation, depth mixing methods allow the computation at one layer to selectively incorporate representations produced at earlier layers.

Several works have explored representation aggregation across layers.  DenseFormer \citep{denseformer} combines preceding hidden representations using learned depth-dependent weights. Cross-Layer Retrospective Retrieving via Layer Attention \citep{fang2023crosslayerretrospectiveretrievinglayer} uses attention to retrieve representations from earlier layers. Although this work considers a formulation similar to our memory, its final methods are different from ours. DeepCrossAttention \citep{heddes2025deepcrossattention} considers richer cross-layer mixing, including input-dependent mechanisms. MUDDFormer \citep{xiao2025muddformer} similarly constructs layer inputs by dynamically combining representations across depth, and LIMe \citep{gerasimov2025fullyutilizetransformersrepresentation} also revisits the use of intermediate transformer representations. Earlier work also introduced Vertical Attention \citep{alkhamissi-etal-2021-adapting} and Depth-Wise Attention \citep{elnokrashy2024depthwiseattentiondwattlayer}, likewise exploring attention or fusion across network layers.

Most closely related to our framing, Attention Residuals (AttnRes) \citep{kimiteam2026attentionresiduals} replaces fixed residual accumulation with softmax attention over representations from the embedding and the preceding layers. \Cref{fig:attnres-architecture} illustrates Full AttnRes. Its blockwise variant applies the same operation to coarser groups of layers. Subsequent work has considered alternative parameterizations of attention-based residual mixing \citep{su2026lowrankattentionresiduals,luo2026multiheadattentionresiduals}.

LDM differs from these approaches in how previous depth representations are made available. Explicit depth mixing mechanisms retain or otherwise expose a collection of previous representations and directly form a mixture over them. LDM instead recurrently compresses key--value associations from the embedding and the previous layers into a fixed-size state. A later layer retrieves from this state using a content-dependent query. As shown in the main text, the simplest LDM recurrence can therefore be interpreted as implementing linearized attention over depth \emph{implicitly}, without explicitly materializing the collection of previous depth representations.

\textbf{Use in large-scale runs.} Richer mechanisms for information propagation across depth have also begun to appear in large-scale language models. Recent models, including DeepSeek-V4 \citep{deepseekai2026deepseekv4highlyefficientmilliontoken}, Kimi K3 \citep{kimiteam2026kimik3openfrontier}, and Qwen3.8-Flash-Next \citep{qiu2026designqwen38nextarchitectureevaluation}, incorporate variants of enriched residual or cross-layer routing mechanisms. This provides additional evidence that modifying information flow across transformer depth works at large scale and appears to be a promising scaling axis.

\subsection{Associative Memories, Linear Attention and Regression}

Our second connection is to the literature on linear attention and associative memory. A linear associative memory~\citep{correlationmatrixmemories} stores key--value associations through a matrix-valued state and retrieves values using a query. Under the simplest outer product update, this recurrence is informally equivalent to unnormalized linear attention~\citep{pmlr-v119-katharopoulos20a,pmlr-v139-schlag21a}.

Recent work has further interpreted such memories as online regression problems in the context of sequence mixing. In particular, Test-Time Regression \citep{wang2025test} provides a regression perspective on sequence models with associative memory. From this view, delta rule models~\citep{pmlr-v139-schlag21a,yang2024parallelizing,yang2025gated,team2025kimi}, whose update goes back to the Widrow--Hoff rule~\citep{widrow1960adaptive}, take one online gradient step on the regression objective per token. DeltaProduct~\citep{siems2025deltaproductimprovingstatetrackinglinear} takes several steps per token, and the later Preconditioned DeltaNet~\citep{tumma2026preconditioneddeltanetcurvatureawaresequence} accounts for the curvature of the least squares loss with a (diagonal) preconditioner. Closely related ideas appear in work on mesa-optimization. The original MesaLayer \citep{vonoswald2024uncoveringmesaoptimizationalgorithmstransformers} uses a recursive least squares update implemented through the Sherman--Morrison identity, maintaining the exact regression solution recurrently along the sequence axis. This recurrence is inherently sequential, which makes efficient sequence training challenging. To address this, subsequent MesaNet work \citep{vonoswald2026mesanetsequencemodelinglocally} instead develops iterative solvers that permit efficient parallel training over the sequence dimension.

LDM transfers this perspective from the sequence axis to the depth axis. Each layer gives a key--value observation, and the memory stream propagated through the network maintains a regression solution over representations produced by preceding layers. This distinction removes an important systems constraint. In particular, transformer layers are naturally evaluated sequentially in depth, so the memory update does not need to be parallelizable along that axis. We can therefore use a Sherman--Morrison rank-one recursive least squares update directly, obtaining the exact solution to the weighted depth regression objective described in the main text. We note, however, that using techniques like unwinding or windowing (\cref{sec:efficiency}) might have additional constraints.

\subsection{Relation of Our Hardware-Aware Implementation to Past Work}
During training, our hardware-aware implementation avoids storing intermediate states during the forward pass by ``unwinding'' the memory, as explained in \cref{app:efficiency}. Although different, this has some connections to past work. Reversible residual networks~\citep{gomez2017reversibleresidualnetworkbackpropagation} split the residual stream into two halves that update each other additively, so that the input of a layer can be recomputed from its output and no activations have to be stored for the backward pass of ``reversible'' layers. Reversible recurrent networks~\citep{NEURIPS2018_4ff6fa96} use a similar technique along the sequence axis to reconstruct earlier hidden states from later ones. Since forgetting makes their update not exactly invertible, they store the information it discards.

Windowed writes also have a counterpart along the sequence axis, although only for inference, since training generally uses chunkwise-parallel algorithms. ReplaySSM \citep{liou2026replayssm} recently used a similar idea to speed up decoding in Mamba-2 and Gated DeltaNet by caching the most recent rank-one updates and writing the state back only when the cache is full. Our windowed writes apply the same idea across depth, during training as well as inference.
\section{Experimental Details}\label{app:exp_details}

We train on the FineWeb-Edu dataset \citep{penedo2024finewebdatasetsdecantingweb}. In particular, we train on (subsets of) the publicly available 100B token subset\footnote{\url{https://huggingface.co/datasets/HuggingFaceFW/fineweb-edu/viewer/sample-100BT}}. For all experiments, data ordering is fixed for every baseline to ensure fairness. Validation loss is taken from a held-out shard of the 100B subset.

Data is tokenized with OLMo-core's GPT-NeoX Dolma v1.5 tokenizer \citep{groeneveld-etal-2024-olmo} (with a vocabulary size of 50,280). Documents are packed into sequences of 2,048 tokens.

All models are trained with AdamW ($\beta_1=0.9$, $\beta_2=0.95$, $\epsilon=10^{-8}$, weight decay $0.1$) and global gradient norm clipping at $1.0$. RMSNorm gains are excluded from weight decay for all methods. We use a global batch of 512 sequences, i.e., 1,048,576 tokens per optimizer step. The learning rate follows a cosine schedule, decaying to $10\%$ of its peak, with a linear warmup from $0$ over the first 1,000 steps at 400M and 2,000 steps at 1.5B. The 400M models are trained for 47,684 steps (50.0B tokens) and the 1.5B models for 94,972 steps (99.6B tokens).

\subsection{Architecture}

For a matched comparison, all baselines use the same backbone configurations, and only the method-specific details vary. Details about our Llama-like and Qwen-like architectures can be found in \cref{tab:model-configs}.
\begin{table}[t]
\centering
\caption{Backbone configurations used in our experiments. The bottom rows list the parameters that each method adds to the backbone.}
\label{tab:model-configs}
\small
\setlength{\tabcolsep}{5pt}
\begin{tabular}{lcc}
\toprule
\textbf{Configuration} & \textbf{400M Llama-like} & \textbf{1.5B Qwen-like} \\
\midrule
Backbone parameters         & 405.4M         & 1.512B          \\
Transformer blocks          & 24             & 28              \\
Model width $D$
                            & 1,024          & 2,048           \\
Attention type              & MHA            & GQA             \\
Query heads                 & 16             & 16              \\
Key/value heads             & 16             & 8               \\
Queries per KV head         & 1              & 2               \\
Head dimension              & 64             & 128             \\
Query projection width      & 1,024          & 2,048           \\
Key/value projection width  & 1,024          & 1,024           \\
FFN type                     & SwiGLU         & SwiGLU          \\
FFN intermediate width      & 2,736          & 6,144           \\
Normalization               & Pre-RMSNorm    & Pre-RMSNorm     \\
RMSNorm $\epsilon$          & $10^{-6}$      & $10^{-6}$       \\
Query/key normalization     & None           & RMSNorm          \\
RoPE base $\theta$          & $10^{4}$       & $10^{6}$        \\
Maximum sequence length     & 2,048          & 2,048           \\
Vocabulary size             & 50,280         & 50,280          \\
Input/output embeddings     & Untied         & Tied            \\
Attention bias              & No             & No              \\
FFN bias                    & No             & No              \\
Dropout                     & 0              & 0               \\
Master/compute precision    & FP32/BF16      & FP32/BF16       \\
Initialization std.\        & 0.02           & 0.02            \\
\midrule
\multicolumn{3}{l}{\textit{Parameters added by the depth mixing method}} \\
AttnRes (Block or Full)     & 0.10M (0.02\%) & 0.23M (0.02\%)  \\
mHC                         & 4.94M (1.22\%) & 11.51M (0.76\%) \\
LDM                         & 3.78M (0.93\%) & 16.25M (1.07\%) \\
LDM, $D_k=4$                & 3.58M (0.88\%) & --              \\
\bottomrule
\end{tabular}
\end{table}

\subsection{Implementation and baselines}

Our codebase is adapted from DepthBench \citep{wang2026depthbenchmeasuringresidualconnections}\footnote{\url{https://github.com/keyu-wang-2002/DepthBench}}, which is itself based on OLMo-core \citep{olmo20242olmo2furious}\footnote{\url{https://github.com/allenai/OLMo-core}}. Each baseline is implemented in our codebase following its description. For some unspecified details, we used public conventions or tried different approaches in our initial explorations, which we describe below.

\textbf{PreLN.}
We retain the standard OLMo-core pre-normalized transformer implementation. For the Qwen-like backbone, attention Q/K normalization is applied independently within each head. The same backbone normalization settings are used across methods.

\textbf{AttnRes.}
We adapt FLA's fused AttnRes implementation\footnote{\url{https://github.com/fla-org/flash-linear-attention/blob/02e270330b4fc5bb2fd738281b5d594bc89e6d5f/fla/ops/attnres/fused.py}}. We use zero-initialized depth queries as described in the original paper. Routing parameters follow the backbone's mixed precision policy. Following the recommendations of \citet{kimiteam2026attentionresiduals}, for the block variant, we use eight depth blocks at 400M and seven at 1.5B. For the latter, we use seven so that every block contains whole transformer layers (in this case, this gives four per block). The RMSNorm used for the keys has a learnable gain but no bias, following popular implementations like FLA or Megatron \citep{shoeybi2020megatronlmtrainingmultibillionparameter}. We did not test removing it.

The original paper does not specify whether the depth queries use weight decay. Decay pushes the queries towards zero and thus biases the mixing towards uniform weights. In preliminary experiments, both choices performed similarly, with a slight but consistent advantage for using decay for the queries. We hence use weight decay for the queries in our main runs.

\textbf{mHC.}
We use DeepSeek's official mHC implementation in TileKernels,\footnote{\url{https://github.com/deepseek-ai/TileKernels}} with four residual streams and 20 Sinkhorn iterations. We use their fused TileLang kernels for the three coefficient mappings and the two application kernels, along with their precision policy (TF32 controller projection, FP32 controller parameters and scalars, and BF16 residual streams), as well as their block-wise activation recomputation, which stores one residual entry per $L_r$ mHC layers and rebuilds the remaining widened streams during the backward pass. We use the block size described in the mHC paper, $L_r \approx \sqrt{NL/(N+2)} = 6$.

Following \citet{xie2026mhcmanifoldconstrainedhyperconnections}, the controller projection is initialized to zero, and its biases are initialized as follows. First, the pre-layer mixing bias is $-3$ except for the layer's selected stream, which is $+3$, and the residual mixing bias is $+3$ on the diagonal and $-3$ elsewhere. The post-mapping bias is initialized to zero, and the selected stream of each connector cycles with the connector's index modulo $N$. The learned logit scales are initialized to $0.01$. In some preliminary experiments where we used other initializations, this did not appear to impact performance. This detail appears to only be available in the ICML version of the paper.

We exclude biases and logit scales from weight decay. Controller RMS normalization, pre-mixing sigmoid stabilization, and Sinkhorn stabilization use $\epsilon=10^{-6}$. We did not try ablating this.

There are other details undisclosed in the mHC paper. First, in our implementation, the first layer expands the $D$-dimensional residual stream into an $N \times D$-dimensional one by replicating it $N$ times. This appears to be in line with other public implementations like Megatron and the official DeepSeek-V4 checkpoints, so we use it. Second, for the final layer, Megatron offers the option to use an average of the $N$ channels to reduce it to a single vector; however, it also supports a final ``head'' that increases expressivity. The released DeepSeek-V4 \citep{deepseekai2026deepseekv4highlyefficientmilliontoken} checkpoint also uses such a learnable head. In early experiments, we found that it had little effect on the training curves but seemed consistently equal to or better than the alternatives. Because of that, and to stay close to DeepSeek-V4, we use the learnable head. Its projection is initialized from $\mathcal{N}(0,0.02^2)$ (similar to the backbone's matrices), its bias to zero, and its logit scale to $0.01$.

\textbf{LDM.}
Our ablations run on a more flexible implementation with no windowing. On the other hand, the main training results follow the implementation in \cref{app:efficiency}. In particular, we use state unwinding (\cref{eq:unwinding}) and windowed writes with a window size of $E=4$. The settings are those of \cref{sec:experiments}: $G=16$ groups, $D_k=8$ (except for the $D_k=4$ result at 400M scale), and $H$ set to 32 at 400M and 64 at 1.5B scales, respectively. We do not use weight decay on the read gains, the gate biases, and the forgetting logits. The linear layers of the address networks (for queries, keys) are initialized in the same way as the backbone linear layers. Their gate heads start with zero weights and biases, so that $\mathbf g=\boldsymbol\omega=\mathbf 1$ at initialization. The read gains are initialized to one. The forgetting factors are sigmoids of learned logits and are initialized to $0.99$.

\subsection{Training}

We use FSDP with FP32 master parameters and BF16 computation, with gradient reductions in FP32.  For the 400M models, we use the equivalent of DDP (no sharding) but utilize the FSDP API to ensure an identical precision policy.

\textbf{Hyperparameter tuning. }To enable a fair comparison, we tune the learning rates of each baseline at both scales. We use a grid anchored at $10^{-3}$, with candidates multiplied or divided by powers of two (e.g.\ $2.5\times10^{-4}$, $5\times10^{-4}$, $10^{-3}$, $2\times10^{-3}$, $4\times10^{-3}$). For each baseline and scale, we run candidate learning rates until we find either three consecutive values, where the middle one has the lowest held-out loss, or two consecutive values whose held-out losses differ by less than 0.001 nats (in these cases, training and held-out validation losses often flip), and we select the value with the lowest held-out loss. The final checkpoint of the selected run is used for all evaluations. Other hyperparameters are kept fixed. The resulting sweeps are reported in \cref{app:lr-sweeps}.

\subsection{Evaluation}

We evaluate the final checkpoint of each selected pretraining run with the \texttt{lm-eval-harness} \citep{eval-harness} in zero-shot mode, with a maximum length of 2,048 tokens. The suite includes ARC-Easy and ARC-Challenge~\citep{clark2018thinksolvedquestionanswering}, HellaSwag~\citep{zellers2019hellaswagmachinereallyfinish}, PIQA~\citep{bisk2019piqareasoningphysicalcommonsense}, WinoGrande~\citep{sakaguchi2020winogrande}, OpenBookQA~\citep{mihaylov2018suitarmorconductelectricity}, WikiText~\citep{merity2016pointersentinelmixturemodels} (word perplexity), and LAMBADA~\citep{paperno2016lambadadatasetwordprediction} (accuracy and perplexity).
\section{Efficiency Measurements}
\label{app:efficiency-results}
\makeatletter\setlength{\@fptop}{0pt}\setlength{\@fpsep}{2\baselineskip}\makeatother

We measure different notions of efficiency for every method on Qwen3 models (1.7B to 14B parameters; we retain the native vocabulary size of 151936 for all of them) on NVIDIA B200 GPUs, in BF16 with compiled models, and report measurements as a percentage of the corresponding PreLN baseline. The models follow the 1.5B Qwen-like configuration of \cref{tab:model-configs}, with the changes listed in \cref{tab:efficiency-models}. Every point in the plots below represents the median over three rounds.

We made several adjustments to increase the fairness of the measurements. Before inference, we fold normalization weights into the mHC controllers \citep{xie2026mhcmanifoldconstrainedhyperconnections}. ``mHC (released kernels)'' denotes DeepSeek's TileKernels as released (though with several adaptations around the exact kernels to make them compatible with our codebase). The mHC row replaces their projection backward with a faster matmul operation and splits the projection at inference, both of which we found to work better in terms of efficiency (with exact improvements varying per setting). ``mHC, no recompute'' keeps every stream instead of recomputing them (recomputation follows the approach in \citet{xie2026mhcmanifoldconstrainedhyperconnections}). Our main figures, for fairness, always report the ``mHC'' setting. Here we report the entire set for completeness. Still, it is possible that other optimizations might produce different results.

\textbf{Training} (\cref{fig:efficiency-training}). We measure the optimizer step of about 1M tokens on 8 GPUs with FSDP and PyTorch's efficient implementation of scaled dot product attention (SDPA). We use the largest microbatch at which every method fits at 2K tokens, and half at 4K tokens (with half the sequences per step, so each step still covers about 1M tokens).

\textbf{Prefill} (\cref{fig:efficiency-prefill}). We measure forward over the entire prompt on one GPU, keeping only the last position's logits, from a single prompt up to the largest batch size that fits for all methods.

\textbf{Decode} (\cref{fig:efficiency-decode,fig:efficiency-first-token}). We measure KV-cached greedy decoding on one GPU with FlashAttention~4. The prompt is prefilled in 512-token chunks, which gives the time to first token. Each generated token then replays a CUDA graph of one decode step.

\textbf{Peak memory} (\cref{tab:efficiency-memory,fig:memory-training,fig:memory-prefill,fig:memory-decode}). We report the peak memory allocated per GPU in GiB in the settings of the main text figure (\cref{tab:efficiency-memory}): the training step with sequences of 2K tokens, and prefill and decode at batch 64 with a 2K token prompt or context. \Cref{fig:memory-training,fig:memory-prefill,fig:memory-decode} report peak memory as a percentage of PreLN in every setting above.

\begin{table}[tbp]
\centering
\small
\caption{Qwen3 shapes of the efficiency measurements. All other settings are those of the 1.5B Qwen-like model in \cref{tab:model-configs}.}
\label{tab:efficiency-models}
\begin{tabular}{lcccc}
\toprule
 & 1.7B & 4B & 8B & 14B \\
\midrule
Transformer blocks & 28 & 36 & 36 & 40 \\
Model width $D$ & 2,048 & 2,560 & 4,096 & 5,120 \\
Query heads & 16 & 32 & 32 & 40 \\
Key/value heads & 8 & 8 & 8 & 8 \\
Head dimension & 128 & 128 & 128 & 128 \\
FFN intermediate width & 6,144 & 9,728 & 12,288 & 17,408 \\
Vocabulary size & 151,936 & 151,936 & 151,936 & 151,936 \\
Input/output embeddings & Tied & Tied & Untied & Untied \\
LDM groups $G$ & 16 & 20 & 32 & 40 \\
\bottomrule
\end{tabular}
\end{table}

\begin{table}[tbp]
\centering
\small
\caption{Peak memory allocated per GPU (GiB). Training runs on 8 GPUs with sequences of 2K tokens; prefill and decode run on one GPU at batch 64 with a 2K token prompt or context. Recomputation does not apply at inference, so the mHC row without recompute is training only.}
\label{tab:efficiency-memory}
\begin{tabular}{lrrrr}
\toprule
 & 1.7B & 4B & 8B & 14B \\
\midrule
\multicolumn{5}{l}{\textit{Training}} \\
PreLN & 80.2 & 82.6 & 107.6 & 97.3 \\
LDM & 94.0 & 92.8 & 123.2 & 107.4 \\
AttnRes (Block) & 80.3 & 82.8 & 107.7 & 97.4 \\
AttnRes (Full) & 90.5 & 94.0 & 126.7 & 111.0 \\
mHC & 80.3 & 81.1 & 106.2 & 95.3 \\
mHC (released kernels) & 80.3 & 81.1 & 104.8 & 94.9 \\
mHC, no recompute & 115.3 & 110.3 & 151.5 & 126.7 \\
\midrule
\multicolumn{5}{l}{\textit{Prefill}} \\
PreLN & 7.8 & 14.3 & 24.4 & 39.9 \\
LDM & 14.7 & 22.9 & 38.1 & 57.1 \\
AttnRes (Block) & 11.3 & 19.9 & 33.4 & 52.4 \\
AttnRes (Full) & 34.8 & 58.0 & 94.4 & 137.4 \\
mHC & 10.8 & 18.1 & 30.5 & 47.5 \\
mHC (released kernels) & 10.8 & 18.1 & 30.5 & 47.5 \\
\midrule
\multicolumn{5}{l}{\textit{Decode}} \\
PreLN & 19.4 & 28.6 & 36.9 & 52.1 \\
LDM & 21.2 & 30.8 & 40.4 & 56.5 \\
AttnRes (Block) & 20.4 & 30.2 & 39.4 & 55.6 \\
AttnRes (Full) & 26.3 & 39.7 & 54.7 & 76.8 \\
mHC & 20.2 & 29.6 & 38.6 & 54.2 \\
mHC (released kernels) & 20.2 & 29.6 & 38.6 & 54.2 \\
\bottomrule
\end{tabular}
\end{table}

\begin{figure}[p]
\centering
\input{media/fig-efficiency-training.pgf}
\caption{Training throughput as a percentage of PreLN.}
\label{fig:efficiency-training}

\vspace{1.5\baselineskip}
\input{media/fig-memory-training.pgf}
\caption{Peak memory per GPU of the training step as a percentage of PreLN.}
\label{fig:memory-training}

\vspace{1.5\baselineskip}
\resizebox{!}{0.4\textheight}{\input{media/fig-efficiency-prefill.pgf}}
\caption{Prefill throughput as a percentage of PreLN, by prompt length (rows) and model size (columns).}
\label{fig:efficiency-prefill}
\end{figure}

\begin{figure}[p]
\centering
\resizebox{!}{0.4\textheight}{\input{media/fig-efficiency-first-token.pgf}}
\caption{Time to first token relative to PreLN (PreLN's time divided by the method's), by prompt length (rows) and model size (columns).}
\label{fig:efficiency-first-token}

\vspace{1.5\baselineskip}
\resizebox{!}{0.4\textheight}{\input{media/fig-memory-prefill.pgf}}
\caption{Peak memory of prefill as a percentage of PreLN, by prompt length (rows) and model size (columns).}
\label{fig:memory-prefill}
\end{figure}

\begin{figure}[p]
\centering
\resizebox{!}{0.4\textheight}{\input{media/fig-efficiency-decode.pgf}}
\caption{Decode throughput per generated token as a percentage of PreLN, by context length (rows) and model size (columns).}
\label{fig:efficiency-decode}

\vspace{1.5\baselineskip}
\resizebox{!}{0.4\textheight}{\input{media/fig-memory-decode.pgf}}
\caption{Peak memory of decode as a percentage of PreLN, by context length (rows) and model size (columns).}
\label{fig:memory-decode}
\end{figure}

\section{Extended Results}
\label{app:extended-results}

\subsection{Learning Rate Sweeps}
\label{app:lr-sweeps}

\begin{figure}[H]
\centering
\input{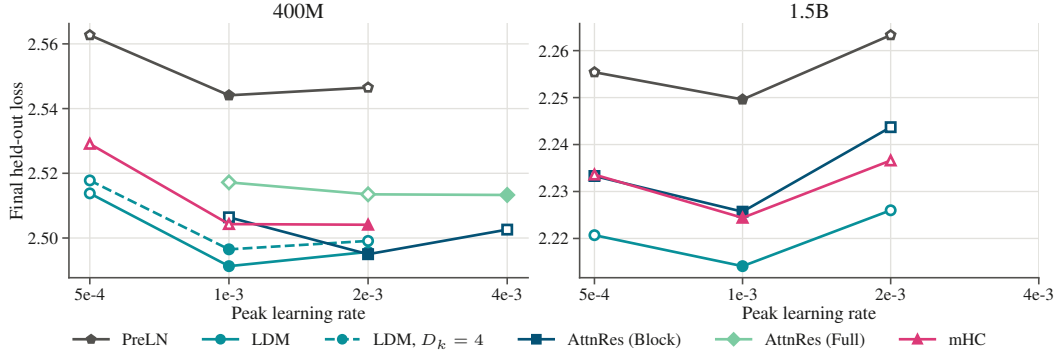}
\caption{Final held-out loss vs peak learning rate (the numbers of \cref{tab:optimal_lrs}). The filled marker is each method's lowest loss.}
\label{fig:lr-sweep}
\end{figure}

\Cref{tab:optimal_lrs} shows the final held-out validation loss of every run in the learning rate sweeps (we describe our protocol in \cref{app:exp_details}). We bold the selected learning rate of each method and scale. \Cref{fig:lr-sweep} plots the losses of each model at each learning rate.

\begin{table}[H]
\centering
\caption{Learning rate sweeps: final held-out FineWeb-Edu cross-entropy of each method, model scale and peak learning rate. The selected learning rate of each method is in bold.}
\label{tab:optimal_lrs}
\small
\setlength{\tabcolsep}{5pt}
\begin{tabular}{llcccc}
\toprule
Scale & Method & $5\times10^{-4}$ & $1\times10^{-3}$ & $2\times10^{-3}$ & $4\times10^{-3}$ \\
\midrule
\multirow{6}{*}{400M}
& PreLN           & 2.5627 & \textbf{2.5441} & 2.5465 & -- \\
& AttnRes (Full)  & --     & 2.5172 & 2.5135 & \textbf{2.5133} \\
& AttnRes (Block) & --     & 2.5064 & \textbf{2.4950} & 2.5026 \\
& mHC             & 2.5291 & 2.5043 & \textbf{2.5041} & -- \\
& LDM (Ours)      & 2.5138 & \textbf{2.4913} & 2.4957 & -- \\
& LDM, $D_k=4$    & 2.5178 & \textbf{2.4965} & 2.4991 & -- \\
\midrule
\multirow{4}{*}{1.5B}
& PreLN           & 2.2554 & \textbf{2.2496} & 2.2633 & -- \\
& AttnRes (Block) & 2.2333 & \textbf{2.2257} & 2.2437 & -- \\
& mHC             & 2.2336 & \textbf{2.2244} & 2.2366 & -- \\
& LDM (Ours)      & 2.2207 & \textbf{2.2141} & 2.2260 & -- \\
\bottomrule
\end{tabular}
\end{table}

\Cref{fig:sweep-400m,fig:sweep-1p5b} show the training loss of every run in \cref{tab:optimal_lrs}.

\begin{figure}[p]
\centering
\input{media/fig-sweep-400m.pgf}
\caption{Training loss of the 400M learning rate sweeps over 50B FineWeb-Edu tokens. The last panel uses each method's selected rate. Top: the whole run. Bottom: from 12.5B tokens (25\% of training) onwards.}
\label{fig:sweep-400m}

\vspace{1.5\baselineskip}
\input{media/fig-sweep-1p5b.pgf}
\caption{Training loss of the 1.5B learning rate sweeps over 100B FineWeb-Edu tokens. Every method's selected rate is $10^{-3}$. Top: the whole run. Bottom: from 24.9B tokens (25\% of training) onwards.}
\label{fig:sweep-1p5b}
\end{figure}

\FloatBarrier

\subsection{Ablations}
\label{app:ablation-definitions}

\begin{figure}[tbp]
\centering
\input{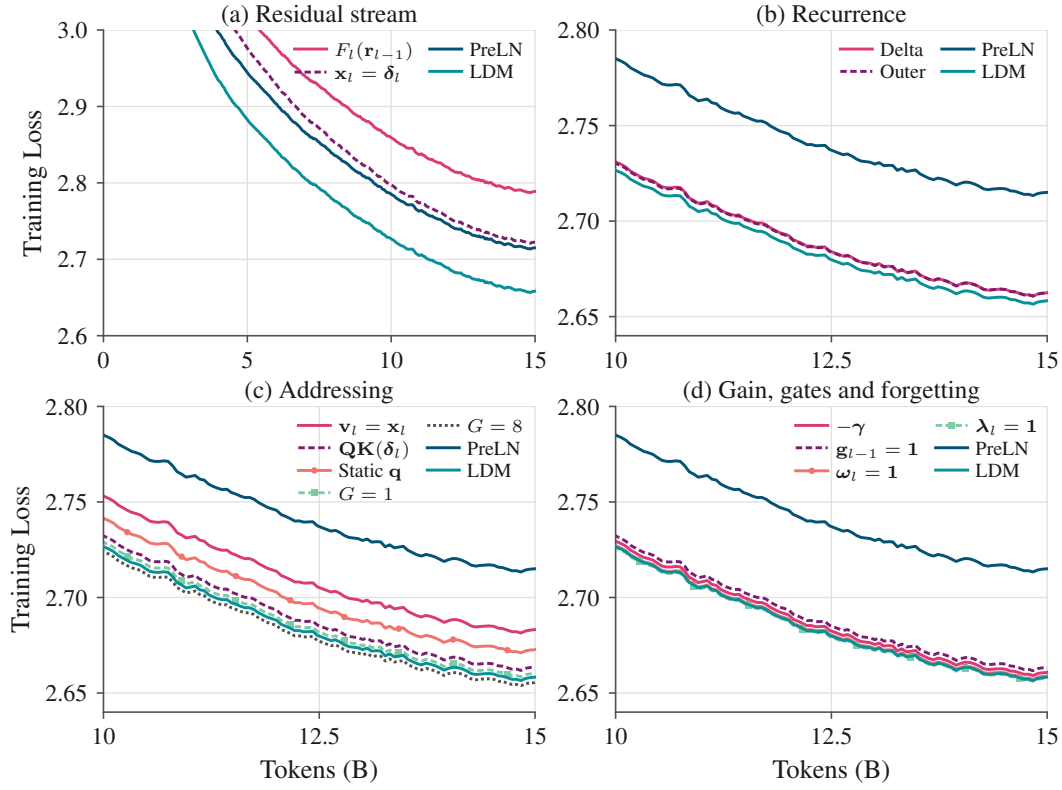}
\caption{Training loss of the 400M ablation runs over 15B FineWeb-Edu tokens, grouped as in \cref{tab:ablation-validation-full}.}
\label{fig:ablations}
\end{figure}

Here, we provide more precise definitions of our ablations. We use the notation of \cref{sec:lDM}, repeated here for clarity. At layer $l$ the incoming residual stream is $\mathbf{x}_{l-1}$, the layer
output is $\bdelta_l$, the state is $\mathbf{S}_l = (\mathbf{x}_l, \mathbf{M}_l, \mathbf P_l)$, and the LDM
equations, with $\mathbf{r}_{l-1}$ denoting the memory readout of the \textread{}, are
\begin{align}
    \mathbf{q}_{l-1} &= \operatorname{Query}(\mathbf{x}_{l-1}),
    &\mathbf{r}_{l-1} &= \boldsymbol{\gamma}_{l-1} \odot \mathbf{M}_{l-1}(\mathbf{g}_{l-1} \odot \mathbf{q}_{l-1}),
    \label{eq:abl-read}\\
    \bdelta_l &= F_l\bigl(\operatorname{Norm}_l(\mathbf{x}_{l-1} + \mathbf{r}_{l-1})\bigr),
    &\mathbf{x}_l &= \mathbf{x}_{l-1} + \bdelta_l,
    \notag\\
    \mathbf{k}_l &= \operatorname{Key}(\mathbf{x}_l), \quad \mathbf{v}_l = \bdelta_l,
    &\mathbf{S}_l &= \mathwrite\bigl(\mathbf{S}_{l-1}, \mathbf{k}_l, \mathbf{v}_l\bigr),
    \label{eq:abl-write}
\end{align}
with the RLS \textwrite{} as in \cref{eq:rls-recursions}, $G$ groups of width $D/G$, the per-group read gate $\mathbf{g}_{l-1} \in \mathbb{R}^G$, the
per-group write weight $\boldsymbol{\omega}_l \in \mathbb{R}^G$ and the learned per-group read
gain $\boldsymbol{\gamma}_{l-1} \in \mathbb{R}^G$. In \Cref{tab:ablation-validation-full}, we show the final held-out loss of every variant, and \cref{fig:ablations} plots their training curves.

\textbf{(a) Residual stream.}
The variant $F_l(\mathbf r_{l-1})$ replaces the stream with the read at every layer (including the final read).
\begin{equation*}
    \bdelta_l = F_l\bigl(\operatorname{Norm}_l(\mathbf{r}_{l-1})\bigr).
\end{equation*}

The variant $\mathbf{x}_l = \bdelta_l$ removes the skip connection after each layer (note that the written value stays $\mathbf{v}_l = \bdelta_l$). Each variant can be understood as removing one of the two identity paths of the residual stream.

\textbf{(b) Recurrence.} We try different \textwrite{} recurrences. Addresses, gates, and the learned forgetting
factor $\lambda_l$ remain the same as in our main variant for comparability. Per group, with the weighted observations
$\overline{\mathbf{k}}_l = \sqrt{\omega_l}\,\mathbf{k}_l$ and
$\overline{\mathbf{v}}_l = \sqrt{\omega_l}\,\mathbf{v}_l$,
\begin{align*}
    \text{Delta rule:}\quad
    &\mathbf{M}_l = \lambda_l\mathbf{M}_{l-1} + \mathbf{u}_l\,\overline{\mathbf{k}}_l^{\top},
    &&\mathbf{u}_l = \overline{\mathbf{v}}_l - \lambda_l\mathbf{M}_{l-1}\overline{\mathbf{k}}_l,\\
    \text{Outer product:}\quad
    &\mathbf{M}_l = \lambda_l\mathbf{M}_{l-1} + \overline{\mathbf{v}}_l\,\overline{\mathbf{k}}_l^{\top}.
\end{align*}
Note that these two do not maintain inverse covariances. RLS can be understood as applying forgetting through $\mathbf P_l$ instead of decaying the state. The delta rule and the outer product update are different approximations to the depth regression problem.

\textbf{(c) Addressing.}
The variant $\mathbf{v}_l = \mathbf{x}_l$ writes the updated residual stream instead of the layer output in \cref{eq:abl-write}. Since $\mathbf{x}_l = \mathbf e + \sum_{i=1}^l \bdelta_i$, the memory stores cumulative sums of earlier outputs rather than the outputs themselves. This means that earlier outputs get stored at every layer iteration.

The variant $\mathbf{QK}(\bdelta_l)$ changes the address sources while the stored value does
not. In particular $\mathbf{q}_{l-1} = \operatorname{Query}(\bdelta_{l-1})$, the most recent layer output,
instead of $\operatorname{Query}(\mathbf{x}_{l-1})$, and $\mathbf{k}_l = \operatorname{Key}(\bdelta_l)$
instead of $\operatorname{Key}(\mathbf{x}_l)$. For the embedding write $\bdelta_0 = \mathbf{x}_0$, so
the two coincide there.

The variant Static $\mathbf q$ replaces the query network by a vector learned for each layer,
$\mathbf{q}_{l-1} = \mathbf{q}^{\mathrm{static}}_{l-1}$, i.e., it is independent of the token. The read gate is computed from the same input as the query, so it is fixed as well, $\mathbf{g}_{l-1} = \mathbf{1}$.

The variants $G = 1$ and $G = 8$ change $G$, the number of groups per token.

\textbf{(d) Gain, gates and forgetting.}
The variant $-\boldsymbol\gamma$ fixes the read gain to one, $\boldsymbol{\gamma}_{l-1} = \mathbf{1}$ in \cref{eq:abl-read}.
The variant $\mathbf{g}_{l-1} = \mathbf{1}$ fixes the read gate to one, so the read is
$\mathbf{r}_{l-1} = \boldsymbol{\gamma}_{l-1} \odot \mathbf{M}_{l-1}\mathbf{q}_{l-1}$.
The variant $\boldsymbol{\omega}_l = \mathbf{1}$ fixes the write weight to one. The corresponding gate heads are removed from the address networks. The variant $\boldsymbol{\lambda}_l = \mathbf{1}$ fixes the forgetting factor to one instead of learning it, so every earlier write keeps its full weight in the least squares objective.

\begin{table}[tbp]
\centering
\caption{Final held-out FineWeb-Edu cross-entropy of the 400M ablation runs (15B tokens). $\Delta$ is relative to LDM.}
\label{tab:ablation-validation-full}
\small
\setlength{\tabcolsep}{8pt}
\renewcommand{\arraystretch}{1.05}
\begin{tabular}{@{}llrr@{}}
\toprule
Group & Variant & Loss $\downarrow$ & $\Delta$ \\
\midrule
Baselines & PreLN & 2.6554 & +0.0567 \\
 & \textbf{LDM} & \textbf{2.5987} & 0.0000 \\
\addlinespace[2pt]
(a) Residual stream & $F_l(\mathbf r_{l-1})$ & 2.7296 & +0.1310 \\
 & $\mathbf{x}_l=\boldsymbol{\delta}_l$ & 2.6611 & +0.0625 \\
\addlinespace[2pt]
(b) Recurrence & Delta & 2.6028 & +0.0041 \\
 & Outer & 2.6037 & +0.0050 \\
\addlinespace[2pt]
(c) Addressing & $\mathbf{v}_l=\mathbf{x}_l$ & 2.6227 & +0.0241 \\
 & $\mathbf{QK}(\boldsymbol{\delta}_l)$ & 2.6040 & +0.0053 \\
 & Static $\mathbf{q}$ & 2.6128 & +0.0141 \\
 & $G=1$ & 2.6007 & +0.0021 \\
 & $G=8$ & 2.5955 & $-$0.0032 \\
\addlinespace[2pt]
(d) Gain, gates and forgetting & $-\boldsymbol\gamma$ & 2.6014 & +0.0027 \\
 & $\mathbf{g}_{l-1}=\mathbf{1}$ & 2.6035 & +0.0048 \\
 & $\boldsymbol{\omega}_l=\mathbf{1}$ & 2.5992 & +0.0006 \\
 & $\boldsymbol\lambda_l=\mathbf 1$ & 2.5981 & $-$0.0006 \\
\bottomrule
\end{tabular}
\end{table}

\FloatBarrier

\subsection{Different Aspect Ratios}
\label{app:depthbench}

We follow the exact training protocol of DepthBench~\citep{wang2026depthbenchmeasuringresidualconnections}. All models are Llama-style decoder-only transformers, trained with a context of 2,048 tokens on FineWeb-Edu with a global batch of 512 sequences (about 1M tokens per step), AdamW ($\beta_1=0.9$, $\beta_2=0.95$, weight decay $0.1$) and a cosine schedule with 10\% warmup. At 400M, DepthBench tunes the learning rate of each baseline and finds $2\times10^{-3}$ optimal for most of them. At 1.6B, it uses $5\times10^{-4}$ for all methods without tuning. We use the same values for LDM and do not tune them. At 400M and at 1.6B, we train for 8B tokens (7,600 steps) and 32B tokens (30,400 steps), respectively. We report the final held-out perplexity on DepthBench's FineWeb-Edu evaluation set. The baseline results are those published by DepthBench. They are exactly comparable to ours, since our runs use the identical training setup, including code, and data and its ordering. The model shapes are listed in \cref{tab:depthbench-shapes}.

LDM uses our main configuration except for weight decay, which follows DepthBench and also applies to the RMSNorm gains. Only the read gains, the forgetting logits and the biases of the address networks are not decayed. The group width equals the attention head width of each shape, which corresponds to 16 groups at every shape. We use $H=32$ and $H=64$ for the 400M and the 1.6B models, respectively.

\begin{figure}[h]
\centering
\resizebox{0.75\linewidth}{!}{\input{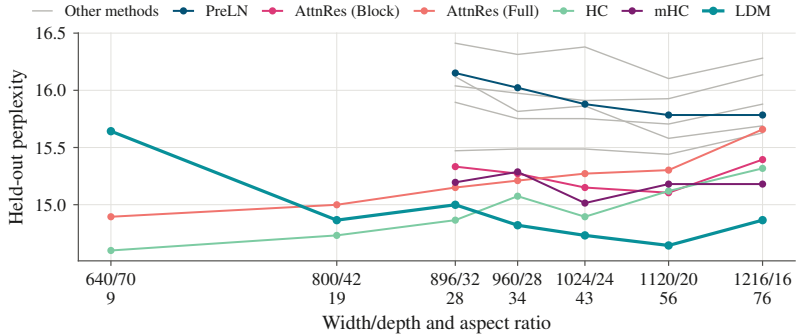}}
\caption{Final held-out perplexity on DepthBench at 400M parameters vs the aspect ratio (width/depth), from 16 to 70 layers. Tick labels give the width/depth of each shape and aspect ratio.}
\label{fig:depthbench-appendix}
\end{figure}

\Cref{fig:depthbench-appendix} extends the 400M panel of \cref{fig:depthbench} to DepthBench's ``extreme'' aspect ratios, with 42 and 70 layers, where we compare against HC and AttnRes (Full) (the only methods reported by DepthBench at these shapes, which appeared to prefer extremely deep shapes). At these shapes, HC is ahead of LDM, and the perplexity of LDM increases at 70 layers. We note that these shapes were run for the purpose of a stress test and do not represent aspect ratios commonly used in models. Moreover, as found in \cref{sec:depthbench}, on bigger models, LDM seems to prefer deeper shapes, and the advantage of HC vanishes.

\begin{table}[h]
\centering
\caption{DepthBench shapes. The 400M shapes have 16 query and 16 key/value heads, the 1.6B shapes 16 query and 8 key/value heads.}
\label{tab:depthbench-shapes}
\small
\setlength{\tabcolsep}{5pt}
\begin{tabular}{lrrrrrrr}
\toprule
Tier & Layers & Width $D$ & $D/L$ & Head width & FFN width & Backbone & LDM adds \\
\midrule
\multirow{7}{*}{400M}
 & 16 & 1,216 & 76.0 & 76 & 3,248 & 406.5M & 2.97M (0.73\%) \\
 & 20 & 1,120 & 56.0 & 70 & 2,992 & 414.1M & 3.42M (0.83\%) \\
 & 24 & 1,024 & 42.7 & 64 & 2,736 & 405.4M & 3.78M (0.93\%) \\
 & 28 & 960 & 34.3 & 60 & 2,560 & 406.3M & 4.16M (1.02\%) \\
 & 32 & 896 & 28.0 & 56 & 2,400 & 399.4M & 4.47M (1.12\%) \\
 & 42 & 800 & 19.0 & 50 & 2,144 & 404.2M & 5.30M (1.31\%) \\
 & 70 & 640 & 9.1 & 40 & 1,712 & 409.2M & 7.31M (1.79\%) \\
\midrule
\multirow{3}{*}{1.6B}
 & 28 & 2,048 & 73.1 & 128 & 6,144 & 1.615B & 16.25M (1.01\%) \\
 & 40 & 1,728 & 43.2 & 108 & 5,184 & 1.607B & 19.73M (1.23\%) \\
 & 54 & 1,504 & 27.9 & 94 & 4,512 & 1.617B & 23.37M (1.45\%) \\
\bottomrule
\end{tabular}
\end{table}

\FloatBarrier

\section{LDM Details}
\label{app:ldm-details}

\subsection{Derivation of the RLS Recursions}
\label{app:rls-derivation}

For conciseness, fix one group, with $\mathbf M_l\in\mathbb R^{D_v\times D_k}$ and
$\mathbf P_l\in\mathbb R^{D_k\times D_k}$. \textwrite{} $l$ maps the state after
\textwrite{} $l-1$ to the state after \textwrite{} $l$. \textwrite{} $0$ stores the embedding.
Let $\alpha>0$, $\lambda_l>0$, and $\omega_l>0$.
The weighted objective is as in \cref{eq:ridge-least-squares}. Define
\begin{equation*}
\begin{aligned}
\mathbf A_l
&=\sum_{i=0}^l\left(\prod_{j=i+1}^l\lambda_j\right)
\omega_i\mathbf k_i\mathbf k_i^\top
+\left(\prod_{j=0}^l\lambda_j\right)\alpha^{-1}\mathbf I,\\
\mathbf B_l
&=\sum_{i=0}^l\left(\prod_{j=i+1}^l\lambda_j\right)
\omega_i\mathbf v_i\mathbf k_i^\top.
\end{aligned}
\end{equation*}
The objective gradient is $2(\mathbf M\mathbf A_l-\mathbf B_l)$.
Since $\mathbf A_l$ is positive definite, the unique minimizer is
$\mathbf M_l=\mathbf B_l\mathbf A_l^{-1}$.
With $\mathbf A_{-1}=\alpha^{-1}\mathbf I$ and $\mathbf B_{-1}=\mathbf 0$,
\begin{equation*}
\mathbf A_l=\lambda_l\mathbf A_{l-1}
+\overline{\mathbf k}_l\overline{\mathbf k}_l^\top,
\qquad
\mathbf B_l=\lambda_l\mathbf B_{l-1}
+\overline{\mathbf v}_l\overline{\mathbf k}_l^\top,
\end{equation*}
where $\overline{\mathbf k}_l=\sqrt{\omega_l}\mathbf k_l$ and
$\overline{\mathbf v}_l=\sqrt{\omega_l}\mathbf v_l$.

\textbf{Forward inverse covariance update.}
Set $\mathbf P_l=\mathbf A_l^{-1}$. Sherman--Morrison gives
\begin{equation*}
\mathbf P_l=\frac{1}{\lambda_l}
\left(\mathbf P_{l-1}
-\frac{\mathbf P_{l-1}\overline{\mathbf k}_l
\overline{\mathbf k}_l^\top\mathbf P_{l-1}}
{\lambda_l+\overline{\mathbf k}_l^\top\mathbf P_{l-1}\overline{\mathbf k}_l}
\right).
\end{equation*}
The denominator is strictly positive. With the gain from
\cref{eq:pseudo-updates}, it is easy to see that this is the inverse covariance update in
\cref{eq:rls-recursions}. In particular,
$\overline{\mathbf k}_l^\top\mathbf P_l=\bm\kappa_l^\top$.

\textbf{Forward memory update.}
Using $\mathbf B_{l-1}=\mathbf M_{l-1}\mathbf A_{l-1}$, we have that
\begin{equation*}
\begin{aligned}
\mathbf M_l
&=(\lambda_l\mathbf B_{l-1}
+\overline{\mathbf v}_l\overline{\mathbf k}_l^\top)\mathbf P_l\\
&=\mathbf M_{l-1}
(\mathbf I-\overline{\mathbf k}_l\overline{\mathbf k}_l^\top\mathbf P_l)
+\overline{\mathbf v}_l\overline{\mathbf k}_l^\top\mathbf P_l\\
&=\mathbf M_{l-1}
+(\overline{\mathbf v}_l-\mathbf M_{l-1}\overline{\mathbf k}_l)
\bm\kappa_l^\top
=\mathbf M_{l-1}+\mathbf u_l\bm\kappa_l^\top.
\end{aligned}
\end{equation*}

\textbf{Gating and regularization.}
The dynamic weight $\omega_l$ sets the ``importance'' of observation $l$ and
$\lambda_l$ discounts all earlier observations (and the initial ridge penalty).
A fixed ridge penalty would instead introduce an additional full-rank term
$(1-\lambda_l)\alpha^{-1}\mathbf I$ into the covariance recurrence.
That alternative remains solvable, but no longer admits the Sherman--Morrison-based update. 

\subsection{Boundary Cases}
\label{app:boundary}
Before writing the embedding $\mathbf e$, we initialize the memory as
\begin{equation*}
\mathbf M_{-1}=\mathbf{0},
\qquad
\mathbf{P}_{-1}=\alpha\mathbf{I}.
\end{equation*}
The initial value written into the memory is $\mathbf v_0 = \mathbf{e}$, and its key and weight are generated from $\mathbf{e}$ using the same write-side parametrization, yielding
\begin{equation*}
(\mathbf e, \mathbf M_{0}, \mathbf P_0)
=
\mathwrite\left(
(\mathbf 0, \mathbf M_{-1}, \mathbf P_{-1}),
\mathbf{k}_{0},
\mathbf v_{0}, \bm{\omega}_0, \bm{\lambda}_0
\right).
\end{equation*}
After the final transformer layer, one final \textread{} is added to the residual stream. We generate
$\mathbf{q}_{\mathrm{out}}$ from
$\mathbf{x}_{L}$ using the same parametrization and define
\begin{equation*}
\mathbf{x}_{\mathrm{out}}
=
\mathread\left(
(\mathbf x_L, \mathbf M_{L}, \mathbf P_L),
\mathbf{q}_{\mathrm{out}}
\right).
\end{equation*}
The final representation $\mathbf x_{\mathrm{out}}$ is then passed to the LM head.

\section{Hardware-Aware Implementation}
\label{app:efficiency}

This appendix provides an expanded explanation of \cref{sec:efficiency}. All quantities below are per token and per group. \Cref{alg:ldm} summarizes our implementation.

\textbf{Notation.} We fix one group, and use $\overline{\mathbf k}_l=\sqrt{\omega_l}\mathbf k_l$ and $\overline{\mathbf v}_l=\sqrt{\omega_l}\mathbf v_l$ for the weighted key and value of a \textwrite{}, and $\tilde{\mathbf q}_l=\gamma_l\,g_l\,\mathbf q_l$ for the gated query. The read is then $\mathbf r_l=\mathbf M_l\tilde{\mathbf q}_l$. We also use $\mathrm d a=\nabla_a\mathcal L$ for brevity. 

When $a$ is used in several places, $\nabla_a\mathcal L$ is a sum with one term per use (by the chain rule). We use a superscript $\mathbf M_l$ to keep only the terms of the uses of $a$ in computing $\mathbf u_l$ and $\mathbf M_l$ (the memory update), and a superscript $\mathbf P_l$ to keep only those in computing $\bm\kappa_l$ and $\mathbf P_l$ (the gain and inverse covariance update). For example, $\bm\kappa_l$ is used to compute $\mathbf M_l$ and $\mathbf P_l$, $\overline{\mathbf k}_l$ to compute $\mathbf u_l$, $\bm\kappa_l$ and $\mathbf P_l$, and $\mathbf M_l$ to compute $\mathbf u_{l+1}$, $\mathbf M_{l+1}$ and the read $\mathbf r_l$, so
\begin{equation*}
\mathrm d\bm\kappa_l=\mathrm d\bm\kappa_l^{\mathbf M_l}+\mathrm d\bm\kappa_l^{\mathbf P_l},
\qquad
\mathrm d\overline{\mathbf k}_l=\mathrm d\overline{\mathbf k}_l^{\mathbf M_l}+\mathrm d\overline{\mathbf k}_l^{\mathbf P_l},
\qquad
\mathrm d\mathbf M_l=\mathrm d\mathbf M_l^{\mathbf M_{l+1}}+\mathrm d\mathbf r_l\tilde{\mathbf q}_l^\top.
\end{equation*}

\textbf{Steps.} We call the \textwrite{} at layer $l$, fused with the \textread{} for layer $l+1$, a ``step''. Step $l$ takes the state $(\mathbf M_{l-1},\mathbf P_{l-1})$, the weighted key and value $\overline{\mathbf k}_l$ and $\overline{\mathbf v}_l$, the gated query $\tilde{\mathbf q}_l$ and the forgetting factor $\lambda_l$. It produces the next state $(\mathbf M_l,\mathbf P_l)$ and the read $\mathbf r_l=\mathbf M_l\tilde{\mathbf q}_l$, which is fed into layer $l+1$.

Step $0$ writes the embedding ($\mathbf v_0=\mathbf e$) into the initial state $(\mathbf M_{-1},\mathbf P_{-1})=(\mathbf 0,\alpha\mathbf I)$, producing the state $(\mathbf M_0,\mathbf P_0)$ after the write, as well as the read $\mathbf r_0$ for layer $1$. Step $L$ ends with the final read into the LM head, with query $\mathbf q_{\mathrm{out}}$ (\cref{app:boundary}) written as $\mathbf q_L$. An $L$-layer model then has $L+1$ steps.

\subsection{Unwinding}
\label{app:unwinding}

\textbf{Needed gradients.} As usual, the backward pass runs the forward steps in reverse order. At step $l$, the backward pass ``receives'' three gradients:
\begin{itemize}[itemsep=1pt,topsep=2pt,parsep=0pt,leftmargin=1.5em]
    \item $\mathrm d\mathbf r_l$, the gradient of the read $\mathbf r_l=\mathbf M_l\tilde{\mathbf q}_l$ of step $l$ (which is fed into layer $l+1$, or into the LM head for $l=L$),
    \item $\mathrm d\mathbf P_l$, the gradient of $\mathbf P_l$,
    \item $\mathrm d\mathbf M_l^{\mathbf M_{l+1}}=\mathrm d\mathbf M_l-\mathrm d\mathbf r_l\tilde{\mathbf q}_l^\top$, the gradient of $\mathbf M_l$ without the term corresponding to its read $\mathbf r_l$ (which step $l$ adds through $\mathrm d\mathbf r_l$).
\end{itemize}
Since $\mathbf M_L$ is only used by the final read and $\mathbf P_L$ is not used, $\mathrm d\mathbf M_L^{\mathbf M_{L+1}}=\mathbf 0$ and $\mathrm d\mathbf P_L=\mathbf 0$. From these, the backward pass of step $l$ must produce
\begin{itemize}[itemsep=1pt,topsep=2pt,parsep=0pt,leftmargin=1.5em]
    \item $\mathrm d\tilde{\mathbf q}_l$, $\mathrm d\overline{\mathbf v}_l$, $\mathrm d\overline{\mathbf k}_l$ and $\mathrm d\lambda_l$, the gradients of the inputs of step $l$,
    \item $\mathrm d\mathbf M_{l-1}^{\mathbf M_l}$ and $\mathrm d\mathbf P_{l-1}$, which are passed on to step $l-1$ (for $l>0$).
\end{itemize}
Since step $l$ computes $\mathbf u_l=\overline{\mathbf v}_l-\mathbf M_{l-1}\overline{\mathbf k}_l$, $\mathbf M_l=\mathbf M_{l-1}+\mathbf u_l\bm\kappa_l^\top$ and $\mathbf r_l=\mathbf M_l\tilde{\mathbf q}_l$ (\cref{eq:pseudo-updates,eq:rls-recursions}), these gradients are
\begin{align}
\mathrm d\tilde{\mathbf q}_l&=\mathbf M_l^\top\mathrm d\mathbf r_l, \quad
\mathrm d\overline{\mathbf v}_l=\mathrm d\mathbf u_l=\mathrm d\mathbf M_l^{\mathbf M_{l+1}}\bm\kappa_l+(\bm\kappa_l^\top\tilde{\mathbf q}_l)\,\mathrm d\mathbf r_l,\nonumber\\
\mathrm d\bm\kappa_l^{\mathbf M_l}&=(\mathrm d\mathbf M_l^{\mathbf M_{l+1}})^\top\mathbf u_l+(\mathbf u_l^\top\mathrm d\mathbf r_l)\,\tilde{\mathbf q}_l, \quad
\mathrm d\overline{\mathbf k}_l=-\mathbf M_{l-1}^\top\mathrm d\overline{\mathbf v}_l+\mathrm d\overline{\mathbf k}_l^{\mathbf P_l},\nonumber\\
\mathrm d\mathbf M_{l-1}^{\mathbf M_l}&=\mathrm d\mathbf M_l^{\mathbf M_{l+1}}+\mathrm d\mathbf r_l\tilde{\mathbf q}_l^\top-\mathrm d\overline{\mathbf v}_l\overline{\mathbf k}_l^\top,
\label{eq:step-vjp}
\end{align}
where $\mathrm d\overline{\mathbf k}_l^{\mathbf P_l}$ is obtained by backpropagating $\mathrm d\bm\kappa_l^{\mathbf M_l}$ and $\mathrm d\mathbf P_l$ through \cref{eq:pseudo-updates,eq:rls-recursions} (which also produces $\mathrm d\lambda_l$ and $\mathrm d\mathbf P_{l-1}$). Note that these need the inverse covariance $\mathbf P_{l-1}$ but do not depend on the state.

\textbf{Unwinding the state.} The only states in \cref{eq:step-vjp} are $\mathbf M_l$, in $\mathrm d\tilde{\mathbf q}_l$, and $\mathbf M_{l-1}$, in $\mathrm d\overline{\mathbf k}_l$. Instead of saving them, the backward pass starts from the final state $\mathbf M_L$ of the forward pass and unwinds it in place. When it reaches step $l$, it holds $\mathbf M_l$ and recovers the previous state from the saved pseudo-value and gain of the step by unwinding (\cref{eq:unwinding}),
\begin{equation*}
\mathbf M_{l-1}=\mathbf M_l-\mathbf u_l\bm\kappa_l^\top,
\end{equation*}
which it uses for the gradients of step $l$ and then for step $l-1$. It is therefore enough to save, for each step, $\mathbf u_l$, $\bm\kappa_l$, $\overline{\mathbf k}_l$, $\tilde{\mathbf q}_l$, $\lambda_l$ and $\mathbf P_{l-1}$ (which we refer to as ``context'' in \cref{sec:efficiency}), and a single state per token is kept in memory. The inverse covariance is saved (instead of unwound). Note that $\mathbf P_{l-1}$ has only $D_k(D_k+1)/2$ entries per group (we only save the nonredundant part, which corresponds to a triangle given it is symmetric).

\subsection{Windowing}
\label{app:windowing}

We group steps into windows $\mathcal W=\{b+1,\ldots,e\}$ of at most $E$ steps, with $b=-1,E-1,2E-1,\ldots$ and $e=\min(b+E,L)$. Inside a window, $\mathbf M_i=\mathbf M_b+\sum_{b<m\le i}\mathbf u_m\bm\kappa_m^\top$, and hence every product of a state with a vector can be computed from the state $\mathbf M_b$ before the window and the pseudo-values and gains of the window (\cref{eq:served}),
\begin{equation}
\label{eq:window-products}
\mathbf M_i\mathbf x=\mathbf M_b\mathbf x+\sum_{b<m\le i}\mathbf u_m\,(\bm\kappa_m^\top\mathbf x),
\qquad
\mathbf M_i^\top\mathbf y=\mathbf M_b^\top\mathbf y+\sum_{b<m\le i}\bm\kappa_m\,(\mathbf u_m^\top\mathbf y).
\end{equation}

\textbf{Forward.} For step $l$, the \textwrite{} needs the pseudo-value $\mathbf u_l=\overline{\mathbf v}_l-\mathbf M_{l-1}\overline{\mathbf k}_l$ for the update $\mathbf M_l=\mathbf M_{l-1}+\mathbf u_l\bm\kappa_l^\top$, and the \textread{} is $\mathbf r_l=\mathbf M_l\tilde{\mathbf q}_l$. Hence, it needs the products $\mathbf M_{l-1}\overline{\mathbf k}_l$ and $\mathbf M_l\tilde{\mathbf q}_l$, which, as shown in \cref{eq:window-products}, can be computed from $\mathbf M_b$. In particular, the states inside the window are never formed, and only the last step of the window writes the new state, $\mathbf M_e=\mathbf M_b+\sum_{b<i\le e}\mathbf u_i\bm\kappa_i^\top$.

\textbf{Backward.} The windows are naturally processed in reverse order. A single unwinding, $\mathbf M_b=\mathbf M_e-\sum_{b<i\le e}\mathbf u_i\bm\kappa_i^\top$, can recover the state before the window. After recovering it, $\mathbf M_i^\top\mathrm d\mathbf r_i$ and $\mathbf M_{i-1}^\top\mathrm d\overline{\mathbf v}_i$ of \cref{eq:step-vjp} are computed from it by \cref{eq:window-products}. Note that the intermediate state gradients inside the window are not needed either. Unrolling the recursion for $\mathrm d\mathbf M_{l-1}^{\mathbf M_l}$ in \cref{eq:step-vjp} over the window gives
\begin{equation}
\label{eq:state-gradient}
\mathrm d\mathbf M_i^{\mathbf M_{i+1}}=\mathrm d\mathbf M_e^{\mathbf M_{e+1}}+\sum_{j=i+1}^{e}\bigl(\mathrm d\mathbf r_j\tilde{\mathbf q}_j^\top-\mathrm d\overline{\mathbf v}_j\overline{\mathbf k}_j^\top\bigr),
\qquad b\le i\le e,
\end{equation}
and substituting it into $\mathrm d\overline{\mathbf v}_i$ and $\mathrm d\bm\kappa_i^{\mathbf M_i}$ of \cref{eq:step-vjp} gives
\begin{align}
\mathrm d\overline{\mathbf v}_i=\mathrm d\mathbf u_i
&=\mathrm d\mathbf M_e^{\mathbf M_{e+1}}\bm\kappa_i+\sum_{j\ge i}(\bm\kappa_i^\top\tilde{\mathbf q}_j)\,\mathrm d\mathbf r_j
-\sum_{j>i}(\bm\kappa_i^\top\overline{\mathbf k}_j)\,\mathrm d\overline{\mathbf v}_j,\nonumber\\
\mathrm d\bm\kappa_i^{\mathbf M_i}&=(\mathrm d\mathbf M_e^{\mathbf M_{e+1}})^\top\mathbf u_i
+\sum_{j\ge i}(\mathbf u_i^\top\mathrm d\mathbf r_j)\,\tilde{\mathbf q}_j
-\sum_{j>i}(\mathbf u_i^\top\mathrm d\overline{\mathbf v}_j)\,\overline{\mathbf k}_j,
\label{eq:window-vjp}
\end{align}
where the sums are over the steps of the given window. These are computed for $i=e,e-1,\ldots,b+1$ and use $\mathrm d\mathbf M_e^{\mathbf M_{e+1}}$ only through $\mathrm d\mathbf M_e^{\mathbf M_{e+1}}\bm\kappa_i$ and $(\mathrm d\mathbf M_e^{\mathbf M_{e+1}})^\top\mathbf u_i$. In particular, the only state gradient that is formed for a window is its $\mathrm d\mathbf M_e^{\mathbf M_{e+1}}$. For the last window, it is $\mathrm d\mathbf M_L^{\mathbf M_{L+1}}=\mathbf 0$. The other windows form it at their start, from the state gradients of their later windows $\{e+1,\ldots,e'\}$, where $e'=\min(e+E,L)$, by \cref{eq:state-gradient}:
\begin{equation*}
\mathrm d\mathbf M_e^{\mathbf M_{e+1}}=\mathrm d\mathbf M_{e'}^{\mathbf M_{e'+1}}+\sum_{j=e+1}^{e'}\bigl(\mathrm d\mathbf r_j\tilde{\mathbf q}_j^\top-\mathrm d\overline{\mathbf v}_j\overline{\mathbf k}_j^\top\bigr).
\end{equation*}
Both the state and $\mathrm d\mathbf M_e^{\mathbf M_{e+1}}$ are then updated only once per window. This reduces state-sized traffic at the cost of accessing more vector-sized values. We use $E = 4$ in our main experiments.

\begin{algorithm}[t]
\caption{Forward and backward pass with unwinding and windowing (fixing one token and one group). $\mathbf M$, $\mathbf P$, $\mathrm d\mathbf M$ and $\mathrm d\mathbf P$ are buffers updated in place.}
\label{alg:ldm}
\small
\begin{algorithmic}[1]
\Statex \textbf{Forward}
\State $\mathbf M\gets\mathbf 0$, \quad $\mathbf P\gets\alpha\mathbf I$
\For{each window $\mathcal W=\{b+1,\ldots,e\}$} \Comment{$\mathbf M=\mathbf M_b$}
  \For{$l=b+1,\ldots,e$}
    \State $\mathbf v_l\gets\bdelta_l$ ($\mathbf v_0\gets\mathbf e$), and $\mathbf k_l$, $\sqrt{\omega_l}$, $\tilde{\mathbf q}_l$ from the address networks
    \State $\overline{\mathbf k}_l\gets\sqrt{\omega_l}\,\mathbf k_l$, \quad $\overline{\mathbf v}_l\gets\sqrt{\omega_l}\,\mathbf v_l$
    \State $\bm\kappa_l$ by \cref{eq:pseudo-updates}, \quad save $\mathbf P_{l-1}=\mathbf P$, \quad $\mathbf P\gets\mathbf P_l$ by \cref{eq:rls-recursions}
    \State $\mathbf u_l$ and $\mathbf r_l$ by \cref{eq:window-products}
    \State save $\mathbf u_l$, $\bm\kappa_l$, $\overline{\mathbf k}_l$, $\tilde{\mathbf q}_l$, $\lambda_l$
    \If{$l=e$}
      \State $\mathbf M\gets\mathbf M+\sum_{b<i\le e}\mathbf u_i\bm\kappa_i^\top$ \Comment{$\mathbf M=\mathbf M_e$}
    \EndIf
  \EndFor
\EndFor
\Statex \textbf{Backward}
\State $\mathrm d\mathbf M\gets\mathbf 0$, \quad $\mathrm d\mathbf P\gets\mathbf 0$, \quad $\mathrm d\mathbf r_L$ from the LM head \Comment{$\mathbf M=\mathbf M_L$}
\For{each window $\mathcal W=\{b+1,\ldots,e\}$, in reverse}
  \If{$e<L$} \Comment{$e'=\min(e+E,L)$}
    \State $\mathrm d\mathbf M\gets\mathrm d\mathbf M+\sum_{e<j\le e'}\bigl(\mathrm d\mathbf r_j\tilde{\mathbf q}_j^\top-\mathrm d\overline{\mathbf v}_j\overline{\mathbf k}_j^\top\bigr)$
  \EndIf
  \State $\mathbf M\gets\mathbf M-\sum_{b<i\le e}\mathbf u_i\bm\kappa_i^\top$ \Comment{$\mathbf M=\mathbf M_b$, \ $\mathrm d\mathbf M=\mathrm d\mathbf M_e^{\mathbf M_{e+1}}$}
  \For{$i=e,\ldots,b+1$}
    \State $\mathrm d\overline{\mathbf v}_i$ and $\mathrm d\bm\kappa_i^{\mathbf M_i}$ by \cref{eq:window-vjp}
    \State $\mathrm d\overline{\mathbf k}_i^{\mathbf P_i}$, $\mathrm d\lambda_i$ and $\mathrm d\mathbf P\gets\mathrm d\mathbf P_{i-1}$ by backpropagating through \cref{eq:pseudo-updates,eq:rls-recursions}
    \State $\mathrm d\tilde{\mathbf q}_i$ and $\mathrm d\overline{\mathbf k}_i$ by \cref{eq:step-vjp,eq:window-products}
    \State $\mathrm d\mathbf r_{i-1}$ ($i>0$) by backpropagating through the address networks and layer $i$
  \EndFor
\EndFor
\end{algorithmic}
\end{algorithm}

\section{AI Use Statement}
\label{app:ai-use}
We used generative AI tools for \LaTeX{} formatting and implementation of figures, to help draft parts of the text that we edited afterward, to polish text, and, through discussions with them, to refine ideas, methodology, and hypotheses. We also used them for some mathematical calculations and derivations, and for code editing and experiment execution (at the direction of the authors). Additionally, they were used to assist with literature search. A particularly relevant use of AI was kernel development. With substantial human involvement, we used AI to explore possible algorithmic implementations and, within the resulting algorithmic framework, to search the space of kernels and to write and tune them. We validated the kernels extensively with a wide range of synthetic tests and real training runs. We have reviewed all AI-assisted work and take responsibility for the final content of this paper.

\end{document}